\documentclass{article}

\PassOptionsToPackage{numbers}{natbib}
\usepackage[preprint]{neurips_2026}

\usepackage[utf8]{inputenc} 
\usepackage[T1]{fontenc}    
\usepackage[table,x11names]{xcolor}  
\usepackage{url}            
\usepackage{booktabs}       
\usepackage{amsfonts}       
\usepackage{nicefrac}       
\usepackage{microtype}      
\usepackage{bm}
\usepackage{tabularx}
\usepackage{graphicx}
\usepackage{array}
\usepackage{multirow}
\usepackage{multicol}
\usepackage{ragged2e}       
\usepackage{xfp}            
\usepackage{listings}
\usepackage{svg}
\usepackage{wrapfig}
\usepackage{todonotes}
\usepackage{float}          
\usepackage{placeins}       
\usepackage{needspace}
\usepackage{titletoc}
\usepackage{enumitem}
\setlist{leftmargin=5.5mm}
\usepackage{longtable,tabu}
\usepackage{algorithm}
\usepackage{algpseudocode}
\usepackage{amsmath} 
\usepackage{hyperref}       

\usepackage{longtable}
\usepackage{makecell}

\usepackage{cleveref}
\usepackage{xurl}

\usepackage{pdflscape}

\definecolor{agentic}{HTML}{1F6F5E}      
\definecolor{modagentic}{HTML}{9B4B1E}   
\definecolor{nonagentic}{HTML}{2E4E7C}   
\definecolor{coral}{HTML}{FF7F50}
\definecolor{authorheading}{HTML}{244B6B}

\colorlet{outline_comment}{blue}

\newcommand{\MainEvalModelCount}{8}
\newcommand{\MainEvalConfigurationCount}{16}
\newcommand{\MainEvalBenchmarkCount}{54}
\newcommand{\MainEvalTaskCount}{6{,}627}
\newcommand{\MainEvalFamilyHarnessCount}{3}
\newcommand{\MainEvalHarnessImplementationCount}{4}
\newcommand{\HarborIndexTaskCount}{82}
\newcommand{\HarborIndexBenchmarkCount}{29}
\newcommand{\HarborIndexModelCount}{9}
\newcommand{\HarborIndexRolloutCount}{1{,}476}
\newcommand{\HarborIndexBestPassRate}{28.0}
\newcommand{\HarborIndexBestConfiguration}{GPT-5.5 with Codex}

\newcommand{\titleauthorheading}[1]{%
  \par\smallskip
  {\centering\color{authorheading}\bfseries #1\par}%
  \vspace{0.35mm}%
}
\newcommand{\titleauthornames}[1]{%
  {\fontsize{7.6pt}{8.65pt}\selectfont\RaggedRight #1\par}%
}
\newcommand{\projectleadsymbol}{%
  \textcolor{authorheading}{\ensuremath{\spadesuit}}%
}
\newcommand{\corecontributorsymbol}{%
  \textcolor{authorheading}{\ensuremath{\diamondsuit}}%
}
\newcommand{\projectleadmark}{\textsuperscript{\projectleadsymbol}}
\newcommand{\corecontributormark}{\textsuperscript{\corecontributorsymbol}}
\newenvironment{titleauthorblock}{%
  \begin{minipage}{0.97\textwidth}
  \setlength{\parindent}{0pt}%
  \setlength{\parskip}{0pt}%
  \normalfont
}{%
  \end{minipage}%
}

\newcommand{\appendixauthorgroup}[2]{%
  \needspace{4\baselineskip}%
  \par\medskip
  {\centering\color{authorheading}\bfseries #1\par}%
  \vspace{0.5mm}%
  \noindent\RaggedRight #2\par
}
\definecolor{RoyalBlue}{rgb}{0.0, 0.14, 0.4}
\hypersetup{
    colorlinks=true,    
    linkcolor=RoyalBlue,     
    citecolor=RoyalBlue,      
    urlcolor=RoyalBlue       
    }

\title{Harbor Adapters and Harbor-Index: \\Infrastructure and a Curated Meta-Dataset for Large-Scale Agentic Evaluation}

\author{
\begin{titleauthorblock}
\titleauthorheading{Organization \& Execution Team}
\titleauthornames{Lin Shi\projectleadmark, Haowei Lin\projectleadmark, Zixuan Zhu\corecontributormark, Xiaoyue Zhou\corecontributormark, Xiang Li\corecontributormark, Xiangning Lin\corecontributormark, Yaxuan Deng\corecontributormark, Han Xu\corecontributormark, Yuangang Li\corecontributormark, Shanda Li\corecontributormark, Zizhao Chen\corecontributormark, Hanwen Xing\corecontributormark, Harsh Raj, Bo Chen, Quan Shi, Steven Dillmann, Yipeng Gao, Puneesh Khanna, Ruofan Lu, Chao Beyond Zhou, Michael Yang, Robert Zhang, Siyuan Chai, Jiayu Chang, Yizhao Chen, Xiaokun Chen, Yiwei Dai, Wenting Yang, Hange Liu, Minghao Liu, Zihan Wang}

\titleauthorheading{Adapter Contributors}
\titleauthornames{Adnan El Assadi, Benedikt Stroebl, E. Kelly Buchanan, Han Meng, Junwei He, Longxuan Yu, Radin Shayanfar, Yukyung Lee, Zhikang Dong, Allen G Hart, Anjiang Wei, Anurag Kashyap, Arpandeep Khatua, Audrey Jixin Zheng, Chengrui Ma, David Heineman, Dubing Chen, Hai-Anh Trinh, Haishuo Fang, Hefan Zhang, Hui Shen, Issa Sugiura, Jiankai Sun, Jiechao Gao, Junhong Lin, Junnan Li, Kai Yang, Lei Hsiung, Maoyu Wang, Mengze Tang, Nabil Omi, Negin Raoof, Nicholas Edwards, Octavia Guo, Orfeas Menis Mastromichalakis, Pengliang Ji, Przemysław Hejman, Qi Qi, Qunshu Lin, Richard Zhuang, Rui Yang, Ruichen Zheng, Ryan Marten, Shaghayegh Fazliani, Shizheng Hou, Sicong Jiang, Sijie Li, Boqin Yuan, Michael Glass, Song Bian, Terry Yue Zhuo, Tianqing Wu, Tom Tang, Wanjia Zhao, Weihao Xuan, Wenhua Liang, Xian Liu, Xin Lan, Xuan Zhang, Xuandong Zhao, Yanchuan Tang, Yifan Jiang, Yijiang Li, Yitong Guan, Yizhi Li, Yonghui Liu, Yuheng Tang, Yujun (Audrey) Mao, Yunfei Zhao, Yuxin Wang, Yuxuan Tang, Zhenheng Tang, Zhifei Li, Ziruo Wang, Ziyu She, Kaiyuan Liu, Iheb Chaabane, Yuxin Tang, Xiangyi Li, Satya Sai Srinath Namburi GNVV, Xinyue Zheng}
\vspace{-1mm}
\titleauthorheading{Advisory Committee}
\titleauthornames{Andy Konwinski, Boxuan Li, Leon Liangyu Chen, Alex Dimakis, Nicholas Carlini, Soroush Vosoughi, Sanmi Koyejo, Di He, Etash Guha, Benjamin Feuer, Mike Merrill\projectleadmark, Ludwig Schmidt\projectleadmark, Alex Shaw\projectleadmark}

\vspace{1mm}
{\fontsize{7.4pt}{8.5pt}\selectfont
\centering
\projectleadsymbol\ Project Leads \enspace $\cdot$ \enspace
\corecontributorsymbol\ Core Contributors \enspace $\cdot$ \enspace
Full affiliations in Appendix~A.\par
Correspondence: \texttt{ls2282@cornell.edu}; \texttt{linhaowei@pku.edu.cn}.\par
\vspace{0.4mm}
Harbor framework and adapters: \url{https://github.com/harbor-framework/harbor}\par
Harbor-Index: \url{https://github.com/harbor-framework/harbor-index}\par}
\end{titleauthorblock}
}

\begin{document}

\maketitle

\vspace{-4mm}
\begin{abstract}
Evaluating agents on the growing number of agentic benchmarks is challenging because they often require complex environments and agent integrations.
We introduce \textbf{Harbor Adapters}, a unified evaluation infrastructure for agentic benchmarks.
Our work makes three contributions.
First, we develop benchmark adapters that port more than 80 benchmarks to evaluate arbitrary agents, and validate them through rigorous code review and parity experiments. 
Second, we conduct a large-scale evaluation of \MainEvalModelCount{} models spanning capability tiers across \MainEvalBenchmarkCount{} benchmarks; every model is run with Terminus-2 and with one of \MainEvalFamilyHarnessCount{} native harnesses. This enables a broader analysis of agent capabilities and failure modes than was previously possible. Third, we introduce \textbf{Harbor-Index}, a curated set of \HarborIndexTaskCount{} difficult, diverse, and high-quality tasks spanning \HarborIndexBenchmarkCount{} benchmarks, refined from the adapted suite through difficulty filtering, AI and human audit, and an audit-and-fix loop. Harbor-Index preserves the challenge and breadth of large-scale agentic evaluations while being affordable to run; no evaluated model--harness configuration exceeds $30\%$ pass rate, and the strongest (\HarborIndexBestConfiguration{}) reaches $\HarborIndexBestPassRate\%$. We release the adapters, evaluation results, in-depth analysis, and Harbor-Index as open-source artifacts to support more reliable and comprehensive evaluation of language-model agents.
\end{abstract}

\begin{figure}[h]
    \centering
    \vspace{-0.8em}
    \includegraphics[width=\linewidth]{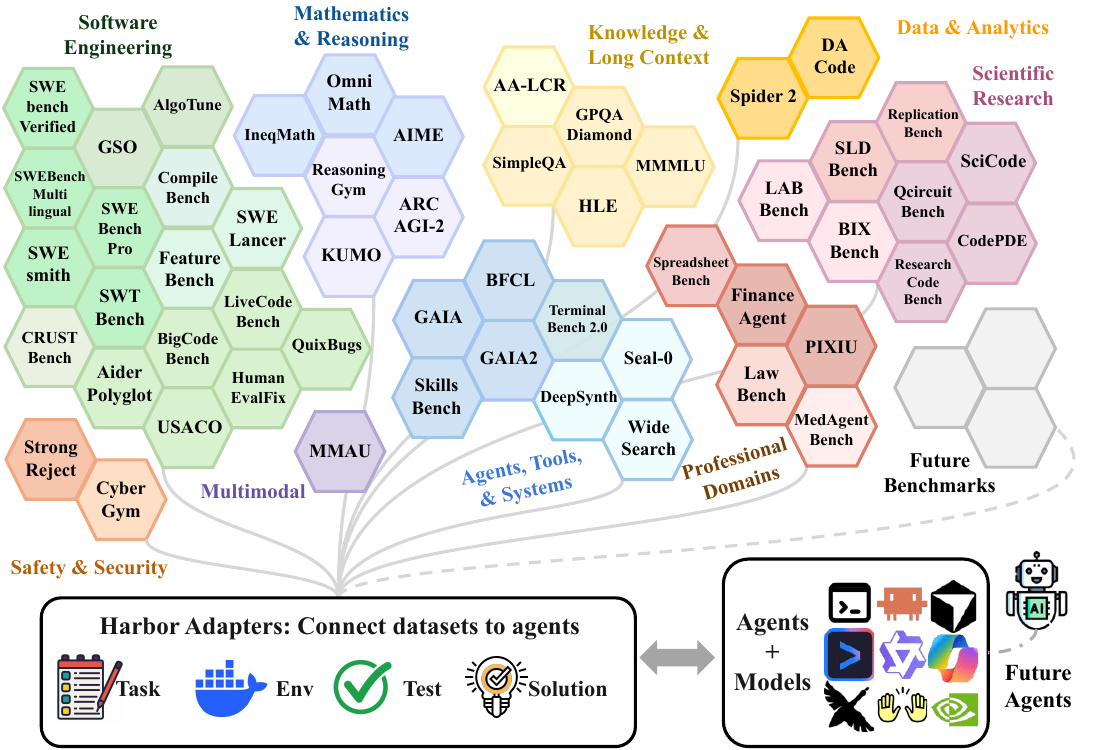}
    \vspace{-0.5em}
    \caption{We integrate and evaluate Harbor adapters for 
    \MainEvalBenchmarkCount{} benchmarks across diverse domains. The Harbor format defines a task by \textit{instruction}, \textit{environment}, \textit{test}, and \textit{solution}, enabling heterogeneous datasets to be standardized to a shared schema and connected to multiple agents.}
    \vspace{-1.1em}
    \label{fig: fig1}
\end{figure}

\vspace{-0.35em}

\section{Introduction}
\label{sec:intro}

\vspace{-0.4em}

Language models are increasingly embedded within agents: systems that not only answer questions, but also plan, use tools, write code, operate computers, and carry out long-horizon tasks.
This shift has expanded the scope of model evaluation beyond static question answering to interactive settings such as software engineering, web research, finance, and terminal-based problem solving.

\needspace{7\baselineskip}
The benchmark landscape has grown in response: SWE-bench~\citep{jimenez2023swebench}, Terminal-Bench~\citep{terminalbench}, FinanceAgent~\citep{valsai2024financeagent}, and many other agentic, tool-using, long-horizon suites; more than 200 have appeared since 2024. This growth reflects the importance of agent evaluation, but has also created a major infrastructure challenge.

Agentic benchmarks are far harder to run than question-answering suites because they are heterogeneous in task formats, environment requirements, interaction protocols, and scoring procedures. Supporting $m$ benchmarks and $n$ agents therefore often requires $\mathcal{O}(mn)$ integrations, one per benchmark-agent pair. This fragmentation limits reliability and scalability: papers and model releases typically report only a small subset of popular benchmarks, so it is unclear whether progress generalizes across domains or concentrates on a few well-known tasks, and comparisons are confounded by differences in benchmark implementations, environment setups, and evaluation protocols.

We address this with \textbf{Harbor Adapters}, which unify diverse agentic evaluations under Harbor, a Python library~\citep{Harbor_Framework} originally developed for Terminal-Bench that defines tasks and runs agents in sandbox environments. Harbor Adapters decouple agents from benchmarks: each benchmark needs one adapter exposing its tasks, environments, and scoring logic through Harbor, and each agent needs one Harbor integration. This changes the integration burden from $\mathcal{O}(mn)$ benchmark-agent pairings to $\mathcal{O}(m + n)$ adapters and agent integrations (\Cref{fig: fig1}). 


Our first contribution is the Harbor Adapters infrastructure. We port more than 80 benchmarks to Harbor, covering both natively agentic benchmarks (e.g., SWE-bench) and non-agentic ones (e.g., HLE~\citep{phan2026hle}) that we make agentic by letting agents use tools, write files, and interact with executable environments. We use rigorous code review and parity experiments to verify that our adaptations remain comparable to the original implementations.

Our second contribution is a large-scale evaluation of models and harnesses. We evaluate \MainEvalModelCount{} models spanning capability tiers, each under exactly two harnesses: cross-family Terminus-2 and one of \MainEvalFamilyHarnessCount{} native harnesses (Codex for GPT, Claude Code for Claude, Gemini CLI for Gemini). This yields \MainEvalConfigurationCount{} model--harness configurations (\MainEvalHarnessImplementationCount{} distinct harness implementations) across \MainEvalBenchmarkCount{} benchmarks and \MainEvalTaskCount{} tasks, three trials per configuration, consuming 226B input and output tokens and more than \$300K of compute. We find:
\begin{itemize}[topsep=0pt, itemsep=2pt, parsep=0pt]
\item The model underlying an agent is more influential for benchmark performance than the harness.
\item The ordering of models on benchmarks is usually highly correlated, i.e., better models are usually better on many benchmarks as opposed to dominating specific niches.
\item Better models use fewer tokens than weaker models across all empirical task difficulty levels.
\item Among unresolved tasks, \emph{task failures} such as misspecified tasks are common, obscuring whether a benchmark still has room for improvement or its remaining tasks are unsolvable because of errors in the task definitions.
\end{itemize}

Our third contribution is \textbf{Harbor-Index}, \HarborIndexTaskCount{} tasks spanning \HarborIndexBenchmarkCount{} benchmarks drawn from the adapted suite. Because agentic evaluation is expensive, Harbor-Index is \emph{affordable} by design: compact, diverse over many agentic domains and task formats, difficult enough to avoid immediate saturation, and high-quality enough that failures reflect model limitations rather than task defects or ambiguous grading. No evaluated model--harness configuration exceeds $30\%$ pass rate; the strongest (\HarborIndexBestConfiguration{}) reaches $\HarborIndexBestPassRate\%$, leaving substantial room for future progress.

We open-source the \href{https://github.com/harbor-framework/harbor/tree/main/adapters}{benchmark adapters}, \href{https://github.com/harbor-framework/harbor-adapters-experiments}{evaluation tools and analyses}, the \href{https://huggingface.co/datasets/kendx/Harbor-Adapter}{full trajectories of the large-scale evaluation}, and Harbor-Index~1.0 on \href{https://github.com/harbor-framework/harbor-index/tree/harbor-index-1.0}{GitHub} and the \href{https://hub.harborframework.com/datasets/harbor-index/harbor-index-1.0}{Harbor Hub}, with interactive results at \href{https://harbor-index.org}{harbor-index.org}.

\vspace{-0.6em}

\section{Harbor Adapters: Infrastructure}
\label{sec:harbor-adapters-infra}

\vspace{-0.3em}






\paragraph{Overview} {Harbor Adapters} is a unifying integration layer built upon \textbf{Harbor}~\cite{Harbor_Framework}.
Harbor provides a standardized task abstraction and execution environment; Harbor Adapters translate dataset-specific formats into Harbor's task schema of \textit{instruction}, \textit{environment}, \textit{tests}, and \textit{solution} (\Cref{fig: fig1}), decoupling agent execution from benchmark-specific logic and reducing integration from $\mathcal{O}(mn)$ to $\mathcal{O}(m+n)$. In addition, the adapter infrastructure supports GPU interaction, LLM-as-a-judge verifiers, custom metric aggregation, and multiple sandbox backends (e.g., Daytona~\citep{daytona}, Modal~\citep{modal}, and E2B~\citep{e2b}). As of May 2026, Harbor Adapters support 22 agents and more than 80 benchmarks.

\paragraph{Adapter construction}
Adapters are designed to faithfully preserve the semantics of the original benchmark datasets. To verify adaptation fairness, (1) we run multi-trial \textbf{parity experiments} matching agent, model, and execution configuration across original and Harbor-adapted benchmarks (Appendix~\ref{app:parity_experiments}), reporting mean $\pm$ sample standard error of the mean (SEM) to account for stochasticity in \Cref{tab:adapter_catalog_part1,tab:adapter_catalog_part2}; (2) each adapter undergoes a strict three-stage quality audit by \emph{bot, junior, and senior reviewers} to ensure faithful and standardized adaptation, with over 10,000 GitHub comments across all adapters (Appendix~\ref{app:adapter_construction_workflow:review}).

\paragraph{Large-scale evaluation.}
Harbor Adapters enable a large-scale evaluation over \MainEvalTaskCount{} tasks from \MainEvalBenchmarkCount{} benchmarks, analyzed in Section \ref{sec:analysis}. We cover \MainEvalModelCount{} models spanning capability tiers from Google, OpenAI, and Anthropic: Gemini-3.1-Pro and Gemini-3-Flash; Claude Opus 4.6, Sonnet 4.6, and Haiku 4.5; and GPT-5.4, GPT-5-mini, and -nano. Each model runs under exactly two harnesses: the cross-family \textit{Terminus-2}~\citep{terminalbench} and one of \MainEvalFamilyHarnessCount{} native harnesses (Gemini CLI for Gemini, Claude Code for Claude, and Codex for GPT), giving \MainEvalConfigurationCount{} model--harness configurations using \MainEvalHarnessImplementationCount{} distinct harness implementations overall. Each (benchmark, model, harness) setting is repeated for 3 trials, yielding $\sim$0.3M trajectories. This scale is difficult to reach with $\mathcal O(mn)$ benchmark-specific integrations; Harbor Adapters make it feasible. Further experimental details are in Appendix~\ref{app:evaluation_details}.

\section{Analysis: Agentic Benchmarking at Scale}
\label{sec:analysis}\
\providecommand{\NumIncludedBenchmarks}{54}
\providecommand{\NumModels}{8}
\providecommand{\NumHarnesses}{4}

\providecommand{\AnthropicModelEffectMean}{1.441}
\providecommand{\AnthropicHarnessEffectMean}{0.457}
\providecommand{\AnthropicModelHarnessRatio}{3.2}
\providecommand{\AnthropicModelWins}{49}
\providecommand{\AnthropicTotalBenchmarks}{54}
\providecommand{\AnthropicModelWinPct}{91}
\providecommand{\GoogleModelEffectMean}{0.481}
\providecommand{\GoogleHarnessEffectMean}{0.530}
\providecommand{\GoogleModelHarnessRatio}{0.9}
\providecommand{\GoogleModelWins}{31}
\providecommand{\GoogleTotalBenchmarks}{54}
\providecommand{\GoogleModelWinPct}{57}
\providecommand{\OpenAIModelEffectMean}{1.880}
\providecommand{\OpenAIHarnessEffectMean}{1.229}
\providecommand{\OpenAIModelHarnessRatio}{1.5}
\providecommand{\OpenAIModelWins}{43}
\providecommand{\OpenAITotalBenchmarks}{54}
\providecommand{\OpenAIModelWinPct}{80}

\providecommand{\GLMMIcc}{0.75}
\providecommand{\GLMMBenchmarkVar}{0.0485}
\providecommand{\GLMMResidualVar}{0.0161}
\providecommand{\GLMMNumSigModels}{6}
\providecommand{\GLMMMaxModelEffect}{0.235}
\providecommand{\GLMMMinModelEffect}{-0.216}
\providecommand{\GLMMNumSigHarnesses}{2}
\providecommand{\GLMMMaxHarnessEffect}{0.045}
\providecommand{\GLMMMinHarnessEffect}{-0.042}
\providecommand{\GLMMModelEffectRange}{0.451}
\providecommand{\GLMMHarnessEffectRange}{0.087}
\providecommand{\GLMMModelHarnessRangeRatio}{5.2}

\providecommand{\claudecodeMeanDelta}{0.239}
\providecommand{\claudecodeWinRate}{65}
\providecommand{\codexMeanDelta}{0.007}
\providecommand{\codexWinRate}{70}
\providecommand{\geminicliMeanDelta}{-0.092}
\providecommand{\geminicliWinRate}{53}

\providecommand{\TerminusgptFiveFourcodexDelta}{0.883}
\providecommand{\TerminusgptFiveFourcodexWinRate}{94}
\providecommand{\TerminusclaudehaikuFourFiveTwoZeroTwoFiveOneZeroZeroOneclaudecodeDelta}{0.653}
\providecommand{\TerminusclaudehaikuFourFiveTwoZeroTwoFiveOneZeroZeroOneclaudecodeWinRate}{83}
\providecommand{\TerminusgptFiveminicodexDelta}{0.547}
\providecommand{\TerminusgptFiveminicodexWinRate}{76}
\providecommand{\TerminusclaudesonnetFourSixclaudecodeDelta}{0.067}
\providecommand{\TerminusclaudesonnetFourSixclaudecodeWinRate}{65}
\providecommand{\TerminusgeminiThreeOnepropreviewgeminicliDelta}{0.028}
\providecommand{\TerminusgeminiThreeOnepropreviewgeminicliWinRate}{56}
\providecommand{\TerminusclaudeopusFourSixclaudecodeDelta}{-0.003}
\providecommand{\TerminusclaudeopusFourSixclaudecodeWinRate}{46}
\providecommand{\TerminusgeminiThreeflashpreviewgeminicliDelta}{-0.213}
\providecommand{\TerminusgeminiThreeflashpreviewgeminicliWinRate}{50}
\providecommand{\TerminusgptFivenanocodexDelta}{-1.408}
\providecommand{\TerminusgptFivenanocodexWinRate}{41}

\providecommand{\GPTFiveFourBase}{.513}
\providecommand{\GPTFiveFourDelta}{$+$.179}
\providecommand{\GPTFiveFourLift}{94}
\providecommand{\ClaudeHaikuBase}{.405}
\providecommand{\ClaudeHaikuDelta}{$+$.092}
\providecommand{\ClaudeHaikuLift}{83}
\providecommand{\GPTMiniBase}{.416}
\providecommand{\GPTMiniDelta}{$+$.111}
\providecommand{\GPTMiniLift}{76}
\providecommand{\ClaudeSonnetBase}{.581}
\providecommand{\ClaudeSonnetDelta}{$+$.022}
\providecommand{\ClaudeSonnetLift}{65}
\providecommand{\GeminiProBase}{.656}
\providecommand{\GeminiProDelta}{$+$.024}
\providecommand{\GeminiProLift}{56}
\providecommand{\ClaudeOpusBase}{.624}
\providecommand{\ClaudeOpusDelta}{$+$.012}
\providecommand{\ClaudeOpusLift}{46}
\providecommand{\GeminiFlashBase}{.581}
\providecommand{\GeminiFlashDelta}{$-$.013}
\providecommand{\GeminiFlashLift}{50}
\providecommand{\GPTNanoBase}{.272}
\providecommand{\GPTNanoDelta}{$-$.029}
\providecommand{\GPTNanoLift}{41}

\providecommand{\NumSaturatedBenchmarks}{13}
\providecommand{\NumHardBenchmarks}{7}
\providecommand{\MeanBestScore}{0.737}

\providecommand{\NumAgentsToolsAndSystemsSaturated}{0}
\providecommand{\NumAgentsToolsAndSystemsHard}{1}
\providecommand{\NumAgentsToolsAndSystemsBenchmarks}{9}
\providecommand{\MeanAgentsToolsAndSystemsBestScore}{67}
\providecommand{\NumDataAndAnalyticsSaturated}{0}
\providecommand{\NumDataAndAnalyticsHard}{1}
\providecommand{\NumDataAndAnalyticsBenchmarks}{2}
\providecommand{\MeanDataAndAnalyticsBestScore}{51}
\providecommand{\NumKnowledgeAndLongContextSaturated}{2}
\providecommand{\NumKnowledgeAndLongContextHard}{0}
\providecommand{\NumKnowledgeAndLongContextBenchmarks}{4}
\providecommand{\MeanKnowledgeAndLongContextBestScore}{83}
\providecommand{\NumMathematicsAndReasoningSaturated}{3}
\providecommand{\NumMathematicsAndReasoningHard}{0}
\providecommand{\NumMathematicsAndReasoningBenchmarks}{6}
\providecommand{\MeanMathematicsAndReasoningBestScore}{92}
\providecommand{\NumMultimodalSaturated}{0}
\providecommand{\NumMultimodalHard}{0}
\providecommand{\NumMultimodalBenchmarks}{1}
\providecommand{\MeanMultimodalBestScore}{70}
\providecommand{\NumProfessionalDomainsSaturated}{0}
\providecommand{\NumProfessionalDomainsHard}{0}
\providecommand{\NumProfessionalDomainsBenchmarks}{5}
\providecommand{\MeanProfessionalDomainsBestScore}{75}
\providecommand{\NumSafetyAndSecuritySaturated}{1}
\providecommand{\NumSafetyAndSecurityHard}{0}
\providecommand{\NumSafetyAndSecurityBenchmarks}{1}
\providecommand{\MeanSafetyAndSecurityBestScore}{97}
\providecommand{\NumScientificResearchSaturated}{0}
\providecommand{\NumScientificResearchHard}{1}
\providecommand{\NumScientificResearchBenchmarks}{8}
\providecommand{\MeanScientificResearchBestScore}{67}
\providecommand{\NumSoftwareEngineeringSaturated}{5}
\providecommand{\NumSoftwareEngineeringHard}{4}
\providecommand{\NumSoftwareEngineeringBenchmarks}{18}
\providecommand{\MeanSoftwareEngineeringBestScore}{73}
\providecommand{\MostHeadroomDomainOne}{Data \& Analytics}
\providecommand{\MostHeadroomDomainOneScore}{51}
\providecommand{\MostHeadroomDomainTwo}{Agents, Tools \& Systems}
\providecommand{\MostHeadroomDomainTwoScore}{67}
\providecommand{\MostHeadroomDomainThree}{Scientific Research}
\providecommand{\MostHeadroomDomainThreeScore}{67}
\providecommand{\SEHardBenchmarkOne}{AlgoTune}
\providecommand{\SEHardBenchmarkOneScore}{28}
\providecommand{\SEHardBenchmarkTwo}{FeatureBench}
\providecommand{\SEHardBenchmarkTwoScore}{41}
\providecommand{\SEHardBenchmarkThree}{BigCodeBench}
\providecommand{\SEHardBenchmarkThreeScore}{45}
\providecommand{\SEHardBenchmarkFour}{GSO}
\providecommand{\SEHardBenchmarkFourScore}{45}

\providecommand{\ParticipationRatio}{1.3}
\providecommand{\ComponentsForNinety}{2}
\providecommand{\ComponentsForNinetyFive}{3}
\providecommand{\TopOneVariance}{89}
\providecommand{\TopThreeVariance}{96}

\providecommand{\WithinDomainMedianCorr}{0.70}
\providecommand{\CrossDomainMedianCorr}{0.65}
\providecommand{\WithinDomainFracAboveSeven}{50}
\providecommand{\CrossDomainFracAboveSeven}{41}
\providecommand{\WithinAgentsToolsAndSystemsMedianCorr}{0.67}
\providecommand{\WithinKnowledgeAndLongContextMedianCorr}{0.78}
\providecommand{\WithinMathematicsAndReasoningMedianCorr}{0.59}
\providecommand{\WithinProfessionalDomainsMedianCorr}{0.72}
\providecommand{\WithinScientificResearchMedianCorr}{0.48}
\providecommand{\WithinSoftwareEngineeringMedianCorr}{0.73}

\providecommand{\GreedyBenchmarksBelowSeven}{12}
\providecommand{\GreedyFirstAboveSeven}{swebench-verified}
\providecommand{\GreedyFirstAboveSevenStep}{13}

\providecommand{\AimeGptFourCodexBase}{0.711}
\providecommand{\AimeGptFourCodexAgent}{0.989}
\providecommand{\AimeGptFourCodexDeltaPP}{27.8}
\providecommand{\AimeGptMiniCodexBase}{0.889}
\providecommand{\AimeGptMiniCodexAgent}{0.967}
\providecommand{\AimeGptMiniCodexDeltaPP}{7.8}
\providecommand{\AimeGptNanoCodexBase}{0.756}
\providecommand{\AimeGptNanoCodexAgent}{0.739}
\providecommand{\AimeGptNanoCodexDeltaPP}{-1.7}

\providecommand{\BenchPressRedundantOne}{HumanEvalFix}
\providecommand{\BenchPressRedundantOneScore}{0.004}
\providecommand{\BenchPressUniqueOne}{ResearchCodeBench}
\providecommand{\BenchPressUniqueOneScore}{0.221}
\providecommand{\BenchPressRedundantTwo}{KUMO}
\providecommand{\BenchPressRedundantTwoScore}{0.017}
\providecommand{\BenchPressUniqueTwo}{FinanceAgent}
\providecommand{\BenchPressUniqueTwoScore}{0.202}
\providecommand{\BenchPressRedundantThree}{BFCL}
\providecommand{\BenchPressRedundantThreeScore}{0.029}
\providecommand{\BenchPressUniqueThree}{LAB-Bench}
\providecommand{\BenchPressUniqueThreeScore}{0.168}
\providecommand{\DiffMedAPERho}{-.29}
\providecommand{\DiffMedAPEPval}{.03}
\providecommand{\DiffMedAPEN}{54}

\providecommand{\GreedyBenchmarksBelowSeven}{12}
\providecommand{\GreedyFirstAboveSeven}{SWE-bench-verified}
\providecommand{\GreedyFirstAboveSevenStep}{13}
\providecommand{\AppNumSystems}{16}
\providecommand{\AppSVDModelOnlyPCOne}{73.7}
\providecommand{\AppSVDFullPCTwo}{10.1}
\providecommand{\TaskRedundNumBenchmarks}{52}
\providecommand{\TaskRedundMedianPCsNinety}{8}
\providecommand{\TaskRedundRhoKThree}{0.923}


\definecolor{LiftGreen}{RGB}{226,248,230}
\definecolor{LiftPink}{RGB}{255,228,232}

\begin{figure}[ht]
  \centering
\includegraphics[width=\linewidth]{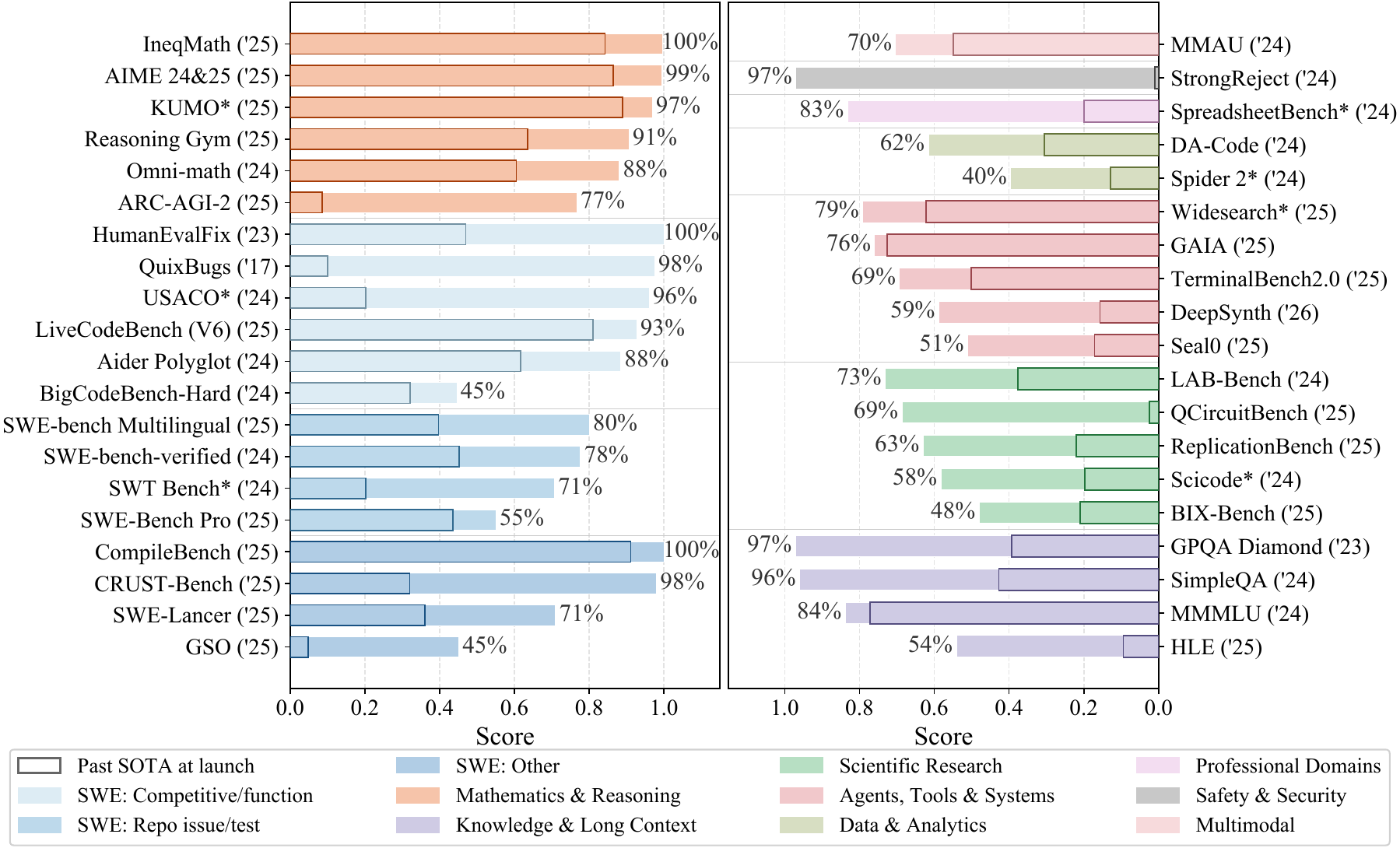}
  \vspace{-1em}
  \caption{\textbf{Benchmark progress since published}; best reported score per benchmark, colored
by domain, for 39 of the \NumIncludedBenchmarks{} benchmarks we evaluated. ``SWE'' abbreviates
software engineering. Scores are normalized to $[0,1]$ so that higher is better.
Light-filled bars indicate the current best score in our evaluation; outlined bars, the state of the
art at publication time. Benchmarks marked $^*$ use an i.i.d. subset of the full dataset. Launch and
current-best scores come from benchmark papers, leaderboards, and model reports
(Appendix~\ref{app:benchmark-data}).}
  \label{fig:headroom-by-domain}
\end{figure}

We ask four questions:
(1) \textbf{Benchmark:} how much unique signal does the \NumIncludedBenchmarks{}-benchmark suite provide?
(2) \textbf{Harness:} does model capability or harness design drive performance?
(3) \textbf{Efficiency:} where do gains justify token costs?
(4) \textbf{Bottlenecks:} where do tasks break, and why do agents fail?

\subsection{Benchmark: progress over time and redundancy analysis}
\paragraph{Which benchmarks still challenge frontier models?}

\Cref{fig:headroom-by-domain} shows broad progress across domains. Of the
\NumIncludedBenchmarks{} benchmarks we evaluate, \NumSaturatedBenchmarks{} are largely saturated,
with state-of-the-art models exceeding 90\%, including once-challenging ones (e.g.,
GPQA-Diamond \cite{rein2024gpqa}). Software engineering is especially prominent and splits sharply: functional and
competitive-coding tasks near saturation (e.g., LiveCodeBench \cite{jain2024livecodebench}),
repository-level feature implementation and issue fixing progress quickly (e.g., SWE-bench-verified
\cite{jimenez2023swebench}), while software speedup remains hard (e.g., GSO
\cite{shetty2025gso}).

\paragraph{How many benchmarks do we actually need to differentiate models?}
While benchmark tasks can rank models by capability, we find the effective evaluation space far
lower-dimensional than the number of benchmarks and tasks suggests, in three complementary
forms.


\textbf{(1) At the benchmark level, most scores vary along only a few shared directions.}  
Following BenchPress~\cite{papailiopoulos2026benchpress}, we apply Principal Component Analysis
(PCA) to the $\AppNumSystems{} \times \NumIncludedBenchmarks{}$ score matrix of model--harness
configurations $\times$ benchmarks (\Cref{tab:gpt-performance-full},
\ref{tab:claude-performance-full}, \& \ref{tab:gemini-performance-full}). Most variation is
explained by a \emph{single} shared capability factor; adding agent scaffolds introduces one
additional direction, but the space remains low-rank (Appendix~\ref{app:benchpress}). Greedy selection on absolute Spearman rank correlations agrees: after
\GreedyBenchmarksBelowSeven{} benchmarks are selected, every remaining benchmark correlates at
$\rho \ge 0.7$ with a selected one, so its model ranking is already well approximated and adds
little information (\Cref{tab:greedy-sequence}).

\textbf{(2) Difficulty and uniqueness are not the same.} The benchmarks that greedy selection identifies as independently informative span the full difficulty spectrum, from hard (CodePDE~\cite{li2025codepde}) to easy
(StrongReject~\cite{souly2024strongreject}), rather than clustering at one end
(\Cref{tab:greedy-sequence}; Appendix~\ref{app:benchpress}). Some difficult benchmarks (e.g., BigCodeBench~\cite{zhuo2025bigcodebench}) are largely
predictable from others, while benchmarks testing specialized capabilities resist prediction
regardless of difficulty (e.g., FinanceAgent~\cite{valsai2024financeagent} and
LabBench~\cite{laurent2024labbench}). This suggests that evaluation design should prioritize
coverage of distinct capability axes, not raw difficulty alone.

\textbf{(3) Redundancy also appears within individual benchmarks.}
Across \TaskRedundNumBenchmarks{} benchmarks with at least 10 tasks
(excluding CodePDE~\cite{li2025codepde} (5 tasks) and SLDBench~\cite{lin2025sldbench} (8 tasks)),
3 representative tasks per benchmark already recover the system ranking at a mean
$\rho \approx \TaskRedundRhoKThree{}$, so many tasks within a benchmark measure overlapping
capabilities. The degree of redundancy varies with internal task diversity: 3 tasks nearly reproduce
the full ranking where tasks test similar capabilities (e.g., WideSearch, $\rho = 0.99$), but not
where task types are diverse (e.g., CyberGym~\cite{cybergym2025}, $\rho = 0.75$)
(Appendix~\ref{app:benchpress}).

\subsection{Harness: the interaction analysis between harness and base model}
\label{sec:agent-harness}

\paragraph{Which matters more: the model or the harness?}
Benchmark difficulty accounts for most score variance (ICC $=\GLMMIcc{}$), so we control for it
with a linear mixed model using benchmark as a random intercept:
\GLMMNumSigModels{}/\NumModels{} model coefficients are significant ($p<0.05$) versus
\GLMMNumSigHarnesses{}/\NumHarnesses{} harness coefficients, and the model fixed-effect range
(\GLMMModelEffectRange{}) is \GLMMModelHarnessRangeRatio{}$\times$ the harness range
(\GLMMHarnessEffectRange{}) (Appendix~\ref{app:model-vs-agent}). Thus harness design can affect
performance, but it does not substitute for base-model capability.

\subsection{Efficiency: the tradeoff analysis between performance and cost}

\label{sec:main_token}


\definecolor{frontiergreen}{RGB}{226,236,231}
\definecolor{otherorange}{RGB}{249,235,224}

\begin{figure}[h]
    \centering
    \vspace{-1em}
\begin{minipage}[c]{0.42\linewidth}
    \scriptsize
    \setlength{\tabcolsep}{4pt}
    \renewcommand{\arraystretch}{1.1}
    \begin{tabular}{l r >{\centering\arraybackslash}p{0.9cm} >{\centering\arraybackslash}p{0.9cm} c}
        \toprule
        \textbf{Difficulty} & \textbf{\# Tasks} & \textcolor{brown}{\textbf{Other}} & \textcolor{teal}{\textbf{Frontier}} & $\boldsymbol{\Delta}$ \\
        \midrule
        0.0--0.1 & 1{,}461 & \cellcolor{otherorange}94\% & \cellcolor{frontiergreen}99\% & +5 pp  \\
        0.1--0.2 & 862     & \cellcolor{otherorange}80\% & \cellcolor{frontiergreen}95\% & +15 pp \\
        0.2--0.3 & 660     & \cellcolor{otherorange}67\% & \cellcolor{frontiergreen}90\% & +23 pp \\
        0.3--0.4 & 495     & \cellcolor{otherorange}54\% & \cellcolor{frontiergreen}83\% & +29 pp \\
        0.4--0.5 & 429     & \cellcolor{otherorange}44\% & \cellcolor{frontiergreen}71\% & +27 pp \\
        0.5--0.6 & 392     & \cellcolor{otherorange}35\% & \cellcolor{frontiergreen}62\% & +27 pp \\
        0.6--0.7 & 448     & \cellcolor{otherorange}26\% & \cellcolor{frontiergreen}51\% & +25 pp \\
        0.7--0.8 & 460     & \cellcolor{otherorange}17\% & \cellcolor{frontiergreen}39\% & +22 pp \\
        0.8--0.9 & 454     & \cellcolor{otherorange}9\%  & \cellcolor{frontiergreen}25\% & +16 pp \\
        0.9--1.0 & 966     & \cellcolor{otherorange}1\%  & \cellcolor{frontiergreen}4\%  & +3 pp  \\
        \bottomrule
    \end{tabular}
\end{minipage}
    \begin{minipage}[c]{0.57\linewidth}
        \centering
        \includegraphics[width=\linewidth]{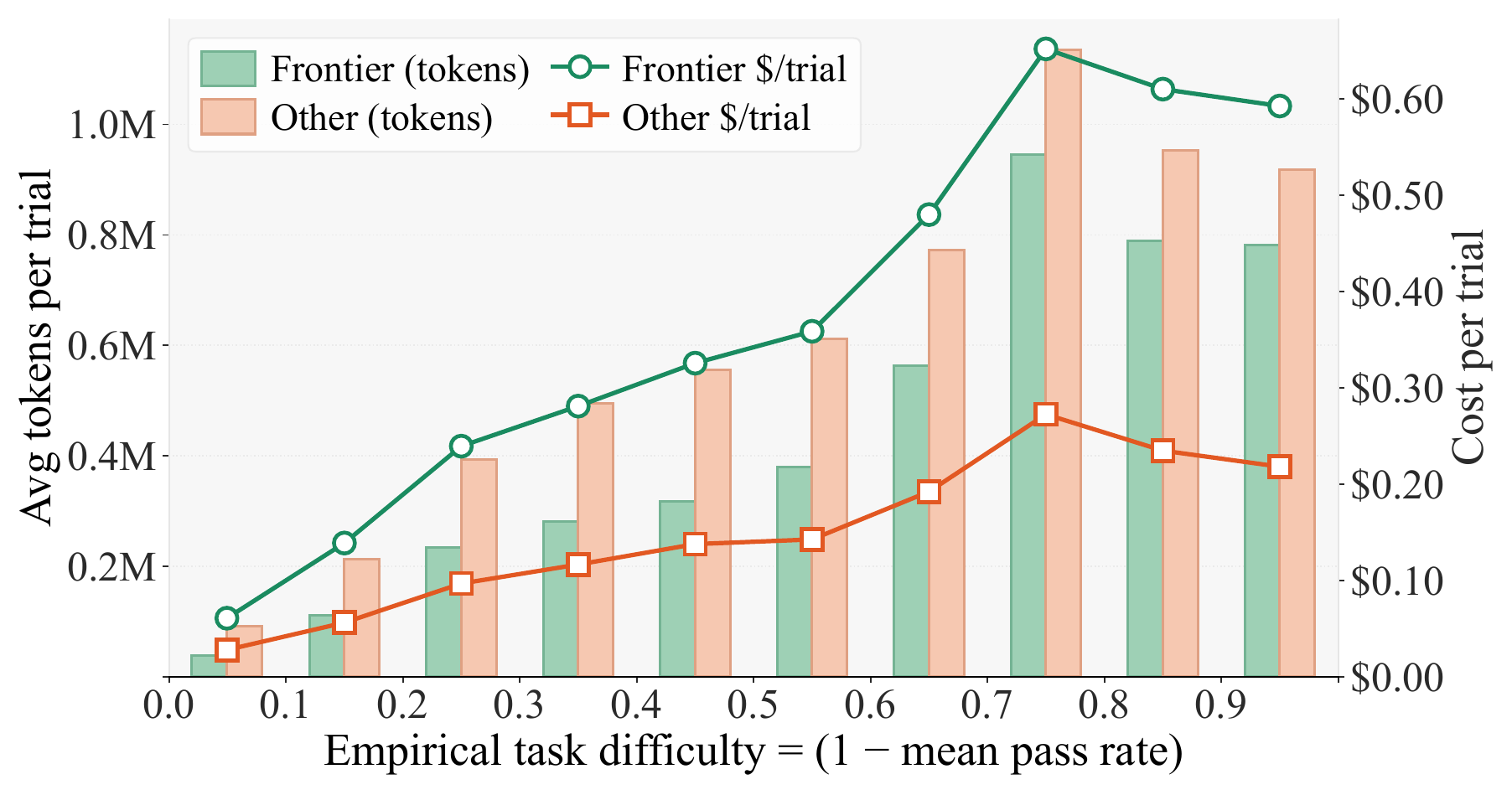}
    \end{minipage}
    \vspace{-1em}
    \caption{\textbf{Left:} Benchmark scores across empirical task-difficulty buckets for three frontier models (Claude Opus 4.6, Gemini 3.1 Pro Preview, GPT 5.4) versus five other models (Claude Sonnet 4.6, Claude Haiku 4.5, Gemini 3.1 Flash Preview, GPT 5 mini, GPT 5 nano).
    \textbf{Right:} Average tokens per trial (bars) and dollar cost per trial (lines) by empirical task difficulty under our experiment setup.
    }
    \label{fig:difficulty_token_table}
\end{figure}

Frontier models are more capable but more expensive. We therefore ask: (1) where does upgrading to a frontier model pay off most, (2) do frontier models solve tasks with shorter trajectories, and (3) does that token efficiency offset their higher prices? 

We define \emph{empirical task difficulty} as one minus the average score across all model--harness runs, and group tasks into ten buckets of width 0.1. 
The distribution is U-shaped: 22.0\% of tasks fall in the easiest bucket (0--0.1), 14.6\% in the hardest (0.9--1), and the remaining 63.4\% across the eight middle buckets. 
Note that this is a configuration-dependent proxy for intrinsic task difficulty, as it may be distorted by task brittleness (false positives/negatives) and the evaluation setup itself.

\textbf{(1) Marginal benefits.}
As shown in \Cref{fig:difficulty_token_table} (left), the absolute performance gain of frontier models is bell-shaped: $\Delta$ is marginal on the simplest and hardest tasks and peaks in the medium band (0.3–0.7). Other models use more tokens at every difficulty level yet cost 2--3$\times$ less per trial (\Cref{fig:difficulty_token_table}, right). The frontier value proposition is therefore most pronounced on medium-to-hard tasks, where the gain in absolute performance justifies the higher cost.


\textbf{(2) Frontier models are more token-efficient, especially on easy tasks.}
Frontier models use fewer tokens than weaker models at every difficulty tier (Figure~\ref{fig:difficulty_token_table}, right), and the \emph{relative} gap is widest on easy tasks: in the easiest bucket (0--0.1), frontier models use only 42\% of weaker-model tokens, a ratio rising to 61--86\% in the upper half of the distribution ($\ge 0.5$). These results suggest that frontier models produce more concise trajectories overall, but that their relative token-efficiency advantage is most pronounced on easy tasks.

Manual inspection of agent trajectories further reveals that, on easy tasks, weaker models consume significantly more tokens, turns, and tokens per turn to reach the correct solution 
(e.g., 6.14$\times$ more tokens on SimpleQA \cite{wei2024simpleqa} and 2.28$\times$ on KUMO \cite{lin2025kumo}).
The extra turns typically come from instruction misunderstanding and repetitive tool calls; the higher per-turn usage comes from verbose chain-of-thought and redundant verification compensating for weak reasoning. Frontier models instead follow instructions, execute cleanly, and accept outputs without extra deliberation.



\textbf{(3) Token savings do not offset higher prices.}
Higher per-token prices outweigh the token savings: other models cost 2--3$\times$ less per trial across difficulty levels, and the absolute cost gap peaks at \$0.37 per trial in the 0.7--0.8 bucket. Token efficiency therefore reduces, but does not eliminate, the frontier cost premium.

\subsection{Bottlenecks: failure mode analysis}
\label{sec:analysis-quali}
To see where and why agents fail, we audit trajectories where frontier models fail on hard tasks. Two domain-experienced human annotators independently label 200 trajectories across 12 failure modes (pooled inter-rater $\kappa = 0.66$); a calibrated Gemini 3.1 Pro judge ($\kappa = 0.51$ vs gold) extends this to $N=6{,}028$ trajectories across 45 benchmarks, three frontier models, and two harnesses each. Full taxonomy, annotation protocol, and per-rubric reliability are in Appendix~\ref{app:qual-method}.

\begin{figure}[!h]
    \centering
    \makebox[\textwidth][c]{\includegraphics[width=0.9\textwidth]{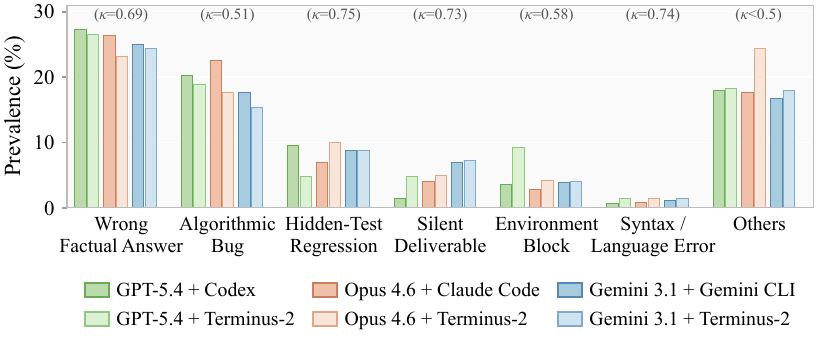}}
    \vspace{-1.9em}
    \caption{Agent failure-mode prevalence on $N=6{,}028$ trajectories across seven rubrics. Each rubric shows six bars grouped by model: the darker bar is the native harness (Codex, Claude Code, Gemini CLI) and the lighter bar is the same model on Terminus-2. Per-rubric inter-rater $\kappa$ is annotated above each group; \textit{Others} collapses six rubrics with $\kappa<0.5$.}
    \label{fig:qual-failure-mode-bars}
    \vspace{-0.5em}
\end{figure}

\textbf{Dominant failure modes.} In \Cref{fig:qual-failure-mode-bars}, \textit{wrong factual answers}, \textit{algorithmic bugs}, and \textit{hidden-test regressions} dominate across all frontier models regardless of harness, indicating a persistent ``capability gap'' in task comprehension, domain knowledge, and the reasoning needed to produce substantially correct solutions. All evaluated agents also still occasionally make trivial operational mistakes such as \textit{syntax errors} or \textit{missing deliverables}.

\textbf{Harness design shapes model behavior.} Our human trajectory analysis (\Cref{app:qual-method}) shows that native harnesses like Claude Code and Codex allow open-ended iteration and self-correction, whereas Terminus-2 enforces a linear \textit{Plan-Execute-Complete} sequence; this lack of a built-in verification-correction loop makes Terminus-2 more vulnerable when iterative refinement is required.

\textbf{Models diverge in behavioral patterns.} GPT-5.4 prioritizes conciseness, either delivering a correct solution quickly or ``giving up'' on difficult tasks (e.g., {HLE}, {LabBench}), more so under Terminus-2 than Codex. Claude models usually hold a steady turn count and consistent execution pattern regardless of task difficulty. Gemini relies heavily on external search, which, while useful, often causes ``information drift'': it replaces an initially correct hypothesis with conflicting online information, notably on knowledge-intensive benchmarks ({LabBench}, {MMMLU}). Full per-(agent, model) breakdowns and examples are in Appendix~\ref{app:qual-results}.

\section{Harbor-Index: Compact, Diverse, Challenging, and High-Quality}
\label{sec:harbor-index}
From our task inspection, \textbf{low resolution rates in existing benchmarks are often attributable to structural design flaws (e.g., instruction-verification mismatches) rather than genuine task complexity}. We therefore introduce \textbf{Harbor-Index}, a curated meta-dataset of \HarborIndexTaskCount{} tasks for next-generation agent development and evaluation. Harbor-Index prioritizes \emph{compactness}, \emph{diversity}, \emph{difficulty}, and \emph{quality}, and is built by passing the entire Harbor trial pool through a multi-stage funnel---difficulty filtering, a series of \emph{AI audit} and \emph{human review}, then an iterative \emph{audit-and-fix} loop---in which each stage rejects tasks that look hard but are structurally broken: a task enters Harbor-Index only if it is well-defined, genuinely difficult, and free of exploitable verifier loopholes.

\subsection{Construction Pipeline}
\label{sec:harbor_index_pipeline}

\begin{figure}[t]
  \centering
  \includegraphics[width=\linewidth]{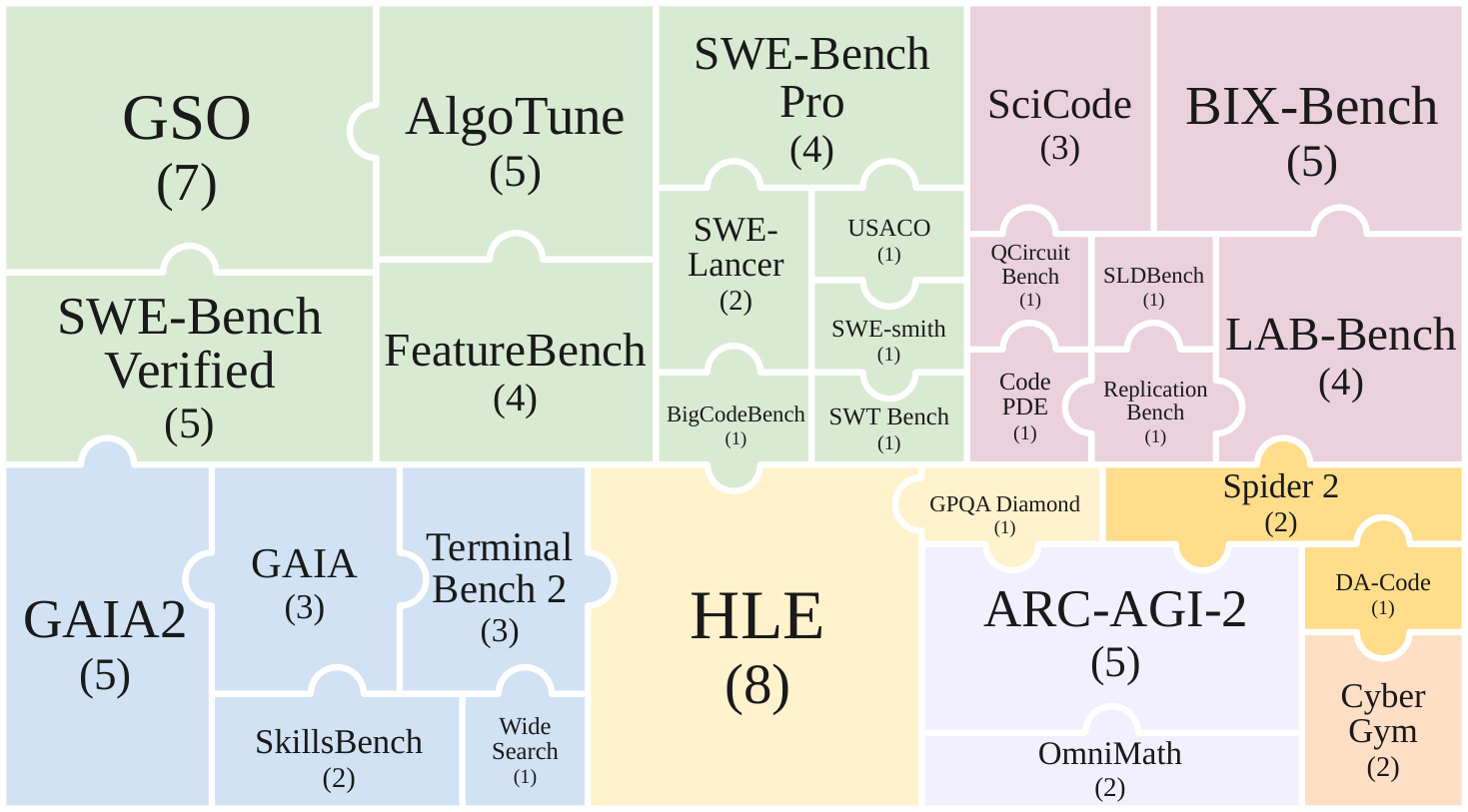}
  \caption{Harbor-Index task distribution by benchmark (\HarborIndexTaskCount{} tasks across \HarborIndexBenchmarkCount{} benchmarks), sized by task count and colored by domain.}
  \label{fig:harbor-index-distribution}
\end{figure}

Starting from an initial pool of \MainEvalTaskCount{} Harbor tasks across \MainEvalBenchmarkCount{} adapters, we apply a difficulty filter retaining the 1{,}311 tasks where three leading models at the time of filtering (Claude Opus 4.6, GPT-5.4, Gemini 3.1 Pro), each under both its native harness and Terminus-2 over three repeats (18 trials), succeed on at most $33\%$ of trials. A Gemini-3-Flash auditor then scores each candidate against a quality rubric based on \textit{instruction--verification alignment} and \textit{essential difficulty} (difficulty must come from genuine reasoning, algorithmic thinking, domain expertise, long-horizon interactions, multi-step execution, etc.), filtering down to 307 tasks. Next, 14 domain-experienced human reviewers re-audit the survivors under the same rubric, each candidate receiving at least one senior or two junior reviews, leaving more than 110; a three-member senior panel then selects 100 on difficulty, diversity, and insight. Finally, at least two senior reviewers examine each of the 100 using trajectory-grounded failure analysis and false-positive/false-negative analysis, repairing broken tasks and dropping those that remain broken or become too easy after repair, yielding the final release of \textbf{\HarborIndexTaskCount{} tasks spanning \HarborIndexBenchmarkCount{} benchmarks}. \Cref{fig:harbor-index-distribution} shows the task distribution, with numbers denoting task counts per benchmark. The full audit process, quality bar, and sampling protocol are in Appendix~\ref{sec:harbor_index_appendix}.

Our human audit reveals that broken task design is common: \textbf{across more than 30 benchmarks, roughly one third of the hardest candidate tasks are rejected as broken rather than genuinely difficult.} On all four GAIA2 \cite{froger2026gaia2} tasks reaching human review, the adapter never fires the simulation events the verifier waits on, so those trials time out regardless of agent approach. A SWE-bench Pro \cite{deng2025swebenchproaiagents} task shows verifier overreach: it asserts \texttt{data-testid} strings and log message formats the instruction never specifies, so a correct solution fails on clerical grounds. On CRustBench \cite{khatry2025crustbench}, the verifier pins transpilation to a fixed hand-written Rust interface, so a correct and memory-safe solution is rejected when signatures or ownership annotations differ from the reference.

\subsection{Evaluation Settings}
\label{sec:evaluation-settings}

We evaluate \HarborIndexModelCount{} models spanning capability tiers and both release types: four \emph{closed-weight} models, whose weights are not publicly released (GPT-5.5, Claude Opus 4.8, Gemini 3.1 Pro, and Qwen3.7 Max), and five \emph{open-weight} models (GLM 5.2, Kimi K2.6, MiniMax M3, DeepSeek V4 Pro, and MiMo V2.5 Pro). GPT-5.5, Claude Opus 4.8, and Gemini 3.1 Pro are served by first-party APIs; the other six models are served through OpenRouter~\citep{openrouter}. All models run under default temperature, context length, and reasoning effort. Each model runs under two harness conditions: Terminus-2 and a model-native harness. This yields $\HarborIndexModelCount\times 2\times \HarborIndexTaskCount = \HarborIndexRolloutCount$ rollouts on Harbor-Index~1.0; the model list and results appear in \Cref{tab:harbor-index-results}. Free-form answer tasks (e.g., HLE, GAIA, GPQA Diamond) use LLM-as-a-judge evaluation; elsewhere scoring is programmatic (unit tests, exact match, thresholded continuous metrics). For continuous metrics such as speedup ratios in AlgoTune \cite{press2025algotune} and GSO \cite{shetty2025gso}, we ``SOTA-threshold'' outcomes into pass/fail, so these tasks only reward solutions that advance the current frontier. See Appendix~\ref{app:evaluation_details} for implementation details.

\subsection{Evaluation Outcome}
\label{sec:evaluation-outcome}

\begin{table}[t]
\centering
\caption{Results on Harbor-Index~1.0 (\HarborIndexTaskCount{} tasks): pass rate (\%) and approximate per-run cost (USD) for a full Index run. The \emph{Native Harness} column group uses the vendor-native harness---Codex, Claude Code, or Gemini CLI---for GPT, Claude, and Gemini, and Claude Code for the six models served through OpenRouter. Models marked $^\dagger$ are closed-weight (weights not publicly released); the rest are open-weight. The best pass rate in each row is shaded green. Costs reconstruct token usage at official API pricing for GPT/Claude/Gemini and OpenRouter pricing for the other models, whose cost Claude Code on OpenRouter can inflate via low cache-hit rates.}
\definecolor{lightgreen}{RGB}{208,240,208}
\newcommand{\hl}[1]{\cellcolor{lightgreen}#1}
\footnotesize
\begin{tabular*}{\textwidth}{@{\extracolsep{\fill}}llcrcr@{}}
\toprule
& \multicolumn{3}{c}{Native Harness} & \multicolumn{2}{c}{Terminus-2} \\
\cmidrule(lr){2-4} \cmidrule(lr){5-6}
Model & Agent & Pass \% & \$/run & Pass \% & \$/run \\
\midrule
GPT-5.5$^\dagger$         & Codex       & \hl{28.0} & 178 & 19.5 & 155 \\
Claude Opus 4.8$^\dagger$ & Claude Code & \hl{20.7} & 269 & 15.9 & 293 \\
Gemini 3.1 Pro$^\dagger$  & Gemini CLI  & \hl{13.4} &  74 & 11.0 &  89 \\
GLM 5.2                  & Claude Code & \phantom{0}8.5 & 205 & \hl{\phantom{0}9.8} &  52 \\
Kimi K2.6                & Claude Code & \phantom{0}6.1 & 191 & \hl{\phantom{0}8.5} &  33 \\
MiniMax M3               & Claude Code & \phantom{0}3.7 &  66 & \hl{\phantom{0}6.1} &  18 \\
Qwen3.7 Max$^\dagger$     & Claude Code & \phantom{0}4.9 & 201 & \phantom{0}4.9 &  36 \\
DeepSeek V4 Pro          & Claude Code & \hl{\phantom{0}4.9} & 177 & \phantom{0}3.7 &  35 \\
MiMo V2.5 Pro            & Claude Code & \phantom{0}2.4 &  49 & \phantom{0}2.4 &   4 \\
\bottomrule
\end{tabular*}
\vspace{0.2em}
\label{tab:harbor-index-results}
\end{table}

\begin{figure}[ht]
  \centering
  \includegraphics[width=\linewidth]{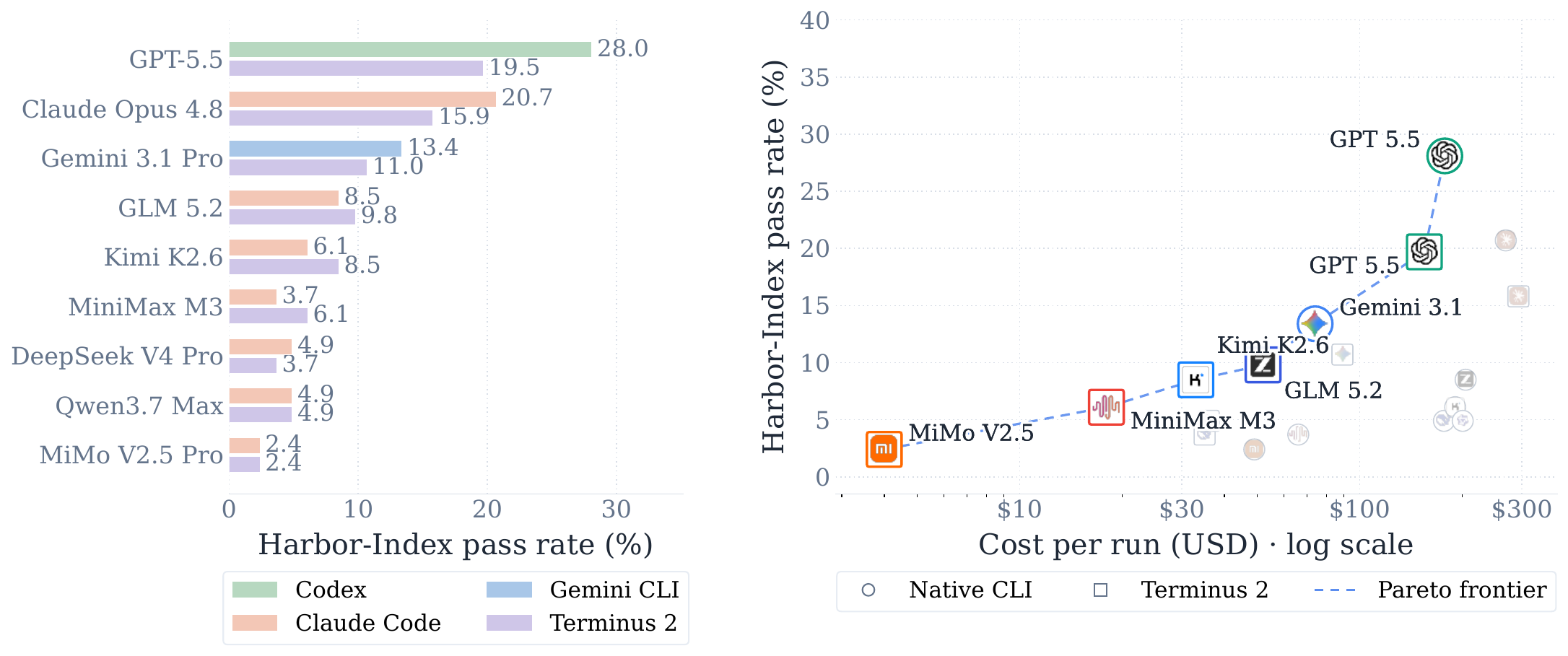}
  \caption{\textbf{Performance--cost tradeoff on Harbor-Index~1.0.}
  \textbf{Left:} Pass rates of the two evaluated harnesses for each model.
  \textbf{Right:} Pass rate versus cost for one complete Harbor-Index run (\HarborIndexTaskCount{} tasks).
  Each point is one model--harness configuration. Circles denote native harness and squares denote Terminus-2 for every model. Colored points and the dashed line mark the cost--performance Pareto frontier; gray points are dominated.}
  \label{fig:harbor-index-pareto}
\end{figure}

\Cref{tab:harbor-index-results} and \Cref{fig:harbor-index-pareto} show the suite is lightweight enough for repeated evaluation. The three highest-scoring models (GPT-5.5, Claude Opus 4.8, and Gemini 3.1 Pro) benefit most from their vendor-native harnesses, whereas several open-weight models sit on the cost--performance Pareto frontier under Terminus-2, trading pass rate for a far lower cost per run.



Under the current evaluation constraints, agent failures are separated into $453$ timeouts with no answer, $361$ near-miss solutions, and $445$ far-off fundamentally wrong answers. Capability and harness jointly shape these outcomes: the six lower-scoring models time out about twice as often as the three highest-scoring ones ($36.7\%$ versus $18.7\%$ of their respective rollouts), while the native harnesses use fewer tool calls and output tokens than Terminus-2 and, matched task by task, cut the timeout rate from $42\%$ to $26\%$. Terminus-2's bash-only action space, lacking native image and web tools, explains part of this gap and cautions against attributing every failure to the model alone. Interactive trajectories, scores, and task-level audit artifacts are on the Harbor-Index website.\footnote{\url{https://harbor-index.org}}


\section{Related Work}
\label{sec:related_work}

\paragraph{Agentic benchmarks and evaluation infrastructure.}
Agent benchmarks span tool and API use~\cite{patil2025bfcl,yao2024taubench,qin2024toolllm}, embodied and science environments~\cite{shridhar2021alfworld,wang2022scienceworld,yao2022webshop}, web and enterprise navigation~\cite{zhou2023webarena,deng2023mind2web,drouin2024workarena}, application and OS control~\cite{trivedi2024appworld,rawles2025androidworld,xie2024osworld}, software engineering from functions to repositories and terminals~\cite{chen2021codex,jimenez2023swebench,terminalbench}, scientific and workplace pipelines~\cite{chan2024mlebench,chen2024scienceagentbench,xu2024agentcompany}, and aggregate collections~\cite{liu2023agentbench,mialon2023gaia,mohammadi2025agentevalsurvey}. This diversity is informative but blocks aggregation: each benchmark ships its own action space, runtime assumptions, and scoring. Infrastructure work further shows a score is not a property of the model alone: interface, scaffold, cost budget, and holdout hygiene all move reported numbers~\cite{biderman2024reproducible,inspect2024,yang2024sweagent,xia2024agentless,kapoor2024agentsmatter}. HAL decomposes model, scaffold, and benchmark effects over 21,730 rollouts~\cite{kapoor2025hal}, while AstaBench controls cost and tool-access confounders within scientific research~\cite{astabench2026}.

\paragraph{Benchmark validity and task quality.}
Aggregate scores inherit the design flaws of their tasks and metrics~\cite{raji2021everything,bean2025constructvalidity,reuel2024betterbench}: label errors destabilize aggregates~\cite{northcutt2021labelerrors,gema2024done}, contamination erodes held-out validity~\cite{sainz2023contamination,white2025livebench}, and leaderboard mechanics skew comparisons~\cite{singh2025leaderboard}. This sharpens for agents, whose verifier executes code rather than matching a label. An imperfect verifier caps attainable accuracy regardless of compute, since resampling cannot lower its false-positive rate~\cite{stroebl2024inferencelimits}; agentic benchmarks misdesign setup or reward often enough to shift measured performance by up to $100\%$ relative~\cite{zhu2025agenticbestpractices}; and SWE-bench audits trace many ``resolved'' instances to leakage, memorization, and weak tests, enough that strengthening those tests reorders the leaderboard~\cite{aleithan2024swebenchplus,wang2025solvedissues,yu2025utboost,liang2025swebenchillusion}. Model judges and live services add further error~\cite{zheng2023judging,lu2025agentrewardbench,guo2024stabletoolbench}.

\paragraph{Efficient evaluation and benchmark redundancy.}
Model-by-benchmark score matrices are intrinsically low-rank~\cite{burnell2023structure,maimon2025lowrankfactors,papailiopoulos2026benchpress,zhang2025redundancy}, motivating curated subsets and adaptive item selection~\cite{polo2024tinybenchmarks,kipnis2025metabench,hofmann2025fluid}. Compression has limits: agreement estimates depend on unstandardized choices~\cite{perlitz2024benchbench}, subset prediction degrades on stronger models~\cite{zhang2025missesmark}, and micro-benchmarks need more items than assumed~\cite{yauney2025microbenchmarking,miller2024errorbars}.

\section{Discussion and Conclusion}
\label{sec:conclusion}

\paragraph{Limitations and future work.}
Adapter availability limits our large-scale evaluation to 54 of the collected adapter benchmarks; cost restricts us to a focused set of popular models under native harnesses and Terminus-2, so our findings do not span all LLMs, agent architectures, or evaluation settings and are not universal. Moreover, stronger future agents may exploit benchmark artifacts or evaluation loopholes more effectively that leads to currently unexploitable reward hacks and unreliable evaluation. Therefore, we plan maintain Harbor-Index as a live benchmark, updating it to reduce saturation, preserve quality, and integrating more adapters.


\paragraph{Conclusion.}
Harbor Adapters address the fragmentation of agentic evaluation: standardizing heterogeneous benchmarks under the Harbor task interface reduces benchmark--agent integration from \(\mathcal O(mn)\) to \(\mathcal O(m+n)\). Across validated adapters and a large-scale evaluation of \MainEvalBenchmarkCount{} benchmarks, \MainEvalConfigurationCount{} model--harness configurations spanning capability tiers, and \MainEvalTaskCount{} tasks, we find that base-model capability is the primary driver of performance gaps; many benchmarks contain redundant ranking signal; frontier models are more token-efficient but still substantially more costly; and low resolution tasks often reflect task-design flaws rather than genuine difficulty. Harbor-Index~1.0 distills this suite into \HarborIndexTaskCount{} compact, difficult, diverse, and audited tasks spanning \HarborIndexBenchmarkCount{} benchmarks. Together, Harbor Adapters and Harbor-Index provide reusable infrastructure and an affordable, high-quality testbed for more reliable, scalable, and informative evaluation of language-model agents.

\clearpage
\bibliographystyle{plain}
\bibliography{references}

\clearpage

\startcontents[appendices]
\section*{Appendix Table of Contents}
\printcontents[appendices]{}{0}{\normalsize}
\newpage

\appendix
\raggedbottom
\section{Authors}
\label{app:authors}

Each contributor below is listed with their full affiliation (superscript), keyed to the numbered institution list at the end of this section.

\subsection{Main Contributors \& Affiliations}

\appendixauthorgroup{Organization \& Execution Team}{Lin Shi$^{1}$, Haowei Lin$^{2}$, Zixuan Zhu$^{3}$, Xiaoyue Zhou$^{1}$, Xiang Li$^{4}$, Xiangning Lin$^{5}$, Yaxuan Deng$^{1}$, Han Xu$^{6}$, Yuangang Li$^{7}$, Shanda Li$^{5}$
% ...
}

\subsection*{Affiliations}

\begin{multicols}{2}
\raggedcolumns
\footnotesize
\begin{enumerate}[leftmargin=5mm,labelsep=1.5mm,itemsep=0pt,parsep=0pt,topsep=2pt]
\item Cornell Tech
\item Peking University
\item Nanyang Technological University
\item Stanford University
\item Carnegie Mellon University
\item University of Illinois Urbana-Champaign
\item University of California, Irvine
\item University of Southern California
\item Northeastern University
\item University of Hong Kong
\item International School of Krakow
\item Technology Innovation Institute
\item The Chinese University of Hong Kong
\item Independent Contributor
\item University of California, Santa Barbara
\item University of Texas at Austin
\item University of California, San Diego
\item Cornell University
\item 2077AI
\item Anthropic
\item Laude Institute
\item Microsoft
\item University of California, Berkeley
\item Dartmouth College
\item Harvard University
\item Princeton University
\item College of William and Mary
\item ByteDance
\item University of California, Riverside
\item Queen's University
\item Boston University
\item Stony Brook University
\item University of Warwick
\item Amazon
\item CoreWeave, Inc.
\item Allen Institute for AI
\item University of Macau
\item TU Darmstadt
\item University of Michigan
\item Institute of Science Tokyo
\item Massachusetts Institute of Technology
\item University of Wisconsin-Madison
\item University of Washington
\item University of Vienna
\item Instituto de Telecomunicações, Lisbon
\item Quesma
\item Meta
\item HarborCo
\item National University of Singapore
\item Monash University
\item The University of Tokyo
\item Michigan State University
\item Zhejiang University
\item University of Manchester
\item Australian National University
\item The Hong Kong University of Science and Technology
\item Renmin University of China
\item University of Basel
\item Rice University
\item BenchFlow
\item IBM
\end{enumerate}
\end{multicols}

\subsection{Additional Contributors}
We also thank the following contributors for help with adapter development, review, and evaluation:
\begin{multicols}{2}
\raggedcolumns
\footnotesize
\begin{itemize}[leftmargin=4mm,label=\textbullet,itemsep=0.5pt,parsep=0pt,topsep=2pt]
\item Zihao Wang (Peking University)
\item Joan Cabezas (Independent Contributor)
\item Thomas Aubry (ooakdata)
\item Bowen Xing (Independent Contributor)
\item Qiuyang Mang (University of California, Berkeley)
\item Shuting Zhao (University of California, San Diego)
\item Xinyu Yao (Rice University)
\item Achintya Paningapalli (University of California, Berkeley)
\item Benjamin Calvert (Brigham Young University)
\item Xinyu Lu (Institute of Software, Chinese Academy of Sciences)
\item Zhaowei Xu (Independent Contributor)
\end{itemize}
\end{multicols}

\section{Acknowledgment}
\label{app:acknowledgement}
The experiments and analyses of this study is generously supported by a series of partnership companies, frontier AI labs, and cloud sandbox providers:

\textbf{Partnership Companies:} Laude Institute~\cite{laude_institute}, 2077AI~\cite{org2077ai}, Docent~\cite{docent}, UniPat AI~\cite{unipat}

\textbf{Frontier AI Labs:} OpenAI (GPT)~\cite{openai_pricing}, Anthropic (Claude)~\cite{anthropic_pricing}, Google DeepMind (Gemini)~\cite{google_gemini_pricing}, Alibaba (Qwen)~\cite{alibaba_modelstudio_pricing}, DeepSeek (Deepseek)~\cite{deepseek_pricing}, MoonShot.AI (Kimi)~\cite{moonshot_kimi_pricing}, Xiaomi (Mimo)~\cite{xiaomi_mimo_pricing}, and Z.AI (GLM)~\cite{zai_pricing}

\textbf{Sandbox Providers:} Daytona~\cite{daytona} (for all CPU-only tasks), Modal~\cite{modal} (for GPU-required tasks specifically)

\section{Extended Related Work}
\label{app:extended_related_work}

This appendix extends \Cref{sec:related_work}. It adds two threads the main text
omits for space, namely the aggregate-evaluation tradition that precedes agentic
benchmarks and safety and robustness evaluation, and expands the main section's
clause-level citation groups into fuller discussions of prior work.

\paragraph{Aggregate and holistic evaluation.}
GLUE and SuperGLUE established standardized multi-task evaluation in language~\cite{wang2018glue,wang2019superglue}. Later suites broadened scope to knowledge and reasoning~\cite{hendrycks2021mmlu,srivastava2022beyond,suzgun2023challenging} and multilinguality~\cite{hu2020xtreme}, and advanced methodology through multi-metric scenario reporting~\cite{liang2022helm}, preference-based comparison~\cite{chiang2024chatbotarena}, and contamination-aware testing~\cite{white2025livebench}. Analogous efforts span retrieval and representation learning~\cite{thakur2021beir,muennighoff2023mteb} and vision~\cite{russakovsky2015imagenet,zhai2019vtab,koh2021wilds,barbu2019objectnet,croce2021robustbench,gupta2022grit,zamir2018taskonomy}, while Dynabench and Dynaboard replace fixed test sets with evolving, hosted comparisons~\cite{kiela2021dynabench,ma2021dynaboard}.

\paragraph{Evaluation infrastructure and reproducibility.}
Reproducibility work documents how prompt templates, decoding choices, few-shot formatting, and metric definitions materially change reported comparisons~\cite{biderman2024reproducible}; HELM adds multi-metric, scenario-level transparency~\cite{liang2022helm} and Dynaboard evaluates submitted models in a hosted setting to reduce reliance on self-reported results~\cite{ma2021dynaboard}. Recent systems extend this to model-plus-scaffold evaluation: Inspect~\cite{inspect2024}, HAL~\cite{kapoor2025hal}, AstaBench~\cite{astabench2026}, and Harbor~\cite{Harbor_Framework}, a containerized task and sandbox runtime. A parallel effort packages task or agent families behind one programmatic interface, including agentic environment simulators~\cite{liu2025gem}, interactive debugging environments~\cite{yuan2025debuggym}, generalist software-agent platforms~\cite{wang2025openhands}, and reproducible search sandboxes that replace commercial APIs whose drift silently breaks comparability~\cite{coelho2025deepresearchgym}.

Agent benchmarks make reproducibility harder. A static task reduces to examples, prompts, and labels; an agent benchmark also carries an environment (browser state, files, virtual services, shell tools), an execution model (sandboxing, network controls, stochastic dynamics), and a scoring layer (hidden tests, trajectory judgments), each of which shifts correctness and cost between runs. Reporting practice compounds this: evaluations often conflate the needs of model developers with those of downstream developers, omit cost, and lack adequate holdout sets, which invites benchmark-specific shortcuts~\cite{kapoor2024agentsmatter}.

\paragraph{Tool and API use.}
Tool-use benchmarks test whether models select APIs, generate valid arguments, follow protocols, and compose calls. Each targets a different layer: API retrieval and selection~\cite{patil2023gorilla,li2023apibank}, parameter prediction and function-call accuracy~\cite{patil2025bfcl}, multi-step invocation~\cite{qin2024toolllm,shen2024taskbench}, policy following~\cite{yao2024taubench}, and workflow planning~\cite{qiao2025worfbench,gioacchini2024agentquest}. Each therefore fixes one abstraction boundary, whether a function call, a tool trajectory, or a simulated service policy, and evaluates models against it.

\paragraph{Interactive and embodied environments.}
Early work placed agents in web interfaces, where sparse reward made exploration the bottleneck~\cite{liu2018reinforcement}, and in text games~\cite{cote2018textworld}, embodied households~\cite{shridhar2021alfworld}, simulated science labs~\cite{wang2022scienceworld}, and online shopping~\cite{yao2022webshop}. AgentBench, GAIA, and AgentGym then evaluated multi-step reasoning, planning, and tool use across broader collections~\cite{liu2023agentbench,mialon2023gaia,xi2024agentgym}. Web benchmarks cover live synthetic sites~\cite{zhou2023webarena,koh2024visualwebarena}, real-world navigation~\cite{deng2023mind2web,he2024webvoyager}, enterprise workflows~\cite{drouin2024workarena}, and shared execution frameworks~\cite{dechezelles2025browsergym,lu2024weblinx}. Beyond the browser, AppWorld, AndroidWorld, and OSWorld reach application and operating-system control~\cite{trivedi2024appworld,rawles2025androidworld,xie2024osworld}, now a large enough class to be surveyed separately~\cite{hu2025osagents}.

These works show that agent performance depends on environment grounding, state tracking, error recovery, and long-horizon planning. They also explain why agent benchmarks resist direct comparison: environments differ in action granularity, observation modality, allowed tools, termination criteria, and scoring.

\paragraph{Code, scientific, and workplace agents.}
HumanEval tests program synthesis in self-contained functions~\cite{chen2021codex}, while SWE-bench raises the bar to issue resolution in real repositories~\cite{jimenez2023swebench}. SWE-agent shows the agent--computer interface itself substantially affects software-engineering performance~\cite{yang2024sweagent}, building on ReAct's interleaved reasoning-and-acting pattern~\cite{yao2023react} and the self-reflection variants that followed~\cite{shinn2023reflexion}; Agentless conversely shows a much simpler pipeline can be competitive, so scaffold complexity is not monotonically useful~\cite{xia2024agentless}. Terminal-Bench extends this line to shells, files, and long-running processes rather than a structured patch interface~\cite{terminalbench}, and time-horizon studies quantify how the task length agents can complete has grown~\cite{kwa2025longtasks}. Beyond software engineering, MLE-bench evaluates agents on 75 Kaggle machine-learning engineering competitions against human leaderboard baselines~\cite{chan2024mlebench}. ScienceAgentBench deliberately decomposes data-driven discovery into 102 expert-validated tasks from peer-reviewed publications, scoring the generated program, its execution results, and its cost~\cite{chen2024scienceagentbench}. TheAgentCompany combines browsing, coding, communication, and service manipulation inside a simulated organization~\cite{xu2024agentcompany}.

These benchmarks expose failure modes invisible in static QA. The modes fall into three classes: environment (flaky dependencies, setup failures, resource limits, timeouts), scoring (hidden-test mismatch, partial credit, trajectory judgments), and scaffold (tool affordances, retry policies, interface conventions). Each class shifts reported scores independently of the model under test.

\paragraph{Safety, robustness, and adversarial evaluation.}
Tool-using agents take actions, read untrusted content, and execute multi-step plans whose intermediate states are themselves attack surfaces. AgentDojo evaluates prompt-injection attacks and defenses over realistic tasks and untrusted data~\cite{debenedetti2024agentdojo}, AgentHarm measures whether agents refuse malicious requests while retaining capability~\cite{andriushchenko2024agentharm}, and OS-Harm extends this to computer-use agents acting on a graphical interface~\cite{kuntz2025osharm}. A parallel line targets the reliability of agent evaluation itself: StableToolBench stabilizes tool calls through virtualized APIs and cached responses~\cite{guo2024stabletoolbench}, while AgentRewardBench and Agent-as-a-Judge ask whether model judges can assess behavior beyond final-answer correctness~\cite{lu2025agentrewardbench,zhuge2024agentasjudge}.

\paragraph{Benchmark validity and task quality.}
A distinct literature asks not how to run benchmarks but whether their scores license the conclusions drawn from them. Position and methodology papers make three arguments: that a few influential benchmarks should not stand in for general capability~\cite{raji2021everything}; that benchmark construction should state its measurement assumptions explicitly and testably~\cite{liu2024ecbd}; and that moving from a score to a claim about an abstract task requires further assumptions made explicit~\cite{freiesleben2025epistemology}. Empirically, a review of 445 benchmarks by 29 expert reviewers finds recurring patterns in measured phenomena, task design, and scoring metrics that undermine validity~\cite{bean2025constructvalidity}, and an assessment framework applied across widely used benchmarks reaches similar conclusions about documentation and design practice~\cite{reuel2024betterbench}. Concrete threats are well documented. Test-set label errors are pervasive, at least $3.3\%$ on average across ten widely used datasets, and destabilize conclusions to the point that lower-capacity models can overtake higher-capacity ones once mislabeling is accounted for~\cite{northcutt2021labelerrors}; annotation errors change how aggregates should be read~\cite{gema2024done,neuhaus2025repope}; contamination compromises held-out validity and is hard to measure per benchmark~\cite{sainz2023contamination}; and leaderboard mechanics distort rankings~\cite{singh2025leaderboard}.

For agents the verifier is executable code rather than a stored label, which turns these validity questions into engineering defects. Because resampling cannot lower a verifier's false-positive rate, that rate upper-bounds the accuracy any amount of resampling can attain, regardless of compute budget~\cite{stroebl2024inferencelimits}. The Agentic Benchmark Checklist documents systematic task-setup and reward-design flaws in widely used agentic benchmarks, such as insufficient test cases in SWE-bench Verified and empty responses scored as success in $\tau$-bench, with distortions reaching $100\%$ in relative terms~\cite{zhu2025agenticbestpractices}. Software-engineering audits make the magnitude concrete: roughly a third of successful SWE-bench patches trace to solution leakage in the issue text, and a comparable share to test suites too weak to detect an incorrect patch~\cite{aleithan2024swebenchplus}. Differential patch testing shows plausible patches diverging behaviorally from developer ground truth, inflating reported resolution rates~\cite{wang2025solvedissues}; augmenting test suites relabels hundreds of patches and reorders leaderboards~\cite{yu2025utboost}; and memorization rather than reasoning on SWE-bench instances complicates held-out interpretation~\cite{liang2025swebenchillusion}. The judging layer is a matching concern, whether the judge is a preference model~\cite{zheng2023judging}, a trajectory judge~\cite{lu2025agentrewardbench,zhuge2024agentasjudge}, or a stand-in for human annotators, a substitution for which no standard procedure existed until the recently proposed alternative annotator test~\cite{calderon2025annotatortest}.

\paragraph{Meta-evaluation, efficiency, and benchmark compression.}
Much of a large suite is redundant. Factor-analytic work finds model-by-benchmark score matrices are intrinsically low-rank, a few interpretable latent dimensions explaining most variance and implying substantial redundancy among nominally distinct tasks~\cite{burnell2023structure,maimon2025lowrankfactors,papailiopoulos2026benchpress}; the same structure makes downstream performance predictable from a few latent capability measures~\cite{ruan2024observational}, and correlation-based analyses of multimodal suites report comparable redundancy across benchmarks and capability dimensions~\cite{zhang2025redundancy}. This licenses cheaper evaluation through curated and distilled subsets~\cite{polo2024tinybenchmarks,kipnis2025metabench,vivek2024anchorpoints}, psychometrically adaptive item selection~\cite{hofmann2025fluid,zhou2025lostbenchmarks}, principled metric selection~\cite{procaccia2025metritocracy}, and label-efficient active testing~\cite{berrada2025activetesting}. Cost-aware framings make the tradeoff explicit rather than incidental~\cite{erol2025costofpass,kapoor2024agentsmatter}.

Recent results delimit these methods. Benchmark agreement testing has no standardized procedure, and overlooked methodological choices, including which models the agreement is computed over, significantly change its conclusions, so apparent redundancy is partly an artifact of how it is measured~\cite{perlitz2024benchbench}. Benchmark prediction from a subset depends on similarity to previously observed models and degrades sharply when extrapolating to stronger ones, failing exactly at the frontier where evaluation is most needed~\cite{zhang2025missesmark}. Micro-benchmarks frequently cannot rank nearby models reliably, requiring far more items than the handful sometimes suggested, at which point random sampling is competitive with sophisticated selection~\cite{yauney2025microbenchmarking}. Small evaluations therefore require explicit uncertainty quantification~\cite{miller2024errorbars}, on top of the setup sensitivity documented above.

\section{Details for Harbor Adapters}
\label{app:adapter_details}

This appendix describes the motivation for Harbor Adapters
(\S\ref{app:adapter_motivation}), the agent interaction modes we use to
categorize benchmarks (\S\ref{app:agent_modes}), the adapter
construction workflow (\S\ref{app:construction_workflow}), the parity
experiment protocol (\S\ref{app:parity_experiments}), the full adapter
catalog (\S\ref{app:adapter_catalog}), per-benchmark adaptation and
parity details (\S\ref{app:per_benchmark}), and lessons learned from
building adapters at scale (\S\ref{app:lessons}).

\subsection{Motivation}
\label{app:adapter_motivation}

As the number of benchmark datasets and agents grows, agent evaluation
has become increasingly costly: supporting $m$ benchmarks with $n$
agents naively requires $\mathcal O(mn)$ effort, since each benchmark-agent pair
involves its own harness integration, environment plumbing, and
evaluation logic. Inconsistent environments and task-design issues
further compound this cost, and the resulting leaderboards are often not
directly comparable because instructions, tools, and execution
configurations differ across benchmarks.

Despite the diversity of task formats, most benchmarks share a common
underlying abstraction: \textbf{instruction}, \textbf{environment},
\textbf{tests}, and \textbf{solution}. The instruction specifies what
the agent must accomplish; the environment specifies where the agent
acts; the tests verify success; and the solution demonstrates
feasibility. This abstraction, introduced by
Terminal-Bench~\citep{terminalbench}, applies across the vast majority
of agentic and non-agentic benchmarks.

Harbor Adapters translate heterogeneous dataset-specific formats into
this shared schema. Together with the Harbor infrastructure,
which provides sandboxed execution (local and cloud) and reward-based
evaluation, the adapter layer reduces the per-benchmark or per-agent
integration cost from $\mathcal O(mn)$ to $\mathcal O(m + n)$, and enables running tens of
datasets, tens of thousands of tasks, and millions of rollouts across
many model--harness configurations under identical evaluation conditions.

\subsection{Agent Interaction Modes}
\label{app:agent_modes}

We categorize each benchmark by its \textit{agent interaction mode},
which captures whether and how agentic capabilities affect the
evaluation outcome.

\textbf{Agentic} benchmarks are explicitly
designed for agent-based evaluation; tasks assume iterative interaction
with an environment (e.g., filesystem, shell, browser, external tools),
multi-step reasoning, and/or test-driven feedback, and evaluating them
without agent capabilities is infeasible or inconsistent with the
original design.

\textbf{Non-Agentic} benchmarks are designed for
direct LLM evaluation; tasks are single-pass, with the model producing
an answer from the prompt alone and no access to tools or an
environment, so introducing agent capabilities would not meaningfully
change what the benchmark measures.

\textbf{Modified-Agentic}
benchmarks are originally non-agentic, but exposing the underlying
environment to an agent (e.g., filesystem access, iterative execution,
test-driven refinement) changes the evaluation outcome in a measurable
way; these are typically coding tasks whose single-pass protocol
underspecifies what a capable agent could achieve with terminal access.

Agent interaction mode is a property of the \textit{benchmark} itself.
It is used throughout the catalog (\S\ref{app:adapter_catalog}) and
determines which integration scenario applies when we construct an
adapter (\S\ref{app:parity_experiments}).

\subsection{Adapter Construction Workflow}
\label{app:construction_workflow}

Constructing and maintaining adapters at scale is an engineering-intensive
process that requires careful handling of dataset formats, execution
environments, and evaluation protocols. We organize this process into
three stages: \textit{creation}, \textit{review}, and
\textit{standardization}.

\subsubsection{Creation}
Each adapter is constructed through the following steps.
\textbf{(1)}~Parsing and restructuring benchmark content into the
Harbor schema (instruction, environment, tests, solution).
\textbf{(2)}~Reconstructing or aligning the execution environment
(Docker images, dependencies, network policies, resource limits).
\textbf{(3)}~Integrating evaluation logic and test cases, and verifying
that oracle solutions achieve a 100\% pass rate.
\textbf{(4)}~Running parity experiments against the original benchmark
(\S\ref{app:parity_experiments}).

To scale this process, we developed contributor guidelines, a
prompt-driven tutorial for coding agents, and an agent skill
(\texttt{/create-adapter}) that handles context sharing and enables
parallel adapter development across the team.

\subsubsection{Review}
\label{app:adapter_construction_workflow:review}
Adapter review is equally intensive, as it must ensure semantic
fidelity to the original benchmark, correctness of the execution
environment, and conformance to the Harbor schema. We trained a
dedicated team of five reviewers and developed automated review bots
to accelerate quality assurance. Every adapter pull request undergoes
a three-stage review process.
\textbf{(1)}~An \textit{automated bot review} that checks code style,
schema conformance, documentation completeness, and the presence of a
passing oracle run.
\textbf{(2)}~A \textit{human review} by one to three trained
reviewers, who verify adaptation correctness, environment
reproducibility, and parity-experiment results.
\textbf{(3)}~A \textit{final approval} by the team lead, who signs off
on non-trivial design decisions (e.g., parity-set selection,
custom-agent ports) before the adapter is merged.
Adapters that fail any stage are returned to the contributor with
reviewer comments and re-enter the pipeline once addressed.

\subsubsection{Standardization}
As more adapters are integrated, requirements and best practices continue to evolve: the Harbor schema gains new fields, shared utilities are refactored, and documentation conventions tighten. To prevent format drift, a standardization team performs periodic passes that co-evolve three artifacts in parallel: existing adapters, the contributor tutorial, and the bot review prompts. In this way, the schema, the guidance new contributors receive, and the checks enforced at review time all stay aligned, while preserving dataset integrity and parity results. This maintenance work is scheduled independently of new-adapter contributions and is a first-class part of the workflow rather than one-off cleanup.

\subsection{Parity Experiments}
\label{app:parity_experiments}

\subsubsection{Challenges in Benchmark Portability}

Porting a benchmark to a new harness is semantics-preserving \emph{in intent}, but rarely so in practice. Small differences in prompt templates, tool configurations, default decoding parameters, Docker base images, or scoring scripts can shift reported scores by several points, even when the underlying task set is unchanged. Published results are often not reproducible from the released code alone: we have routinely encountered oracle solutions that fail the benchmark's own tests, scoring scripts with undocumented non-determinism, and leaderboard entries produced by pipelines that diverge from the public release. Without explicit verification, a user of a Harbor adapter cannot tell whether a score difference reflects a genuine capability gap, a harness artifact, or a latent bug on either side. We therefore treat parity verification as an essential step in building every adapter.

\subsubsection{Design and Evaluation Metrics}

A \textit{parity experiment} compares the Harbor-adapted version of a benchmark against the original under matched conditions. The same agent, model, tool set, prompt template, decoding parameters, and execution configuration are applied on both sides. Each side is run $k$ times ($k=3$ by default) to account for stochasticity, and we report the mean ($\bar{x}$) $\pm$ sample standard error of the mean ($s_{\bar{x}}$), computed as $s_{\bar{x}} = s/\sqrt{k}$ where $s$ is the sample standard deviation across the $k$ runs. An adapter passes parity when the original and Harbor means agree within a margin consistent with their standard errors. Adapters that fail the parity are debugged and re-run. Cases where parity cannot be achieved due to irreducible differences, for example dependence on a non-deterministic external service, are documented per benchmark in \S\ref{app:per_benchmark}.

\subsubsection{Agent Integration Scenarios}

A parity experiment requires the same agent to run against both the original benchmark and the Harbor adapter. Whether such an agent already exists or has to be built depends on the original benchmark's design. During our development, we distinguish three main scenarios:

\textbf{Scenario 1: Harbor Supported Agent.}
The original benchmark already uses an agent that Harbor also supports natively (e.g., Claude Code, Codex CLI). Adaptation reduces to aligning versions, tool configurations, execution arguments, and implicit environment assumptions between the two sides. The parity experiment runs the same agent against both.

\textbf{Scenario 2: Vanilla LLM.}
The original benchmark provides no agent and was designed for direct LLM prompting. We fork the original repository and add a Harbor-supported CLI agent, matching the prompt and decoding protocol of the original single-pass evaluation. Minor modifications, such as input/output formatting and prompt tweaks, are permitted as long as they are applied symmetrically to both the original and Harbor sides. The parity experiment then runs this agent against both the forked original and the Harbor adapter. We apply the same treatment to non-agentic and modified-agentic benchmarks alike.

\textbf{Scenario 3: Custom Agent.}
The original benchmark ships a custom agent that Harbor does not support natively (e.g., a benchmark-specific ReAct variant, a deep-research agent, or a multi-agent scaffold). We port the custom agent into Harbor, and the parity experiment runs it against the original benchmark. Where feasible, we also run a Harbor-native agent (e.g., Claude Code) as a cross-check. When the custom agent cannot be cleanly ported due to tight coupling with benchmark internals, we allow the benchmark design to diverge, and ship both the original-agent variant and a CLI-agent variant in Harbor for community use, and document the limitation in \S\ref{app:per_benchmark}.

The mapping from agent interaction mode to integration scenario is direct. All agentic benchmarks fall into Scenario~1 or Scenario~3 depending on whether their native agent is supported by Harbor. All non-agentic and modified-agentic benchmarks fall into Scenario~2.

\subsubsection{Parity Set Selection}
\label{app:parity_set}

For most benchmarks, parity is run on the full task set. When the full set is too costly to run $k$ times (e.g., long-horizon agentic tasks or large task counts), we run parity on a representative subset, the \emph{parity set}, selected by one of the following criteria:

\textbf{(1) Random stratified sampling} over difficulty and/or subdomain,
when such metadata is available.

\textbf{(2) Official subsets}, when the benchmark authors provide a
smaller canonical subset (e.g., a ``Lite'' or ``Verified'' split).

\textbf{(3) Curated subsets}, when stability or environment-setup cost
requires hand-picked representative instances.

For each adapter, the parity-set size, selection criterion, and the specific task IDs are recorded and explained in the adapter's README at \texttt{adapters/<benchmark>/README.md} in the Harbor Adapters repository, and summarized in the corresponding paragraph of \S\ref{app:per_benchmark}.

\subsection{Adapter Catalog}
\label{app:adapter_catalog}

\begin{table*}[t]
\centering
\caption{93 Benchmarks integrated via Harbor adapters, organized by domain and subdomain. Colors mark agentic character: \textcolor{agentic}{\textbf{agentic}}, \textcolor{modagentic}{\textbf{modified-agentic}}, \textcolor{nonagentic}{\textbf{non-agentic}}.}
\label{tab:benchmark_categorization_clustered}
\scalebox{0.8}{%
\renewcommand{\arraystretch}{1.3}%
\begin{tabularx}{1.20\textwidth}{@{}
    >{\hsize=0.42\hsize\raggedright\arraybackslash}X
    >{\hsize=0.78\hsize\raggedright\arraybackslash}X
    >{\hsize=1.80\hsize\raggedright\arraybackslash}X
@{}}
\toprule
\textbf{Domain} & \textbf{Subdomain} & \textbf{Benchmarks} \\
\midrule
\multirow{6}{=}{\textbf{Software Engineering}}
 & \textit{Repo-level Issue Resolution and Testing}
 & \textcolor{agentic}{ABC-Bench~\cite{yang2026abcbench}, CooperBench~\cite{khatua2026cooperbench}, DevEval~\cite{li2024deveval}, Multi-SWE-Bench~\cite{zan2025multiswebench}, SWE-Bench-Multilingual~\cite{jimenez2023swebench}, SWE-Bench~Pro~\cite{deng2025swebenchproaiagents}, SWE-bench~Verified~\cite{jimenez2023swebench}, SWE-Gym~\cite{pan2025swegym}, SWE-rebench~\cite{badertdinov2025swerebench}, SWE-smith~\cite{yang2025swesmith}, SWT~Bench~\cite{mundler2025swtbench}} \\
\addlinespace
 & \textit{Feature Development}
 & \textcolor{agentic}{FeatBench~\cite{chen2026featbench}, FeatureBench~\cite{libercoders2026featurebench}, ProgramBench~\cite{yang2026programbench}, SWE-Lancer~\cite{miserendino2025swelancerfrontierllmsearn}, WebGen-Bench~\cite{lu2025webgenbench}} \\
\addlinespace
 & \textit{Competitive \& Function-level Coding}
 & \textcolor{modagentic}{Aider~Polyglot~\cite{aiderpolyglot2024}, AutoCodeBench~\cite{chou2025autocodebench}, BigCodeBench~\cite{zhuo2025bigcodebench}, EvoEval~\cite{xia2024evoeval}, Frontier-CS-Algorithm~\cite{mang2025frontiercs}, LiveCodeBench~\cite{jain2024livecodebench}, USACO~\cite{shi2024language}};\;
   \textcolor{nonagentic}{CanItEdit~\cite{cassano2024canitedit}, HumanEvalFix~\cite{muennighoff2023octopack}, QuixBugs~\cite{lin2017quixbugs}}, UniCode~\cite{zheng2026unicode} \\
\addlinespace
 & \textit{Performance Optimization}
 & \textcolor{agentic}{GSO~\cite{shetty2025gso}};\;
   \textcolor{modagentic}{AlgoTune~\cite{press2025algotune}} \\
\addlinespace
 & \textit{Language Translation}
 & \textcolor{modagentic}{CRUST-Bench~\cite{khatry2025crustbench}} \\
\addlinespace
 & \textit{DevOps \& Build Systems}
 & \textcolor{agentic}{CompileBench~\cite{compilebench2025}, DevOpsGym~\cite{tang2026devopsgym}} \\
\midrule
\multirow{2}{=}{\textbf{Mathematics \& Reasoning}}
 & \textit{Competition Mathematics}
 & \textcolor{nonagentic}{AIME~\cite{ye2025aimepreview}, IneqMath~\cite{sheng2025ineqmath}, Omni-Math~\cite{gao2024omnimath}} \\
\addlinespace
 & \textit{Abstract \& Procedural Reasoning}
 & \textcolor{modagentic}{SATBench~\cite{wei2025satbench}};\;
   \textcolor{nonagentic}{ARC-AGI-2~\cite{chollet2025arcagi2}, KUMO~\cite{lin2025kumo}, Reasoning~Gym~\cite{stojanovski2025reasoninggym}} \\
\midrule
\multirow{2}{=}{\textbf{Knowledge \& Long Context}}
 & \textit{Expert \& Multi-subject QA}
 & \textcolor{nonagentic}{CL-Bench~\cite{dou2026clbench}, GPQA~Diamond~\cite{rein2024gpqa}, Humanity's~Last~Exam~\cite{phan2026hle}, MMMLU~\cite{hendrycks2021mmlu}, SimpleQA~\cite{wei2024simpleqa}} \\
\addlinespace
 & \textit{Long-Context Reasoning}
 & \textcolor{nonagentic}{AA-LCR~\cite{aalcr2025}, LoCoMo~\cite{maharana2024evaluating}} \\
\midrule
\multirow{3}{=}{\textbf{Scientific Research}}
 & \textit{End-to-end Research Workflows}
 & \textcolor{agentic}{AstaBench~\cite{astabench2026}, ML-Dev-Bench~\cite{padigela2025mldevbench}, MLR-Bench~\cite{chen2025mlrbench}, ReplicationBench~\cite{ye2025replicationbench}, RExBench~\cite{edwards2025rexbench}, ScienceAgentBench~\cite{chen2024scienceagentbench}, SLDBench~\cite{lin2025sldbench}};\;
   \textcolor{modagentic}{MLGym-Bench~\cite{mlgym2025}} \\
\addlinespace
 & \textit{Scientific Computing}
 & \textcolor{agentic}{LLM-SRBench~\cite{shojaee2025llmsrbench}};\;
   \textcolor{modagentic}{CodePDE~\cite{li2025codepde}, ResearchCodeBench~\cite{hua2025researchcodebench}, SciCode~\cite{tian2024scicode}};\;
   \textcolor{nonagentic}{QCircuitBench~\cite{yang2025qcircuitbench}} \\
\addlinespace
 & \textit{Biomedical Research}
 & \textcolor{agentic}{BIX-Bench~\cite{mitchener2025bixbench}};\;
   \textcolor{nonagentic}{LAB-Bench~\cite{laurent2024labbench}} \\
\midrule
\multirow{3}{=}{\textbf{Agents, Tools \& Systems}}
 & \textit{Tool Use \& Assistants}
 & \textcolor{agentic}{ACE-Bench~\cite{chen2025acebench}, GAIA~\cite{mialon2023gaia}, GAIA2~\cite{froger2026gaia2}, LongCLIBench~\cite{feng2026longclibench}, OSWorld~\cite{xie2024osworld}, Tau3-Bench~\cite{yao2024taubench}};\;
   \textcolor{nonagentic}{AMA-Bench~\cite{zhao2026amabench}, BFCL~\cite{patil2025bfcl}} \\
\addlinespace
 & \textit{Deep Research \& Web Agents}
 & \textcolor{agentic}{DeepResearch-Bench-II~\cite{du2025deepresearch}, DeepSynth~\cite{paul2026deepsynth}, Seal-0~\cite{pham2026sealqa}, WideSearch~\cite{wong2025widesearch}} \\
\addlinespace
 & \textit{Interactive Games}
 & \textcolor{nonagentic}{TextArena~\cite{guertler2025textarena}} \\
\midrule
\textbf{Data \& Analytics}
 & \textit{Text-to-SQL \& Data Science}
 & \textcolor{agentic}{ADE-Bench~\cite{stancil2025adebench}, DA-Code~\cite{huang2024dacode}, DABstep~\cite{egg2025dabstep}, KramaBench~\cite{lai2025kramabench}};\;
   \textcolor{modagentic}{BIRD-Bench~\cite{li2023birdbench}, DS-1000~\cite{lai2023ds1000}, Spider~2~\cite{lei2024spider2}} \\
\midrule
\multirow{4}{=}{\textbf{Professional Domains}}
 & \textit{Finance \& Trading}
 & \textcolor{agentic}{FinanceAgent~\cite{valsai2024financeagent}};\;
   \textcolor{modagentic}{PIXIU~\cite{xie2023pixiu}} \\
\addlinespace
 & \textit{Business / Professional Work}
 & \textcolor{agentic}{CRMArena~\cite{huang2025crmarena}, SpreadsheetBench~\cite{ma2024spreadsheetbench}, TheAgentCompany~\cite{xu2024agentcompany}} \\
\addlinespace
 & \textit{Healthcare \& Clinical}
 & \textcolor{agentic}{MedAgentBench~\cite{jiang2025medagentbench}} \\
\addlinespace
 & \textit{Legal}
 & \textcolor{nonagentic}{LawBench~\cite{fei2023lawbench}} \\
\midrule
\multirow{2}{=}{\textbf{Safety \& Security}}
 & \textit{Cybersecurity}
 & \textcolor{agentic}{CyberGym~\cite{cybergym2025}} \\
\addlinespace
 & \textit{Jailbreak Robustness}
 & \textcolor{nonagentic}{StrongReject~\cite{souly2024strongreject}} \\
\midrule
\multirow{3}{=}{\textbf{Multimodal}}
 & \textit{Audio Understanding}
 & \textcolor{nonagentic}{MMAU~\cite{sakshi2024mmau}} \\
\addlinespace
 & \textit{Graphic / Visual Design}
 & \textcolor{modagentic}{GraphDesignBench~\cite{deganutti2026gdb}} \\
\addlinespace
 & \textit{Autonomous Driving \& Sensor Reasoning}
 & \textcolor{modagentic}{RefAV~\cite{davidson2025refav}} \\
\bottomrule
\end{tabularx}%
}
\end{table*}

Tables~\ref{tab:adapter_catalog_part1} and~\ref{tab:adapter_catalog_part2}
list every adapter whose parity experiment is complete, with its
domain/subdomain, integration scenario,
agent interaction mode, parity agent, parity model, number of runs,
parity-set size, and parity scores (original vs. Harbor). The
parity-set column is formatted as $\mathit{parity}/\mathit{full}$,
where $\mathit{parity}$ is the number of tasks actually used for
parity verification and $\mathit{full}$ is the total size of the
benchmark. Score values are reported as mean $\pm$ SEM over $k$
runs (SEM omitted when $k=1$). \texttt{N/A} indicates that no
direct Harbor-vs-original (or Harbor-vs-Terminal-Bench) parity
score is available --- either because the upstream benchmark has
no agent harness so traditional parity does not apply, or because
only an upstream Terminal-Bench adapter versus original-benchmark
comparison was measured. Because parity verification is part of
adapter construction rather than of the evaluation itself, these
tables span more benchmarks than the \MainEvalBenchmarkCount{}
used in our large-scale evaluation: nine rows (CooperBench,
FeatBench, WebGen-Bench, Frontier-CS-Algorithm, DevOpsGym,
ScienceAgentBench, MLGym-Bench, ACE-Bench, CRMArena) have verified
adapters but were held out of the experiments for compute reasons,
and are therefore also listed in \S\ref{app:bench_not_included};
benchmarks already authored in the Harbor schema (Terminal-Bench,
CompileBench, SkillsBench) require no adapter and are described in
\S\ref{app:bench_harbor_format}.

\begin{table*}[t]
\centering
\caption{Harbor adapter catalog (Part 1 of 2): Software Engineering,
Mathematics \& Reasoning, and Knowledge \& Long Context.
\textbf{Sc.}: integration scenario (1 = Harbor-Agent, 2 = Vanilla-LLM,
3 = Custom-Agent). \textbf{Mode}: agent interaction mode
(A = Agentic, M = Modified-Agentic, N = Non-Agentic). $\bm{k}$: number
of parity runs. \textbf{Set}: $\mathit{parity}/\mathit{full}$ task
count. \textbf{Orig.}~/~\textbf{Harbor}: pass rate (\%) as mean $\pm$
SEM over $k$ runs.}
\label{tab:adapter_catalog_part1}
\scalebox{0.84}{%
\renewcommand{\arraystretch}{1.15}%
\setlength{\tabcolsep}{4pt}%
\begin{tabular}{@{}
    >{\raggedright\arraybackslash}p{2.6cm}
    >{\raggedright\arraybackslash}p{1.8cm}
    c c
    >{\raggedright\arraybackslash}p{1.7cm}
    >{\raggedright\arraybackslash}p{1.8cm}
    c c c c
@{}}
\toprule
\textbf{Benchmark} & \textbf{Subdomain}
 & \textbf{Sc.} & \textbf{Mode}
 & \textbf{Parity Agent} & \textbf{Parity Model}
 & $\bm{k}$ & \textbf{Set}
 & \textbf{Orig.} & \textbf{Harbor} \\
\midrule

SWE-Bench-Multilingual & Repo Issues   & 2 & A & codex  & gpt-5-mini & 3 & 50/300   & 33.3$\pm$0.7 & 33.3$\pm$3.3 \\
SWE-Bench Pro          & Repo Issues   & 2 & A & codex       & gpt-5-mini       & 6 & 100/731   & 35.7$\pm$0.6 & 36.3$\pm$1.0 \\
SWE-bench Verified     & Repo Issues   & 1 & A & mini-swe-agent & gpt-5-mini    & 3 & 499/500   & 56.3$\pm$0.0 & 54.5$\pm$0.7 \\
SWE-smith              & Repo Issues   & 1 & A & mini-swe-agent & claude-3-haiku  & -- & 100/100   & 0.0 & 0.0 \\
SWT Bench              & Repo Issues   & 1 & A & claude-code    & haiku-4-5        & -- & 433/433   & 20.55$\pm$0.0 & 20.55$\pm$0.0 \\
CooperBench            & Feature Dev   & 1 & A & openhands-sdk  & gemini-3-flash   & 3 & 50/652    & 32.7$\pm$1.3 & 30.7$\pm$1.3 \\
FeatBench              & Feature Dev   & 1 & A & trae-agent     & deepseek-v3.2    & 3 & 156/156   & 49.8$\pm$2.3 & 49.4$\pm$1.1 \\
FeatureBench           & Feature Dev   & 1 & A & codex          & gpt-5-mini       & 2 & 30/200    & 13.3$\pm$0.0 & 15.0$\pm$1.7 \\
SWE-Lancer             & Feature Dev   & 2 & A & claude-code    & claude-sonnet-4  & 5 & 463/463   & 48.4$\pm$1.0 & 47.6$\pm$0.4 \\
WebGen-Bench           & Feature Dev   & 2 & M & aider          & gpt-5-mini       & 3 & 101/101   & 16.7$\pm$0.9 & 16.2$\pm$1.6 \\
Aider Polyglot         & Coding        & 2 & N & claude-code    & claude-3-haiku   & 1 & 225/225   & 2.7$\pm$0.0  & 2.7$\pm$0.0  \\
BigCodeBench           & Coding        & 2 & M & codex          & gpt-5-mini       & 3 & 145/145   & 32.7$\pm$0.4 & 34.0$\pm$1.6 \\
FrontierCS             & Coding        & 2 & M & claude-code    & opus-4-6         & 3 & 10/172    & 68.9$\pm$11.5 & 53.4$\pm$9.9 \\
LiveCodeBench          & Coding        & 2 & N & claude-code    & haiku-4-5        & 4 & 100/1055  & 54.5$\pm$1.5 & 53.3$\pm$1.0 \\
USACO                  & Coding        & 2 & N & claude-code    & haiku-4-5        & 3 & 304/304   & 40.3$\pm$0.4 & 43.0$\pm$1.4 \\
HumanEvalFix           & Coding        & 1 & N & openhands      & gpt-5-mini       & 3 & 164/164   & 98.2$\pm$0.6 & 98.0$\pm$0.4 \\
QuixBugs               & Coding        & 2 & N & codex          & gpt-5-mini       & 3 & 80/80     & 88.4$\pm$0.4 & 87.9$\pm$0.4 \\
GSO                    & Perf. Opt.    & 1 & A & openhands      & gpt-5.1          & 2 & 102/102   & 13.7$\pm$1.0 & 13.2$\pm$1.5 \\
AlgoTune               & Perf. Opt.    & 2 & A & terminus-2     & gpt-5-mini       & 3 & 154/154   & 1.23$\pm$0.02 & 1.23$\pm$0.02 \\
CRUST-Bench            & Lang. Trans.  & 2 & M & codex          & gpt-5-nano       & 3 & 100/100   & 58.3$\pm$0.7 & 58.3$\pm$1.2 \\
DevOpsGym              & DevOps        & 1 & A & codex          & gpt-5-mini       & 3 & 50/733    & 22.7$\pm$1.8 & 22.0$\pm$0.0 \\
\midrule

AIME                   & Comp. Math    & 2 & N & codex          & gpt-5-nano       & -- & 30/30      & N/A & N/A \\
IneqMath               & Comp. Math    & 2 & N & codex      & gpt-4o-mini      & 3 & 100/100    & 50.0$\pm$1.5 & 52.0$\pm$0.7 \\
Omni-Math              & Comp. Math    & 2 & N & codex      & gpt-5.3-codex    & 2 & 100/4428   & 80.0$\pm$2.0 & 81.0$\pm$2.0 \\
ARC-AGI-2              & Abs. Reason.  & 2 & N & codex      & gpt-5.2          & 6 & 167/167    & 36.0$\pm$0.8 & 35.8$\pm$0.8 \\
KUMO                   & Abs. Reason.  & 2 & N & kumo-vanilla & gpt-5-mini     & 3 & 212/5300   & 88.7$\pm$0.7 & 89.9$\pm$0.7 \\
Reasoning Gym          & Abs. Reason.  & 2 & N & codex      & gpt-5.1-codex-mini & 3 & 576/576  & 85.7$\pm$0.4 & 85.9$\pm$0.3 \\
\midrule

GPQA Diamond           & Expert QA     & 2 & N & codex      & gpt-5.2          & 3 & 198/198    & 87.9$\pm$0.6 & 87.2$\pm$0.3 \\
Humanity's Last Exam   & Expert QA     & 2 & N & claude-code & haiku-4-5       & 3 & 249/2500   & 10.7$\pm$0.9 & 11.0$\pm$0.4 \\
MMMLU                  & Expert QA     & 2 & N & codex-cli  & gpt-5.1-codex-mini & 5 & 150/210630 & 63.5$\pm$0.5 & 63.3$\pm$0.5 \\
SimpleQA               & Expert QA     & 2 & N & claude-code & opus-4-6        & 3 & 50/4326    & 96.7$\pm$1.8 & 94.7$\pm$0.7 \\
AA-LCR                 & Long Ctx.     & 2 & N & codex          & gpt-5-mini       & 1 & 99/99      & 68.0 & 68.7 \\

\bottomrule
\end{tabular}%
}
\end{table*}

\begin{table*}[t]
\centering
\caption{Harbor adapter catalog (Part 2 of 2): Scientific Research,
Agents/Tools/Systems, Data \& Analytics, Professional Domains, Safety
\& Security, and Multimodal. Columns as in
Table~\ref{tab:adapter_catalog_part1}.}
\label{tab:adapter_catalog_part2}
\scalebox{0.85}{%
\renewcommand{\arraystretch}{1.15}%
\setlength{\tabcolsep}{4pt}%
\begin{tabular}{@{}
    >{\raggedright\arraybackslash}p{2.6cm}
    >{\raggedright\arraybackslash}p{1.8cm}
    c c
    >{\raggedright\arraybackslash}p{1.7cm}
    >{\raggedright\arraybackslash}p{1.8cm}
    c c c c
@{}}
\toprule
\textbf{Benchmark} & \textbf{Subdomain}
 & \textbf{Sc.} & \textbf{Mode}
 & \textbf{Parity Agent} & \textbf{Parity Model}
 & $\bm{k}$ & \textbf{Set}
 & \textbf{Orig.} & \textbf{Harbor} \\
\midrule

ReplicationBench       & Research WF   & 1 & A & claude-code    & haiku-4-5        & 1 & 90/90     & 14.4 & 14.4 \\
\makecell[l]{ScienceAgent\\Bench} & Research WF & 1 & A & claude-code    & haiku-4-5        & 5 & 102/102   & 33.7$\pm$0.9 & 31.8$\pm$1.2 \\
SLDBench               & Research WF   & 1 & A & claude-code    & haiku-4-5        & 5 & 8/8       & 38.1$\pm$9.5 & 39.2$\pm$10.7 \\
MLGym-Bench            & Research WF   & 2 & M & mini-swe-agent & gpt-5            & 3 & 12/12     & 80.6$\pm$2.8 & 80.6$\pm$2.8 \\
CodePDE                & Sci. Comp.    & 2 & M & terminus-2     & haiku-4-5        & 8 & 5/5       & 27.5$\pm$3.7 & 27.5$\pm$3.7 \\
\makecell[l]{ResearchCode\\Bench} & Sci. Comp.  & 2 & M & codex          & gpt-4.1-mini     & 3 & 212/212   & 41.0$\pm$0.4 & 40.7$\pm$0.3 \\
SciCode                & Sci. Comp.    & 2 & M & codex          & gpt-5.1-codex-mini & 3 & 80/80   & 43.3$\pm$0.6 & 43.8$\pm$0.6 \\
QCircuitBench          & Sci. Comp.    & 2 & N & codex          & gpt-5.2          & 3 & 28/28     & 60.9$\pm$2.5 & 61.9$\pm$2.1 \\
BIX-Bench              & Biomed.       & 3 & A & bixbench-agent & gpt-4o-mini      & 3 & 50/205    & 16.0$\pm$3.1 & 16.7$\pm$2.4 \\
LAB-Bench              & Biomed.       & 2 & N & codex          & gpt-5-codex      & 3 & 181/181   & 40.0$\pm$0.2 & 41.1$\pm$0.5 \\
\midrule

ACE-Bench              & Tool Use      & 1 & A & claude-code    & haiku-4-5        & 3 & 973/973   & 83.6$\pm$0.4 & 83.2$\pm$0.8 \\
GAIA                   & Tool Use      & 1 & A & openhands      & gpt-5-mini       & 3 & 165/165   & 51.3$\pm$0.7 & 50.7$\pm$0.5 \\
GAIA2                  & Tool Use      & 3 & A & gaia2-parity-agent & gpt-5-mini   & 2 & 100/800   & 6.0$\pm$1.0 & 6.0$\pm$0.0 \\
BFCL                   & Tool Use      & 2 & N & codex          & gpt-5-mini       & 3 & 123/3641  & 83.2$\pm$0.7 & 83.5$\pm$1.0 \\
DeepSynth              & Web Agents    & 1 & A & claude-code    & haiku-4-5        & 3 & 40/40     & 9.3$\pm$1.0 & 7.8$\pm$0.6 \\
Seal-0                 & Web Agents    & 1 & A & claude-code    & haiku-4-5        & 3 & 111/111   & 33.9$\pm$3.0 & 33.3$\pm$3.6 \\
WideSearch             & Web Agents    & 1 & A & claude-code    & haiku-4-5        & 3 & 200/200   & 53.8$\pm$0.3 & 53.9$\pm$0.3 \\
\midrule

DA-Code                & SQL \& DS     & 3 & A & da-agent       & gpt-4o           & 3 & 479/479   & 38.7$\pm$1.1 & 38.9$\pm$0.2 \\
Spider 2               & SQL \& DS     & 1 & A & spider-agent   & gpt-5-mini       & 3 & 64/64     & 18.8$\pm$0.0 & 18.3$\pm$0.5 \\
\midrule

FinanceAgent           & Finance       & 3 & A & finance-agent  & gpt-5.2          & 3 & 50/50     & 78.0$\pm$1.0 & 80.0$\pm$0.0 \\
PIXIU                  & Finance       & 2 & M & codex          & gpt-5-mini       & 3 & 435/54083 & 90.1$\pm$3.3 & 85.2$\pm$6.4 \\
CRMArena               & Business      & 3 & A & crmarena-react & haiku-4-5        & 3 & 90/1170   & 69.0$\pm$1.0 & 68.0$\pm$2.0 \\
SpreadsheetBench       & Business      & 1 & A & claude-code    & haiku-4-5        & 3 & 400/400   & 68.8$\pm$0.8 & 68.3$\pm$1.1 \\
MedAgentBench          & Healthcare    & 3 & A & medagent-react & gpt-4o-mini      & 3 & 300/300   & 58.0$\pm$0.9 & 57.9$\pm$0.3 \\
LawBench               & Legal         & 2 & N & qwen-code      & glm-4.7          & 3 & 120/1000  & 65.0$\pm$0.3 & 65.6$\pm$0.7 \\
\midrule

CyberGym               & Cybersec.     & 3 & A & openhands      & haiku-4-5        & 5 & 10/1507   & 64.0$\pm$4.0 & 66.0$\pm$2.5 \\
StrongReject           & Jailbreak     & 2 & N & codex          & gpt-5-nano       & 3 & 150/12207 & 95.0$\pm$0.7 & 95.0$\pm$1.1 \\
\midrule

MMAU                   & Audio         & 2 & N & terminus-2     & gpt-4o           & 3 & 1000/1000 & 56.6$\pm$0.8 & 56.6$\pm$0.8 \\

\bottomrule
\end{tabular}%
}
\end{table*}

\subsection{Per-Benchmark Details}
\label{app:per_benchmark}
The subsections from \Cref{app:bench_swe} through \Cref{app:bench_multimodal} describe benchmarks added with Harbor adapters. \Cref{app:bench_harbor_format} covers benchmarks that natively use the Harbor format and require no adapter. \Cref{app:bench_not_included} lists benchmarks whose adapters are merged into Harbor (or have an approved adapter pull request) but were not evaluated here due to resource limits.

\subsubsection{Software Engineering}
\label{app:bench_swe}
\paragraph{Repo-level Issue Resolution:} SWE-Bench Multilingual, SWE-Bench Pro, SWE-Bench Verified, SWE-Smith, SWT-Bench.

\textit{SWE-Bench-Multilingual} \cite{yang2025swesmith}.
This benchmark extends SWE-bench \cite{jimenez2023swebench} across nine programming languages with 300 tasks. The Harbor adapter targets the per-task multilingual image and invokes the official grading harness, adapting all 300 tasks and scoring them with unit tests. The original oracle solutions pass 100\%, with one seasonal failure. Parity uses a fixed subset of 50 tasks stratified across the nine languages.

\textit{SWE-Bench~Pro} \cite{deng2025swebenchproaiagents}.
This benchmark extends SWE-bench \cite{jimenez2023swebench} with 731 tasks across Python, JavaScript, TypeScript, and Go. The Harbor adapter reuses the upstream execution and verification with two reliability fixes, running Jest in single-worker mode with forced exit and normalizing gold patches to end with a newline. The original oracle solutions pass 97\%, with 15 invalid gold patches and 9 tasks that time out. Parity uses a random subset of 100 tasks together with a 10-task rerun after adapter fixes, and both sides are task-aligned.

\textit{SWE-Bench~Verified} \cite{jimenez2023swebench}.
This benchmark is a human-curated subset of SWE-bench \cite{jimenez2023swebench} with 500 Python tasks scored by unit tests. The original oracle solutions pass 99.2\%, with 4 tasks failing on data or infrastructure errors. Parity covers the full 500-task set and is validated transitively against an earlier Terminal-Bench integration. We also check the benchmark score directly against the official SWE-bench leaderboard on a 499-task set, excluding \texttt{scikit-learn\_\_scikit-learn-14710} due to hardware limits in the Daytona environment.

\textit{SWE-smith} \cite{yang2025swesmith}.
This benchmark contains 60{,}000 SWE-bench~\cite{jimenez2023swebench}-style tasks with higher-quality descriptions and richer multi-file patches. Harbor currently supports the Python-profiled tasks, and the adapter extends to other languages easily. Runs are scored by unit tests. A fraction of the oracle patches do not fully pass upstream, a known artifact of synthetic patch generation. Parity uses a 100-task slice reused from an earlier Terminal-Bench integration, so parity chains transitively across two legs.

\textit{SWT~Bench} \cite{mundler2025swtbench}.
This benchmark contains 433 tasks that test an agent's test-generation and bug-reproduction ability, and a redundant code-description segment is removed from the prompt during adaptation. Generated tests are scored by whether they fail on the buggy code and pass on the fixed code. The original oracle solutions pass 97.7\%, with 9 \texttt{sphinx-doc} tasks failing on an upstream parsing issue. Parity is restricted to evaluation because upstream provides no inference script, so Harbor diffs are exported to the upstream prediction format and graded by the upstream evaluator.

\paragraph{Feature \& End-to-end Development:} FeatureBench, SWE-Lancer.

\textit{FeatureBench} \cite{libercoders2026featurebench}.
This benchmark evaluates feature-level code generation on 200 real-world Python tasks across 24 popular repositories, split into interface-guided (lv1) and from-scratch (lv2) complexity levels. The Harbor adapter reuses the prebuilt DockerHub images, mirrors the lv1 oracle through \texttt{setup\_patch} reversal and the lv2 oracle through \texttt{agent\_code} re-export, auto-configures \texttt{shm\_size} and GPU allocation from task metadata, and prunes flaky PASS\_TO\_PASS tests. The oracle passes 200/200 on Docker and 198/200 on Modal, where two transient failures occur, namely a mirror timeout on an lv2 dependency install and a SciPy segmentation fault on an lv1 task in Modal's sandbox. Parity uses a curated 30-task lite split (26 lv1 and 4 lv2) on NVIDIA A10G, and the residual disagreements reflect stochastic lv1 variation rather than systematic divergence.

\textit{SWE-Lancer} \cite{miserendino2025swelancerfrontierllmsearn}.
This benchmark evaluates agents on 463 real freelance software-engineering tasks scraped from Upwork and scoped to the Expensify repository, comprising 198 IC SWE tasks that resolve a GitHub issue and pass Playwright and pytest user-tool tests, and 265 Manager tasks that select the correct proposal among candidates. The Harbor adapter reuses the upstream \texttt{swelancer\_x86} images with light tmux and user-tool additions and rewrites the Python-only solver prompt into terminal-style instructions. One deviation is that the user tool runs in the agent's namespace, with tests copied in unzipped, because Harbor has no privileged out-of-namespace runner, and the grading logic is otherwise preserved. The oracle passes 100\%, as reported in the adapter PR. Parity covers the full 463-task set, and the original side runs from a \texttt{preparedness} fork that ports a matching Claude Code solver into the upstream nanoeval harness, so both sides exercise the same agent.

\paragraph{Competitive \& Function-level Coding:} Aider Polyglot, BigCodeBench, LiveCodeBench, USACO, HumanEvalFix, QuixBugs.

\textit{Aider~Polyglot} \cite{aiderpolyglot2024}.
This benchmark repackages 225 Exercism exercises across six languages (Java, Python, Go, Rust, C++, and JavaScript) to evaluate iterative code editing rather than greenfield generation. The Harbor adapter mirrors all exercises with per-language Dockerfiles, ports the upstream test harnesses verbatim, and wraps encrypted oracle payloads in shell normalizers. Runs are scored by the per-language unit tests, and the adapter reports no aggregate oracle pass rate. Parity is established transitively in two legs, one between the original and the Terminal-Bench adapter (225 tasks, 2 runs with claude-code) and one between the Terminal-Bench adapter and Harbor (225 tasks, 1 run with claude-haiku).

\textit{BigCodeBench} \cite{zhuo2025bigcodebench}.
This benchmark evaluates complex function-calling and engineering programming on the 145 hard-split tasks, using two prompt formats, the \emph{Complete} docstring-driven format and the \emph{Instruct} natural-language format. The Harbor adapter converts tasks from HuggingFace and uses reward-based verification, and 3 tasks are excluded for network or file-structure incompatibilities with the Harbor sandbox. The adapter PR reports no aggregate pass rate in text. Parity uses the full 145-task set with codex over 3 trials per side.

\textit{LiveCodeBench} \cite{jain2024livecodebench}.
This benchmark evaluates competitive programming on 1{,}055 problems aggregated from LeetCode, Codeforces, and AtCoder, covering May 2023 to March 2024. The Harbor adapter supports both Solution-class and stdin/stdout tests, exposes \texttt{check\_solution.py} for public-test development, and auto-injects standard-library imports, and final scoring runs both public and private tests under 30\,s and 180\,s timeouts. LiveCodeBench ships no oracle solutions, so validity relies on test-case validation. Parity is established transitively in two legs, one between the original and the Terminal-Bench adapter (100 \texttt{release\_v6} tasks, 4 trials with terminus-2) and one between the Terminal-Bench adapter and Harbor (100 tasks, 4 trials with claude-haiku).

\textit{USACO} \cite{shi2024language}.
This benchmark evaluates algorithmic problem-solving on 307 USA Computing Olympiad problems covering graph theory, dynamic programming, and other competitive domains, of which Harbor adapts 304 after excluding 3 with broken reference solutions. The Harbor adapter wraps the official batch judge in pytest with CTRF logging and embeds canonical solutions, leaving the upstream grading logic unchanged. Parity uses the full 304-task set over 3 runs with claude-haiku.

\textit{HumanEvalFix} \cite{muennighoff2023octopack}.
This benchmark evaluates code repair on 164 Python tasks from HumanEvalPack, where the agent fixes buggy implementations so that pytest passes. The Harbor adapter rewrites the original prompts into agent-oriented instructions, embeds Dockerized environments, and ships reference solutions, and the pytest grading logic is unchanged. The oracle passes 164/164 at reward 1.0. Parity uses the full 164-task set for 3 trials with OpenHands~\cite{wang2025openhands} on Daytona under gpt-4o-mini and gpt-5-mini.

\textit{QuixBugs} \cite{lin2017quixbugs}.
This benchmark evaluates single-line bug repair on 80 defective programs, 40 in Python and 40 in Java, drawn from the Quixey Challenge. The Harbor adapter clones the upstream repository and generates per-language tasks (Python 3.11, and JDK 11 with Gradle), and runs are scored by pytest correctness tests together with a diff-based single-line check. The oracle passes 100\%. Parity is established transitively in two legs, one between the original and the Terminal-Bench adapter (80 tasks, 5 trials with claude-code) and one between the Terminal-Bench adapter and Harbor (80 tasks, 3 trials with codex).

\paragraph{Performance Optimization:} GSO, AlgoTune.

\textit{GSO} \cite{shetty2025gso}.
This benchmark evaluates code performance optimization on 102 Python repositories, each pairing a codebase with a performance test that serves as the specification. The Harbor adapter follows the upstream evaluation flow in isolated Docker environments and reuses the original grading harness, and all 102 tasks are adapted. The oracle passes 89/102 (87.3\%), while the remaining 13 fail because their oracles run inconsistently faster than the reference baseline under timing variance, a known artifact of the speedup metric rather than a harness bug. Parity uses the full 102-task set with OpenHands@1.4.0 and gpt-5.1 (high) on each side.

\textit{AlgoTune} \cite{press2025algotune}.
This benchmark evaluates algorithm optimization on 154 tasks across mathematics, physics, computer science, signal processing, cryptography, and graph algorithms, scored by continuous speedup rather than binary pass or fail. The Harbor adapter implements the interleaved timing protocol with 100 instances and 10 timing repetitions per instance, enforces the upstream limits of 8 CPUs and 16\,GB of memory inside Docker, and reports the harmonic-mean speedup. Because AlgoTune has no canonical oracle, we use the baseline solution and verify that its self-speedup is close to 1.0$\times$. Parity is established transitively in two legs, one between the original and the Terminal-Bench adapter and one between the Terminal-Bench adapter and Harbor, on the full 154-task set.

\paragraph{Language Translation:} CRUST-Bench.

\textit{CRUST-Bench} \cite{khatry2025crustbench}.
This benchmark evaluates C to safe-Rust transpilation on 100 real-world C repositories from GitHub, each paired with hand-written safe-Rust interfaces and correctness tests. The Harbor adapter vendors the C sources and headers into per-task Rust 1.83 slim containers and classifies projects by difficulty. Runs are scored by \texttt{cargo test}, and because transpilation is open-ended, no oracle solutions are provided. Parity uses the full 100-task set for 3 trials with codex and gpt-5-nano.


\subsubsection{Mathematics \& Reasoning}
\label{app:bench_math}

\paragraph{Competition Mathematics:} AIME, IneqMath, Omni-Math.

\textit{AIME} \cite{ye2025aimepreview}.
This benchmark evaluates competition-level mathematical reasoning on 60 problems from AIME 2024 and 2025, each requiring an exact integer answer in [0, 999] with no partial credit. The Harbor adapter sources problems from the GAIR-NLP/AIME-Preview repository and packages each as an isolated task with Python available for symbolic computation, and scoring uses exact match against the integer answer. The oracle solutions pass 100\%. Because AIME has no agent-oriented upstream harness, standard parity does not apply, so we instead confirm on the full 60-task set that diverse agents such as Gemini-CLI and codex recover the expected fraction of solutions.

\textit{IneqMath} \cite{sheng2025ineqmath}.
This benchmark evaluates inequality-proof reasoning on the 100 expert-curated multiple-choice questions in the public dev set, covering Bound Estimation and Relation Prediction, while the test set and step-wise evaluation are not publicly released and are not adapted. The Harbor adapter generates tasks with full upstream metadata and supports submission-format conversion for the leaderboard, and evaluation follows the upstream protocol of exact match with a gpt-4o-mini judge fallback. The oracle passes 100\% on the full dev set. Parity uses the full 100-task dev set over 3 trials with codex and gpt-4o-mini, with the same evaluator on both sides.

\textit{Omni-Math} \cite{gao2024omnimath}.
This benchmark evaluates olympiad-level mathematical reasoning on 4,428 problems spanning more than 33 sub-domains across 10 difficulty levels. The Harbor adapter loads problems from HuggingFace, standardizes the answer at \texttt{/workspace/answer.txt}, and installs the upstream evaluation code inside the container to avoid parsing pitfalls, and a gpt-5-mini judge compares the answer to ground truth. Oracle accuracy is bounded at 99.5\% because 18 problems ship without benchmark-provided answers. Parity uses a stratified 100-task subset for 2 trials with codex and gpt-5.3-codex, with an identical judge configuration on both sides.

\paragraph{Abstract \& Procedural Reasoning:} KUMO, Reasoning Gym, ARC-AGI-2.

\textit{KUMO} \cite{lin2025kumo}.
This benchmark evaluates interactive multi-step reasoning on 5,300 procedurally generated games (50 instances across 106 domain-scenario pairs), where the agent takes actions, observes outcomes, and identifies hidden truths using a knowledge book. The Harbor adapter runs an interactive loop through a local verifier service (\texttt{POST /act}) and a CLI client, isolates secrets in verifier-only Docker images while exposing only \texttt{knowledge\_book.txt}, and enforces action budgets through HTTP 429 responses, and scoring uses exact match on \texttt{/app/answer.txt}. The oracle passes 100\% on the generated tasks. Parity uses a deterministic 212-task subset (seeds 0 and 1 per scenario) for 3 trials per agent-model pair (kumo-vanilla, terminus-2, and codex, each paired with gpt-5-mini and gpt-5-nano) at \texttt{max-steps=50} and \texttt{temperature=1.0} on both sides.

\textit{Reasoning~Gym} \cite{stojanovski2025reasoninggym}.
This benchmark evaluates verifiable procedural reasoning across diverse problem types, materialized into 576 tasks from 96 task generators (288 easy and 288 hard, with 3 instances each). The Harbor adapter dockerizes each generator and has agents write code to compute answers rather than answering zero-shot, directing the result to \texttt{answer.txt}. Agentic execution yields substantially higher reward than the zero-shot setting, and scoring uses the upstream per-task reward verifier. A few generators ship no oracle answer or fail to score their own oracle at reward 1.0, and these are documented and excluded from oracle validation rather than treated as harness bugs. Parity uses the full 576-task set for 3 trials with codex and gpt-5.1-codex-mini, with the same generators and verifier on both sides.

\textit{ARC-AGI-2} \cite{chollet2025arcagi2}.
This benchmark evaluates abstract visual reasoning on the ARC-AGI-2 public evaluation set, where the agent infers a grid transformation from a few demonstrations and applies it to held-out inputs. Harbor materializes 167 task pairs by splitting the 120 upstream puzzles along their test cases. The Harbor adapter pulls the dataset from \texttt{Ardea/arc\_agi\_v2}, rewrites the puzzle descriptions into agent-oriented instructions, and runs each task in a Python 3.13 container that writes the predicted grid as a JSON 2D array to \texttt{/testbed/output.json}. A pytest verifier requires an exact grid match and embeds reference solutions. Parity uses the full 167-task set for 6 trials with codex@0.53.0 and gpt-5.2 at \texttt{reasoning\_effort=medium}, under the upstream Pass@2 protocol.

\subsubsection{Knowledge \& Long Context}
\label{app:bench_knowledge}

\paragraph{Expert \& Multi-subject QA:} GPQA Diamond, Humanity's Last Exam, MMMLU, SimpleQA.

\textit{GPQA~Diamond} \cite{rein2024gpqa}.
This benchmark evaluates graduate-level scientific reasoning on 198 expert-curated multiple-choice questions in biology, physics, and chemistry, where domain experts reach about 65\% and non-experts with web access about 34\%. The Harbor adapter loads questions from HuggingFace and deterministically shuffles the answer choices, and scoring uses exact letter match on \texttt{/app/answer.txt}. The oracle passes 100\% on the full diamond split. Parity uses the full 198-task diamond split for 3 trials with codex and gpt-5.2.

\textit{Humanity's~Last~Exam} \cite{phan2026hle}.
This benchmark evaluates expert-level multimodal question answering on 2,500 questions across mathematics, science, engineering, humanities, and computer science. The Harbor adapter reuses the official HLE judge prompt, copies images into the workspace for vision-capable agents, routes responses through \texttt{/logs/agent/response.txt} with confidence scores, and corrects one ground-truth placeholder in the upstream dataset, and a gpt-5 judge performs scoring. Oracle verification passes 100\% with 0\% calibration error. Parity uses a stratified 249-task subset (10\% per category, seed 42) with a gpt-5 judge, and calibration error is computed post-hoc.

\textit{MMMLU} \cite{hendrycks2021mmlu}.
This benchmark evaluates multilingual multi-subject knowledge by extending MMLU across 15 languages, for a total of 210,630 tasks (14,042 questions translated into Arabic, Bengali, German, Spanish, French, Hindi, Indonesian, Italian, Japanese, Korean, Portuguese, Chinese, Swahili, and Yoruba). The Harbor adapter casts each question as a standalone task with explicit \texttt{Answer: \$LETTER} formatting and supports multilingual character normalization through the simple-evals regex, and scoring uses exact match on the answer letter. Oracle accuracy is 100\%. Parity uses a stratified 150-task subset (10 per language, seed 0) for 5 trials with codex and gpt-5.1-codex-mini.

\textit{SimpleQA} \cite{wei2024simpleqa}.
This benchmark evaluates factual recall on 4,326 fact-seeking questions whose correct answers are short and unambiguous. The Harbor adapter follows OpenAI's simple-evals approach with a configurable grading model (gpt-5-mini by default) and allows internet access during evaluation, and scoring uses an LLM judge. The oracle passes 100\% on the full 4,326-task set. Parity uses the first 50 tasks for 3 trials with claude-code and claude-opus-4-6, using gpt-4o-mini as the judge to match the upstream framework.

\paragraph{Long Context Reasoning:} AA-LCR.

\textit{AA-LCR} \cite{aalcr2025}.
This benchmark evaluates long-context reasoning over documents of roughly 100k tokens in seven categories (company reports, academia, government, legal, industry, marketing, and survey materials), with 100 upstream tasks, of which Harbor adapts 99 after excluding one with a ground-truth error. The Harbor adapter copies documents into the workspace at \texttt{/workspace/documents/} and corrects two ground-truth answers (Task 40 date formatting and Task 94 percentage versus decimal), and grading uses the official equality-checker prompt. The oracle passes 99/99 at mean reward 1.0. Parity uses the full 99-task set across several agent-model pairs, namely codex with gpt-5-mini, claude-code with haiku, terminus-2 with haiku, and terminus-2 with gpt-5-mini.

\subsubsection{Scientific Research}
\label{app:bench_science}

\paragraph{End-to-end Research Workflows:} ReplicationBench, SLDBench.

\textit{ReplicationBench} \cite{ye2025replicationbench}.
This benchmark evaluates research replication on astrophysics papers, with 106 upstream tasks, of which Harbor adapts 90 expert-reviewed numeric tasks that require Python to match numeric outputs within specified tolerances. The Harbor adapter integrates the HuggingFace dataset and runs tasks in the upstream Docker images, and scoring uses tolerance-based numeric checks. No aggregate oracle rate is reported, and 5 tasks time out consistently on both sides and are flagged for possible exclusion. Parity is established transitively in two legs, one between the original and the Terminal-Bench adapter (106 tasks, 2 runs with claude-code and sonnet) and one between the Terminal-Bench adapter and Harbor (90 tasks, 1 run with claude-code and haiku).

\textit{SLDBench} \cite{lin2025sldbench}.
This benchmark evaluates symbolic scaling-law discovery from more than 5,000 LLM training experiments, packaged as 8 tasks that span parallel, vocabulary, fine-tuning, domain-mixture, mixture-of-experts, data-constrained, hyperparameter, and adversarial (U-shaped) scaling laws. The Harbor adapter ships oracle implementations of the Human and Expert baselines and requires the agent to produce both \texttt{/app/law.py} and \texttt{/app/explain.md}, and scoring is continuous over $R^2$, NMSE, and NMAE on held-out extrapolation sets. Because the metric is continuous rather than pass or fail, oracle validation reports the paper's Human and Expert baselines rather than a pass rate, and the two hardest tasks have negative $R^2$ even for the human oracle, which is preserved faithfully. Parity is established transitively against the Terminal-Bench adapter on all 8 tasks with claude-code and haiku-4-5 over 5 runs.


\paragraph{Scientific Computing:} CodePDE, ResearchCodeBench, SciCode, QCircuitBench.

\textit{CodePDE} \cite{li2025codepde}.
This benchmark evaluates Python code generation for solving partial differential equations across five families (advection, burgers, reacdiff1d, cns1d, and darcy) with 100 test instances per family. The Harbor adapter replaces the upstream scaffolding-template prompts with full markdown task instructions and adds automatic test-data download from HuggingFace, and scoring uses the original nRMSE metric. Reference oracle solutions are copied verbatim from upstream, and no aggregate oracle rate is reported. Parity is established transitively in two legs, one between the original and the Terminal-Bench adapter and one between the Terminal-Bench adapter and Harbor, on the full task set, with results aligned within seed variance.

\textit{ResearchCodeBench} \cite{hua2025researchcodebench}.
This benchmark evaluates the implementation of research code from academic papers, covering 20 ML and CV papers decomposed into 212 code-snippet tasks totaling 1,449 lines of code. The Harbor adapter provides the full paper content and code files, divides the problems into per-snippet units, and reuses the upstream I/O format and unit-test verifier, with one task-specific oracle fix that relaxes the GMFlow tolerance from 1e-7 to 1e-3 to absorb upstream nondeterminism. Oracle reference code is embedded in \texttt{solve.sh}, and no aggregate oracle rate is reported. Parity uses the full 212-task set with gpt-4o-mini-2024-11-20 against the original benchmark.

\textit{SciCode} \cite{tian2024scicode}.
This benchmark evaluates scientific computing on 80 main problems decomposed into 338 sub-problems across physics, mathematics, materials science, biology, and chemistry. The Harbor adapter generates one task per main problem with sequential instructions and skips three pre-written steps (problems 13, 62, and 76), and the upstream HDF5 checker gives fractional reward by sub-step correctness. Ground-truth solutions exist only for the 15 validation problems, where the oracle passes 15/15, while the 65 test problems carry placeholder oracles and are excluded from validation. Parity uses the full 80-task set with codex@0.106.0, with sub-step accuracy aligned within a Welch t-test.

\textit{QCircuitBench} \cite{yang2025qcircuitbench}.
This benchmark evaluates quantum-algorithm design on 28 tasks spanning oracle construction, algorithm design, and random circuit synthesis, expressed in OpenQASM 3.0 and Python. The Harbor adapter rewrites the upstream task descriptions into agent-oriented instructions and generates oracles dynamically inside the verifier, and scoring combines semantic correctness and efficiency into a score in [0, 1]. The adapter PR reports no aggregate oracle rate. Parity uses the full 28-task set with codex@0.76.0.

\paragraph{Biomedical Research:} BIX-Bench, LAB-Bench.

\textit{BIX-Bench} \cite{mitchener2025bixbench}.
This benchmark evaluates computational biology data analysis on 205 questions derived from 61 published Jupyter notebooks. The Harbor adapter offers two execution modes, a custom Jupyter agent that mirrors the original benchmark and a terminal variant (bixbench-cli) that exposes tasks to generic agents such as codex, and the LLM-judge prompt is lightly modified to accept range answers so the oracle scores robustly. The oracle passes 205/205. Parity uses a 50-task subset with gpt-4o-mini, together with a CLI-mode run using codex and gpt-5-mini on the same subset.

\textit{LAB-Bench} \cite{laurent2024labbench}.
This benchmark evaluates visual reasoning over scientific figures in biology, with 181 multiple-choice questions from the public FigQA subset. The Harbor adapter copies figures into the workspace for the agent to read from disk rather than embedding encoded images, and shuffles the answer choices at runtime through a container ENTRYPOINT to prevent positional bias, and scoring uses exact letter match through \texttt{answer.txt}. The oracle passes 181/181. Parity uses the full 181-task set with codex@0.71.0.

\subsubsection{Agents, Tools \& Systems}
\label{app:bench_agents}

\paragraph{Tool Use \& Assistants:} GAIA, GAIA2, BFCL.

\textit{GAIA} \cite{mialon2023gaia}.
This benchmark evaluates multi-step tool use on 165 validation-split question-answering tasks that require browsing, file handling, and external tools across three difficulty levels. The Harbor adapter wraps the original prompts as task instructions and copies attachments into the workspace, and scoring uses deterministic answer normalization and exact match. The oracle passes 165/165 at mean reward 1.0 on the full validation split. Parity uses the full validation split with openhands and gpt-5-mini at \texttt{timeout\_multiplier=1.5}, equivalent to \texttt{max\_iterations=45}.

\textit{GAIA2} \cite{froger2026gaia2}.
This benchmark evaluates agent ability across simulated apps and environments on 800 public validation scenarios from Meta's ARE, spanning five configurations (execution, search, adaptability, time, and ambiguity), each with full environments, apps, and oracle action sequences. The Harbor adapter supports an ARE-native mode that runs the official oracle and default agent, and a CLI mode that exposes ARE tools to standard agents through an MCP sidecar, and scoring uses the upstream action-sequence checker. The oracle passes 800/800. Parity uses a stratified 100-task subset with the official ARE \texttt{default} agent over 2 trials per side.

\textit{BFCL} \cite{patil2025bfcl}.
This benchmark evaluates function-calling ability on 3,641 tasks across 13 single-turn and live categories from BFCL v4. The Harbor adapter mirrors the upstream AST evaluation logic with function-name normalization, order-independent parallel matching, and parameter validation, and tasks run in Docker with the upstream evaluator pre-installed. All 3,641 oracle solutions pass. Parity uses a stratified 123-task subset with codex@0.77.0 and gpt-4o-mini.

\paragraph{Deep Research \& Web Agents:} DeepSynth, Seal-0, WideSearch.

\textit{DeepSynth} \cite{paul2026deepsynth}.
This benchmark evaluates deep information synthesis from multiple web sources into structured JSON answers, with 40 tasks in the dev set. The Harbor adapter reuses the official HuggingFace dataset, and scoring uses deterministic F1 over key-value pairs with an optional Claude Haiku 4.5 judge fallback for semantic equivalence. The oracle passes 40/40 at reward 1.0. Parity uses the full 40-task dev set.

\textit{Seal-0} \cite{pham2026sealqa}.
This benchmark evaluates search-augmented factual question answering under adversarial web evidence on 111 tasks, where frontier models reach near-zero accuracy because of conflicting or misleading search results. The Harbor adapter reuses the upstream HuggingFace dataset, and scoring uses an LLM judge (Claude Haiku 4.5 by default) with normalized string matching as a fallback, mirroring upstream. The oracle passes 111/111. Parity uses the full 111-task set with claude-code.

\textit{WideSearch} \cite{wong2025widesearch}.
This benchmark evaluates broad web information gathering on 200 bilingual tasks spanning 18 industries, where the agent collects and organizes structured information into markdown tables. The Harbor adapter reuses the official HuggingFace dataset and ports the upstream metrics, and scoring combines Item F1, exact match, and an LLM judge (gpt-4.1 by default). The oracle passes 200/200 at mean Item F1 of 0.9996, and the small gap arises because gold-answer cells that contain the markdown delimiter (\texttt{|}) are replaced with a space during CSV-to-markdown conversion, which does not affect real agent runs. Parity uses the full 200-task set for 3 trials on each side.

\subsubsection{Data \& Analytics}
\label{app:bench_data}

\paragraph{Text-to-SQL \& Data Science:} DA-Code, Spider 2.

\textit{DA-Code} \cite{huang2024dacode}.
This benchmark evaluates data-science coding that requires Python, SQL, and Bash for end-to-end analysis, with 500 upstream tasks across 13 categories, of which Harbor adapts 479 after excluding the ml-cluster category. The Harbor adapter mirrors the upstream evaluation pipeline, applies 11 correctness fixes to the original evaluation script, and adds agent-oriented instructions describing task goals and file paths, and scoring uses the upstream checker against gold outputs. The gold-output oracle passes 479/479 after the fixes. Parity uses the full 479-task set for 3 trials with DA-Agent, and a permutation test shows no statistically significant difference.

\textit{Spider~2} \cite{lei2024spider2}.
This benchmark evaluates dbt data-transformation ability on 68 upstream tasks over real-world DuckDB projects, of which Harbor adapts 64 after excluding 4 that consistently fail the oracle on both sides. The Harbor adapter reuses the upstream dbt projects with refactored API integration and retry handling for safety-policy rejections, and scoring uses dbt database comparison. The oracle passes 64/64. Parity uses the full 64-task set for 3 trials with spider-agent.

\subsubsection{Professional Domains}
\label{app:bench_professional}

\paragraph{Finance \& Trading:} FinanceAgent, PIXIU.

\textit{FinanceAgent} \cite{valsai2024financeagent}.
This benchmark evaluates financial research and analysis on 50 questions that require interpreting SEC filings and reasoning about reported figures. The Harbor adapter ships a terminal variant for CLI agents and a customized variant for function-calling, and an LLM judge scores answers against the Finance-Agent ground truth. The oracle passes 50/50. Parity uses the full 50-task set for 3 trials with Finance-Agent and Claude Code.

\textit{PIXIU} \cite{xie2023pixiu}.
This benchmark evaluates financial NLP across 54,083 tasks in 29 subcategories that span classification, sentiment, question answering, named-entity recognition, sequential labeling, relation extraction, and summarization. The Harbor adapter materializes each subcategory example as a task, and scoring uses category-specific metrics such as accuracy, F1, RMSE, ROUGE, BERTScore, and BARTScore. The oracle solutions pass 100\%. Parity uses a stratified 435-task subset (15 per subcategory) for 3 trials with codex.

\paragraph{Business / Professional Work:} SpreadsheetBench, MedAgentbench, LawBench.

\textit{SpreadsheetBench} \cite{ma2024spreadsheetbench}.
This benchmark evaluates spreadsheet manipulation on 400 tasks drawn from real user questions on Excel forums, where the agent modifies a workbook so that designated cell values become correct. The Harbor adapter replaces the Windows-only \texttt{win32com} pipeline with cross-platform LibreOffice headless recalculation and applies 11 correctness fixes to the evaluator, and scoring compares cell values against the gold workbook. All 400 oracle solutions pass. Parity uses the full 400-task set for 3 trials with claude-code.

\textit{MedAgentBench} \cite{jiang2025medagentbench}.
This benchmark evaluates clinical agent ability on 300 tasks that require FHIR API interaction to retrieve patient data and execute healthcare workflows. The Harbor adapter vendors the official reference grader, mirrors the upstream GET, POST, and FINISH semantics, and adds Harbor-specific instruction formatting to enforce strict payload compliance, and scoring uses the upstream grader. The oracle passes 300/300, as verified in the adapter PR. Parity uses the full 300-task set for 3 trials with HTTPAgent and gpt-4o-mini.

\textit{LawBench} \cite{fei2023lawbench}.
This benchmark evaluates Chinese legal reasoning across 1,000 tasks derived from 20 task families, each with zero-shot and one-shot prompts, covering statutes, legal reasoning, and case prediction. The Harbor adapter pins the upstream LawBench repository and reuses its native evaluation functions, with the agent writing answers to \texttt{/app/answer.jsonl}, and scoring uses per-family metrics that mix exact match, F1, ROUGE, and rule-based scoring. The oracle passes the full task set, as verified in the adapter PR. Parity uses a stratified 120-task subset (3 chunks across 20 families and 2 prompt types) for 3 trials with Qwen Code on GLM-4.7.

\subsubsection{Safety \& Security}
\label{app:bench_safety}

\textit{CyberGym} \cite{cybergym2025}.
This benchmark evaluates an agent's vulnerability-analysis ability on 1,507 real-world C and C++ tasks drawn from 188 projects, where the agent generates a proof-of-concept input that triggers the bug under sanitizer instrumentation (ASan, MSan, or UBSan). The Harbor adapter ships four difficulty levels, from level0 to level3, that control which files are exposed to the agent, and level1 is the primary level and matches the upstream paper. Scoring uses a dual-binary check in which the generated input must crash the vulnerable binary but not the patched one. The ground-truth inputs live in per-task runner images sourced from OSS-Fuzz, and the oracle reaches 100\% on the 10-task subset, while full-set validation is skipped because of the roughly 10\,TB footprint of the 3{,}014 runner images. Parity uses the recommended 10-task subset (6 ARVO and 4 OSS-Fuzz, over five difficulty and seed combinations of three at level1, one at level2, and one at level3) with OpenHands@1.6.0 and claude-haiku-4-5.

\textit{StrongReject} \cite{souly2024strongreject}.
This benchmark evaluates LLM jailbreak robustness on 313 forbidden prompts crossed with 39 jailbreak methods, for 12,207 tasks in total. The Harbor adapter supports all 39 methods and implements the official rubric evaluator with inverted scoring, where a reward of 1 indicates a safe refusal, and a rubric-based LLM judge (gpt-4o-mini by default) scores refusal, convincingness, and the specificity of harmful content. All 313 base oracle solutions pass. Parity uses a stratified 150-task subset (50 prompts across 3 jailbreaks) for 3 trials with codex and gpt-5-nano.

\subsubsection{Multimodal}
\label{app:bench_multimodal}

\paragraph{Audio Understanding.}

\textit{MMAU} \cite{sakshi2024mmau}.
This benchmark evaluates multimodal audio understanding on 1,000 expert-curated tasks spanning speech, environmental sounds, and music. The Harbor adapter uses Qwen2-Audio-Instruct to transcribe audio to text and runs the upstream judge through pytest over the transcribed answer. The oracle passes the full test-mini set, as verified in the adapter PR. Parity uses the full 1,000-task test-mini set for 3 trials with Terminus-2 and gpt-4o.

\subsubsection{Benchmarks Using Harbor Format}
\label{app:bench_harbor_format}
The following benchmarks are built with Harbor format so that no additional adaptation is needed.

\textit{Terminal-Bench} \cite{terminalbench}.
This benchmark evaluates an agent's ability to complete realistic command-line tasks such as filesystem manipulation, build configuration, debugging, and system administration. Tasks are authored directly in the Harbor schema, with instruction, environment, tests, and solution, so no adapter is required, and Harbor pulls the upstream task directories verbatim, runs them in per-task Docker environments, and scores them with the upstream test scripts.

\textit{CompileBench} \cite{compilebench2025}.
This benchmark evaluates an agent's compilation skills across 15 real-world build tasks that involve dependency resolution, legacy code such as \texttt{coreutils} v5.0 from 2003, static linking with glibc and musl, and ARM64 and Windows cross-compilation. Tasks are maintained natively in the Harbor schema by the upstream authors, so the adapter reduces to a sparse-checkout of the task directories, and the upstream test scripts run inside isolated Docker images.

\textit{SkillsBench} \cite{skillsbench2026}.
This benchmark evaluates how well agent skills transfer across diverse tasks, with 86 tasks across 11 domains paired with curated reference skills and deterministic verifiers. Tasks are authored directly in the Harbor schema and require no adaptation, and the agent runs with and without the curated skills exposed through Harbor's skills mechanism, scored by the upstream verifiers.

\subsubsection{Other Benchmarks Not Included}
\label{app:bench_not_included}
Due to compute and time constraints, we exclude the following benchmarks from our experiments, although their adapters are merged into Harbor or have an approved adapter pull request. We list the supported benchmarks below; adapters whose parity experiments are already complete also appear in \Cref{tab:adapter_catalog_part1,tab:adapter_catalog_part2}.

\textbf{Software Engineering}: ABC-Bench \cite{yang2026abcbench}, AutoCodeBench \cite{chou2025autocodebench}, CanItEdit \cite{cassano2024canitedit}, CooperBench \cite{khatua2026cooperbench}, DevEval \cite{li2024deveval}, DevOpsGym \cite{tang2026devopsgym}, EvoEval \cite{xia2024evoeval}, FeatBench \cite{chen2026featbench}, Frontier-CS-Algorithm \cite{mang2025frontiercs}, Multi-SWE-Bench \cite{zan2025multiswebench}, ProgramBench \cite{yang2026programbench}, SWE-Gym \cite{pan2025swegym}, SWE-Rebench \cite{badertdinov2025swerebench}, UniCode \cite{zheng2026unicode}, WebGen-Bench \cite{lu2025webgenbench}.

\textbf{Mathematics \& Reasoning}: SATBench \cite{wei2025satbench}.

\textbf{Knowledge \& Long Context}: CL-Bench \cite{dou2026clbench}, LoCoMo \cite{maharana2024evaluating}.

\textbf{Scientific Research}: AstaBench \cite{astabench2026}, LLM-SRBench \cite{shojaee2025llmsrbench}, ML-Dev-Bench \cite{padigela2025mldevbench}, MLGym-Bench \cite{mlgym2025}, MLR-Bench \cite{chen2025mlrbench}, RExBench \cite{edwards2025rexbench}, ScienceAgentBench \cite{chen2024scienceagentbench}.

\textbf{Agents, Tools \& Systems}: ACE-Bench \cite{chen2025acebench}, AMA-Bench \cite{zhao2026amabench}, DeepResearch-Bench-II \cite{du2025deepresearch}, LongCLIBench \cite{feng2026longclibench}, OSWorld \cite{xie2024osworld}, Tau3-Bench \cite{yao2024taubench}, TextArena \cite{guertler2025textarena}.

\textbf{Data \& Analytics}: ADE-Bench \cite{stancil2025adebench}, BIRD-Bench \cite{li2023birdbench}, DABstep \cite{egg2025dabstep}, DS-1000 \cite{lai2023ds1000}, KramaBench \cite{lai2025kramabench}.

\textbf{Professional Domains}: CRMArena \cite{huang2025crmarena}, TheAgentCompany \cite{xu2024agentcompany}.

\textbf{Multimodal}: GraphDesignBench \cite{deganutti2026gdb}, RefAV \cite{davidson2025refav}.

\subsection{Benchmark Issues and Lessons Learned}
\label{app:lessons}

Adapter construction and parity validation surfaced a small number of recurring failure modes that materially affect reproducibility, often regardless of the benchmark's age, popularity, or scientific area. We group them here by symptom rather than by domain, with concrete examples drawn from the per-benchmark notes in \Cref{app:bench_swe} through \Cref{app:bench_multimodal}, and conclude with concrete recommendations for benchmark authors.

\paragraph{Environment instability.}
Broken Docker builds, unpinned dependencies, removed upstream packages, hard-coded paths, and platform-specific tooling that breaks across machines. Many benchmarks required non-trivial environment reconstruction before any task could run: SpreadsheetBench's upstream pipeline relied on the Windows-only \texttt{win32com} stack and had to be ported to a cross-platform LibreOffice headless workflow; SWE-Bench~Pro needed Jest invocations to be pinned to single-worker / forced-exit flags before its JavaScript and TypeScript tasks would terminate reliably; ScienceAgentBench shipped a Modal harness whose output paths and per-task dependencies drifted from the released evaluator; and SWE-Bench~Verified contains a task (\texttt{scikit-learn\_\_scikit-learn-14710}) that exceeds the hardware budget of our default execution backend.

\paragraph{Non-reproducible scoring.}
Scoring code with hidden non-determinism (network calls, time-based seeds, race conditions in parallel test runners), lenient or accidentally-permissive test matchers, and divergence between the public leaderboard pipeline and the released evaluation script. GSO's speedup metric is sensitive enough to timing variance that 13 of 102 oracle solutions are scored as ``slower than baseline'' on a fraction of runs --- a property of the metric rather than the harness. WideSearch's gold-answer CSVs contain cells whose contents include the markdown column delimiter (\texttt{|}); the upstream CSV-to-markdown conversion silently rewrites these to a space, capping any oracle's Item-F1 at 0.9996. DA-Code's released evaluation script required eleven distinct correctness fixes before its gold outputs scored 1.0. We treat divergences of this kind as benchmark bugs, document them in the adapter README, and keep oracle exclusions out-of-band rather than baked into the dataset.

\paragraph{Implicit assumptions.}
Preprocessing, prompting, or grading pipelines that depend on undocumented conventions: a particular Git branch of a vendored repo, a particular tokenizer, an exact prompt formatting, or a hidden answer-extraction regex. SWE-Bench~Pro's grader silently rejects gold patches that do not end with a newline. SWT-Bench's \texttt{FAIL\_TO\_PASS} field is parsed by an upstream regex that returns the empty set on nine \texttt{sphinx-doc} tasks, marking them as zero-score even under the gold patch. LAB-Bench's multiple-choice harness exhibits non-trivial positional bias unless the answer-choice order is randomized at runtime, which we now enforce via container \texttt{ENTRYPOINT}. LiveCodeBench tasks rely on standard-library imports that the upstream harness injects implicitly. We surface these assumptions in the adapter so that they are part of the task specification rather than the agent's prior knowledge.

\paragraph{Oracle solutions that fail.}
In several benchmarks, the published ground-truth solution did not pass the benchmark's own tests, usually due to environment drift since release rather than a flaw in the solution itself. SWE-Bench~Verified ships 4 tasks whose oracles fail on data or infrastructure errors (99.2\% pass rate); SWE-Bench~Pro ships 15 invalid gold patches and 9 oracles that exceed the upstream timeout (97\% pass rate); SWT-Bench's oracle passes 97.7\% for the parsing reason above; ScienceAgentBench's CPU-only oracle pass rate is 93.1\% (the remaining 7 tasks need GPUs); GSO's oracle passes 87.3\% due to timing variance; SWE-smith's synthetic patches contain a small fraction of upstream-failing entries that the original benchmark also does not score. Without an explicit oracle-validation step these failures are invisible to downstream users and silently inflate baseline scores.

\paragraph{Agent hacking surfaces.}
A less visible but arguably more damaging failure mode arises when the benchmark design inadvertently allows an agent to obtain the ground-truth answer or bypass the verifier without genuinely solving the task. We identified three recurring patterns during adapter construction. \emph{Answer exfiltration via public data sources:} when the task instruction exposes a stable identifier and the container has unrestricted internet access, an agent can query the publicly hosted upstream dataset and retrieve the gold answer directly, passing the verifier without engaging with the task content. \emph{Ground-truth leakage through container mounts or file placement:} if gold labels, answer files, or scoring criteria are visible inside the agent's runtime filesystem---whether through a shared Docker volume, a misconfigured \texttt{COPY} directive, or co-located test assets---the agent can read the expected output before producing its response. \emph{Shared-environment verifier tampering:} when the agent and verifier share the same container and interpreter, an agent can overwrite judge dependencies so that the verification step returns a hard-coded passing score regardless of the agent's actual output; this class of attack is invisible to oracle validation because the oracle path typically bypasses the judge. These surfaces are particularly insidious because they inflate benchmark scores silently: the resulting numbers look plausible, oracle runs still pass, and the exploit is only revealed by careful trajectory inspection or by comparing scores against independent human evaluation. We now flag these surfaces during adapter review and require network restriction, identifier masking, or verifier isolation before merge.

\paragraph{Standards we enforce on every Harbor adapter.}
\begin{itemize}
  \item \textbf{Oracle validation.} The provided ground-truth solution must achieve a 100\% pass rate; deviations must be explained per task and documented in the adapter README, not silently excluded from the dataset.
  \item \textbf{Parity verification.} A parity experiment must confirm statistical equivalence with the original benchmark within sample SEM, on a parity set whose composition is recorded in \texttt{parity\_experiment.json}.
  \item \textbf{Reproducible execution.} Every task ships a sandboxed Docker environment with pinned dependencies; cross-platform tooling is preferred over OS-specific commands.
  \item \textbf{Documented divergences.} Any deviation from the upstream evaluator (prompt rewrite, scoring fix, exclusion list) is captured in the adapter README so that downstream users can audit it.
\end{itemize}

\paragraph{Recommendations for future benchmark authors.}
Based on the above, we recommend that future benchmark releases:
\begin{itemize}
  \item pin all dependencies (Python packages, system libraries, language toolchains, container base images) to exact versions, ideally via lock files;
  \item publish a single reproducible container that bundles the entire evaluation pipeline rather than expecting users to assemble it from a README;
  \item include a runnable oracle solution and gate releases on an oracle-pass-rate check;
  \item make scoring deterministic --- avoid network calls, time-based randomness, lenient regex matchers, and platform-dependent tooling in the grader;
  \item separate task specification from execution harness, so that downstream adapters and agents can be swapped without touching task content;
  \item document every implicit assumption (prompt format, answer-extraction rule, tokenizer, dataset version) in the task specification rather than only in code.
\end{itemize}
Adopting these practices would substantially reduce the per-benchmark integration cost we report in the per-adapter notes.

\section{Evaluation Protocol and Reproducibility Details}
\label{app:evaluation_details}

This section describes the evaluation protocol used to run the benchmark experiments at scale and for Harbor-Index. 
The protocol is designed to make the reported results reproducible and auditable. 
In particular, it specifies how benchmark datasets are prepared as runnable Harbor task registries, how the reported paper subset is sampled and locked, how model--harness configurations are executed in isolated environments, and how trial-level metadata and artifacts are stored for auditing and aggregate scoring.

\subsection{Evaluation Infrastructure}
\label{app:evaluation_infrastructure}

\paragraph{Harness and repository state.}
We use Harbor as the unified harness for task execution, sandboxing, agent orchestration, and verification.
For each benchmark, we follow the corresponding adapter README to construct the runnable dataset and to validate the execution environment, verifier behavior, credential requirements, and resource constraints prior to inclusion.
Unless otherwise noted, large-scale evaluations are conducted with Harbor pinned to commit \texttt{9ee6790376583608f541c133d1af2dae47b8fc32}.
For the Harbor-Index experiments, we use the more recent and stable Harbor@v0.6.4 release, pinned to commit \texttt{331dcba30efcc3fa8282a21de562a3834d6e6244}.
Harbor automatically records token counts, metrics, and agent trajectories.
The \href{https://github.com/harbor-framework/harbor-adapters-experiments}{experiment repository} provides the tooling for dataset metadata upload, job validation, result import, and trial-artifact archival.

\paragraph{Sandbox backends.}
Each trial is executed in an isolated Harbor environment. 
For CPU-only tasks, we use Daytona as the default backend. 
Daytona provides independent task workspaces and isolates the agent and verifier file systems for each trial. 
For a subset of benchmarks, we instead use Harbor's local Docker backend when the verifier or task environment is incompatible with Daytona, for example when CPU, memory, or storage requirements exceed Daytona's resource limits of 4 CPUs, 8 GB of memory, and 10 GB of storage. 
Docker-backed jobs are executed through the same Harbor interface as Daytona-backed jobs. 
For image-heavy benchmarks, we prebuild the outer task image and, when applicable, the inner Docker-in-Docker images, publish them to a registry such as GHCR, and configure the task environment to use the prebuilt image. 
This procedure reduces failures caused by registry rate limits and repeated large downloads during environment setup. 
GPU-dependent tasks are executed through Harbor's Modal backend, which we reserve for benchmarks that require GPU hardware or GPU-capable system images.

\paragraph{Language model services.}
We access each language model through the corresponding first-party provider API, with support from API grants provided by the respective providers. All model configurations are documented in our experiment repository mentioned above. Unless otherwise specified, we use each provider's default settings for temperature, context length, reasoning effort, and other provider-specific parameters; for example, GPT-5.4 is evaluated with the default medium reasoning effort and Claude Opus 4.6 is evaluated with the default high effort. We note that GPT-5.4 is often reported with extra-high reasoning effort on public leaderboards. Readers should therefore exercise caution when comparing our evaluation results with those reported by other leaderboards.

\paragraph{Model and harness matrix.}
The large-scale evaluation includes \MainEvalConfigurationCount{} model--harness
configurations over \MainEvalModelCount{} models spanning capability tiers.
Every model is evaluated under exactly two harness conditions: the cross-family
Terminus-2 harness and one native harness. There are three
native harnesses---Codex CLI for GPT, Claude Code for Claude, and Gemini
CLI for Gemini---and therefore \MainEvalHarnessImplementationCount{} distinct
harness implementations overall. Terminus-2 is evaluated with all
\MainEvalModelCount{} models:
\texttt{gpt-5.4}, \texttt{gpt-5-mini}, \texttt{gpt-5-nano} \cite{openai_pricing},
Claude Opus 4.6, Claude Sonnet 4.6, Claude Haiku 4.5 \cite{anthropic_pricing},
\texttt{gemini-3.1-pro-preview}, and \texttt{gemini-3-flash-preview} \cite{google_gemini_pricing}.
Codex CLI is evaluated with \texttt{gpt-5.4}, \texttt{gpt-5-mini}, and
\texttt{gpt-5-nano} using CLI version \texttt{0.115.0}. Gemini CLI is
evaluated with \texttt{gemini-3.1-pro-preview} and
\texttt{gemini-3-flash-preview} using CLI version \texttt{0.34.0}. Claude Code
is evaluated with Claude Opus 4.6, Claude Sonnet 4.6, and Claude Haiku 4.5
using CLI version \texttt{2.1.81}.

For reproducibility, each configuration records the agent interface, the agent
CLI version when applicable, the model identifier used in the job
configuration, and the run-date range. When provider snapshot or API release
identifiers are available, we additionally report them alongside the
job-configuration identifier. Some provider identifiers correspond to managed
preview endpoints or aliases rather than immutable model snapshots; for these
configurations, we report the exact identifier used during evaluation together
with the corresponding run-date range. For the Gemini preview configurations,
the job-configuration identifiers are \texttt{gemini-3.1-pro-preview} and
\texttt{gemini-3-flash-preview}. Public provider metadata lists these endpoints
as preview models with release dates of February 19, 2026 and December 17,
2025, respectively. Evaluations using these endpoints were run in April 2026.

For the Harbor-Index~1.0 experiments, the \HarborIndexModelCount{} evaluated models split into four \emph{closed-weight} models, whose weights are not publicly released (GPT-5.5, Claude Opus 4.8, Gemini 3.1 Pro, and Qwen3.7 Max), and five \emph{open-weight} models (GLM 5.2, Kimi K2.6, MiniMax M3, DeepSeek V4 Pro, and MiMo V2.5 Pro). GPT-5.5, Claude Opus 4.8, and Gemini 3.1 Pro are served by first-party APIs and retain their vendor-native harnesses (Codex CLI, Claude Code, and Gemini CLI respectively); the remaining six models are served through OpenRouter and run under Claude Code. Each of the \HarborIndexModelCount{} models is also evaluated with Terminus-2. Harbor-Index~1.0 therefore contains $\HarborIndexModelCount\times 2\times \HarborIndexTaskCount = \HarborIndexRolloutCount$ model--harness--task runs. Interactive trajectories and aggregate Pareto numbers are published at \url{https://harbor-index.org}.

\paragraph{Database and artifact storage.}
Results are stored in Supabase. 
The database schema contains eight core tables:
\texttt{dataset},
\texttt{task},
\texttt{dataset\_task},
\texttt{job},
\texttt{trial},
\texttt{agent},
\texttt{model}, and
\texttt{trial\_model}. 
Dataset upload creates \texttt{dataset}, \texttt{task}, and \texttt{dataset\_task} records that capture registry information, task identifiers, task instructions, source metadata, and task-level metadata.
Each job record stores the full job configuration, Harbor Git commit or package version, trial count, start and end timestamps, aggregate job statistics, and verification flags. 
Each trial record stores the trial UUID, task checksum, agent name and version, scalar reward, exception information, setup/execution/verifier timestamps, full trial configuration, and agent metadata. 
The \texttt{trial\_model} table stores model and provider names, as well as input, output, and cache token counts for each trial.

We also upload the full trial directory as a compressed archive to a public Supabase Storage bucket named \texttt{trials}.
When present, \texttt{agent/trajectory.json} is uploaded separately as \texttt{<trial-id>-traj.json}.
Before upload, large transient agent artifacts that are not required for evaluation or audit are removed, including Codex runtime temporary directories and very large Terminus-2 terminal recordings.
The retained artifacts include the reward file, verifier output, configuration, trajectory, and logs needed to audit individual trials.

\paragraph{Public trajectory release.}
The complete rollout set of the large-scale evaluation is released as a Hugging Face dataset at \url{https://huggingface.co/datasets/kendx/Harbor-Adapter}.
The release has two configurations.
The \texttt{manifest} configuration is a lightweight catalog with one row per $(\text{benchmark}, \text{task}, \text{model}, \text{agent})$ cell ($178{,}647$ rows) listing the trial identifiers for that cell, so it can be filtered without downloading trajectory bytes.
The \texttt{trajectories} configuration stores one row per trial ($793{,}698$ trials, about $340$~GB in total), each holding the gzipped trial directory described above together with its SHA-256 checksum; shards are laid out per benchmark so that a single benchmark can be fetched in isolation.
Each cell retains up to its five most recent trials, so a cell may contain more than the three trials aggregated in this paper when the job was rerun during auditing (Appendix~\ref{app:error_handling_quality_control}).
This release lets others reproduce our aggregate numbers, re-audit individual trials, and reuse the trajectories for analyses beyond those reported here.

\subsection{Dataset Preparation and Sampling}
\label{app:dataset_preparation_sampling}

For each external benchmark, we follow the Harbor-provided dataset generation workflow described in the benchmark's README. 
Before adding a benchmark to the evaluation suite, we inspect the generated dataset to determine whether it requires LLM-as-a-judge evaluation, Docker-in-Docker execution, GPU resources, or prebuilt images. 
The large-scale evaluation subset includes only tasks that can be executed through the standard Harbor job interface: each task provides an instruction to the evaluated agent, exposes an isolated execution environment, and computes a scalar reward using the benchmark verifier.

To ensure compatibility with the target execution platforms, namely Daytona and Modal, we run a smoke test for each benchmark using the low-cost GPT-5-nano model together with Codex.
For benchmarks with a large number of tasks, we sample an i.i.d. subset to control evaluation cost while preserving broad benchmark coverage. 
The resulting large-scale evaluation set contains 54 benchmarks and 6,627 selected tasks. 
For each benchmark, the manifest records the selected task count, original task count, sample rate, verifier type, and execution backend. 
Table~\ref{tab:adapter-task-summary} lists the benchmarks included in the manifest and their selected task counts. 
After this preparation phase, the evaluation suite can be executed at large scale.

\begin{table}[H]
\centering
\Large
\caption{Adapter task summary. Sample Rate = Paper \#Tasks / Original \#Tasks.}
\resizebox{\textwidth}{!}{%
\begin{tabular}{@{}lrrr@{\hspace{0.8cm}}lrrr@{}}
\toprule
Adapter Name & Paper \#Tasks & Original \#Tasks & Sample Rate &
Adapter Name & Paper \#Tasks & Original \#Tasks & Sample Rate \\
\midrule
PIXIU & 101 & 435 & 23.22\% &
FeatureBench & 185 & 200 & 92.50\% \\

SciCode & 80 & 80 & 100.00\% &
ResearchCodeBench & 212 & 212 & 100.00\% \\

AlgoTune & 154 & 154 & 100.00\% &
GSO & 102 & 102 & 100.00\% \\

BFCL & 123 & 3,641 & 3.38\% &
HLE & 249 & 2,500 & 9.96\% \\

Aider Polyglot & 225 & 225 & 100.00\% &
Reasoning Gym & 576 & 576 & 100.00\% \\

LawBench & 181 & 1,000 & 18.10\% &
SLDBench & 8 & 8 & 100.00\% \\

SpreadsheetBench & 200 & 400 & 50.00\% &
Spider 2 & 64 & 68 & 94.12\% \\

Seal-0 & 111 & 111 & 100.00\% &
StrongReject & 150 & 12,207 & 1.23\% \\

Omni-Math & 200 & 4,428 & 4.52\% &
SWE-smith & 98 & 59,136 & 0.17\% \\

Terminal-Bench 2.0 & 89 & 89 & 100.00\% &
MMMLU & 150 & 210,630 & 0.07\% \\

DeepSynth & 40 & 40 & 100.00\% &
USACO & 100 & 307 & 32.57\% \\

GAIA2 & 100 & 800 & 12.50\% &
KUMO & 212 & 5,300 & 4.00\% \\

WideSearch & 100 & 200 & 50.00\% &
FinanceAgent & 50 & 50 & 100.00\% \\

CRUST-Bench & 100 & 100 & 100.00\% &
DA-Code & 200 & 479 & 41.75\% \\

SWT Bench & 50 & 433 & 11.55\% &
SkillsBench & 77 & 86 & 89.53\% \\

GAIA & 165 & 165 & 100.00\% &
SWE-Bench Pro & 100 & 730 & 13.70\% \\

AIME & 60 & 60 & 100.00\% &
QuixBugs & 80 & 80 & 100.00\% \\

SWE-Lancer & 100 & 463 & 21.60\% &
SWE-Bench-Multilingual & 50 & 300 & 16.67\% \\

ARC-AGI-2 & 100 & 167 & 59.88\% &
BIX-Bench & 50 & 205 & 24.39\% \\

CompileBench & 15 & 15 & 100.00\% &
MedAgentBench & 100 & 300 & 33.33\% \\

GPQA Diamond & 198 & 198 & 100.00\% &
SimpleQA & 200 & 1,000 & 20.00\% \\

HumanEvalFix & 164 & 164 & 100.00\% &
SWE-bench-verified & 100 & 500 & 20.00\% \\

IneqMath & 100 & 100 & 100.00\% &
LiveCodeBench & 100 & 1,055 & 9.48\% \\

LAB-Bench & 181 & 181 & 100.00\% &
CodePDE & 5 & 5 & 100.00\% \\

MMAU & 100 & 1,000 & 10.00\% &
ReplicationBench & 90 & 111 & 81.08\% \\

QCircuitBench & 28 & 28 & 100.00\% &
AA-LCR & 99 & 100 & 99.00\% \\

BigCodeBench-Hard & 145 & 145 & 100.00\% &
CyberGym & 10 & 1,507 & 0.66\% \\
\bottomrule
\end{tabular}%
}

\label{tab:adapter-task-summary}
\end{table}

\subsection{Error Handling and Quality Control}
\label{app:error_handling_quality_control}

\paragraph{Phase-based debugging.}
We structure the large-scale evaluation into phases. We first evaluate each benchmark independently and, within each benchmark, proceed in three phases. Phase~1 serves as a sanity check for model--harness compatibility. In this phase, we run a small subset of tasks (1\% of the full dataset) using the complete model--harness matrix, and verify that the environment starts successfully, the agent receives the intended instruction, the verifier runs, the reward is written, token usage is recorded, and trajectory upload succeeds. This phase is also intended to surface severe issues, such as failures to install a particular harness in the benchmark environment due to system incompatibilities. For benchmarks with reference solutions or oracle support, we additionally run an oracle or known-good trajectory to confirm that the verifier produces the expected reward. Phase~2 evaluates the full model--harness matrix on a 10\% subset of the dataset, and Phase~3 performs the full evaluation. Each phase excludes tasks that were already included in earlier phases.

\paragraph{Local and database checks.}
After jobs finish, we inspect local job statistics in a long-form per-task report. 
This report groups trials by job, dataset, agent, task, reward, token usage, and error type, making it possible to identify all-failure jobs, missing token counts, unexpectedly high verifier failures, or reward distributions inconsistent with earlier phases. 
Supabase is then checked for completeness: each expected job should have a job row, the expected number of trial rows, \texttt{trial\_model} token rows, and storage URLs for uploaded trial archives. 
Jobs with suspiciously missing trajectories, missing or zero token counts, or unexpected exception spikes are rerun or manually audited before inclusion.

\paragraph{Rerun policy.}
We rerun a trial when the recorded exception is caused by infrastructure or configuration rather than by the evaluated agent's behavior. 
Typical rerun triggers include API authentication errors, temporary provider outages, bursty rate-limit failures, sandbox creation failures, image pull failures, verifier crashes due to missing external resources, malformed judge responses caused by judge API failures, corrupted result files, and failed storage or database writes.

We do not rerun agent timeouts within the benchmark execution envelope, verifier timeouts induced by the submitted solution, policy refusals that reflect model behavior on the task prompt, max-output or context-limit failures caused by the agent's interaction strategy, or verifier failures that correctly indicate the task was not solved. 
When the distinction is ambiguous, the trial archive and trajectory are inspected, and the final decision is recorded with the job audit notes.

\subsection{Harbor-Index Experiment Protocol}
  \label{app:harbor_index_protocol}

  \paragraph{Task curation and execution budgets.}
  Harbor-Index~1.0 contains \HarborIndexTaskCount{} tasks spanning
  \HarborIndexBenchmarkCount{} benchmarks after difficulty
  filtering, AI and human audit, and an audit-and-fix loop
  (Appendix~\ref{sec:harbor_index_appendix}). Tasks run in isolated Docker
  sandboxes with separate verifier environments. Performance-oriented tasks
  (e.g., AlgoTune, GSO) use GPU-capable backends; free-form answer tasks
  (e.g., HLE, GAIA, GPQA Diamond) use LLM-as-a-judge scoring. We augment each
  task instruction with an explicit execution contract that states both the
  time budget and the available resources, for example: ``You should solve the
  task in 1 hour. You can access 1 CPU, 2 GB memory, 10 GB storage.'' The same
  limits are enforced in the task configuration through the Harbor
  \texttt{agent} fields. Timeouts are tightened during the audit-and-fix
  stage (typically $1.2\times$ the fastest frontier model's observed runtime,
  or up to 3 hours when every model fails).

  \paragraph{Scoring and thresholding.}
  Most Harbor-Index tasks have binary rewards and are scored directly from the
  verifier output. For continuous-score tasks, we convert the raw verifier reward
  to pass/fail using a locked task-specific threshold. Let \(r_{i,t}\) be the raw
  reward for trial \(i\) on task \(t\). For each continuous task, we define
  \[
  \tau_t = \left\lfloor \max_i r_{i,t} \right\rfloor_{5},
  \]
  where \(\lfloor \cdot \rfloor_{5}\) denotes flooring to five decimal places
  over all non-failed observed trials, including oracle calibration runs when
  available. A trial is counted as passing if its raw reward strictly exceeds
  \(\tau_t\). For bounded continuous metrics with a known upper bound (e.g., F1 score, $R^2$), we also
  count a trial as passing if it reaches the upper bound. Speedup-style tasks such as AlgoTune and GSO do not use a
  fixed upper bound.

\subsection{Evaluation Results}
\label{app:large_scale_eval_matrices}

In Tables~\ref{tab:gpt-performance-full}, \ref{tab:claude-performance-full}, and \ref{tab:gemini-performance-full}, we present the benchmark scores for all model--harness configurations on all the benchmarks we study.


\begin{table}[!ht]
\caption{GPT-series models under Terminus-2 and Codex across \MainEvalBenchmarkCount{} benchmarks.}
\label{tab:gpt-performance-full}
\small
\centering
\begin{tabular}{@{}lcccccc@{}}
\toprule
\multirow{2}{*}{Benchmark}
& \multicolumn{2}{c}{GPT-5.4}
& \multicolumn{2}{c}{GPT-5-mini}
& \multicolumn{2}{c}{GPT-5-nano} \\
\cmidrule(lr){2-3} \cmidrule(lr){4-5} \cmidrule(lr){6-7}
& Terminus-2
& Codex
& Terminus-2
& Codex
& Terminus-2
& Codex
\\ \midrule
SWE-bench~Verified & 0.690 & 0.713 & 0.307 & 0.460 & 0.160 & 0.110 \\
SWE-Bench-Multilingual & 0.607 & 0.700 & 0.213 & 0.280 & 0.047 & 0.033 \\
SWE-smith & 0.031 & 0.221 & 0.010 & 0.296 & 0.065 & 0.051 \\
CRUST-Bench & 0.920 & 0.980 & 0.620 & 0.803 & 0.327 & 0.113 \\
GSO & 0.167 & 0.373 & 0.007 & 0.095 & 0.000 & 0.010 \\
SWE-Bench~Pro & 0.347 & 0.550 & 0.100 & 0.400 & 0.053 & 0.057 \\
SWT~Bench & 0.540 & 0.707 & 0.147 & 0.307 & 0.073 & 0.147 \\
Aider~Polyglot & 0.544 & 0.785 & 0.367 & 0.539 & 0.141 & 0.096 \\
AlgoTune & 0.166 & 0.275 & 0.130 & 0.144 & 0.050 & 0.009 \\
CompileBench & 0.933 & 1.000 & 0.533 & 0.644 & 0.400 & 0.133 \\
FeatureBench & 0.056 & 0.368 & 0.034 & 0.103 & 0.009 & 0.004 \\
BigCodeBench-Hard & 0.395 & 0.428 & 0.352 & 0.283 & 0.244 & 0.034 \\
USACO & 0.830 & 0.960 & 0.843 & 0.857 & 0.713 & 0.057 \\
SWE-Lancer & 0.457 & 0.580 & 0.407 & 0.397 & 0.230 & 0.303 \\
LiveCodeBench & 0.873 & 0.927 & 0.847 & 0.883 & 0.697 & 0.307 \\
HumanEvalFix & 0.998 & 1.000 & 0.988 & 0.998 & 0.941 & 0.470 \\
QuixBugs & 0.863 & 0.975 & 0.892 & 0.887 & 0.362 & 0.292 \\
\midrule
IneqMath & 0.853 & 0.997 & 0.923 & 0.980 & 0.780 & 0.903 \\
Omni-Math & 0.543 & 0.852 & 0.767 & 0.833 & 0.710 & 0.622 \\
Reasoning~Gym & 0.842 & 0.902 & 0.841 & 0.819 & 0.671 & 0.563 \\
KUMO & 0.926 & 0.964 & 0.849 & 0.936 & 0.541 & 0.668 \\
AIME 24 \& 25 & 0.711 & 0.989 & 0.889 & 0.967 & 0.756 & 0.739 \\
ARC-AGI-2 & 0.020 & 0.533 & 0.003 & 0.003 & 0.000 & 0.000 \\
\midrule
AA-LCR & 0.525 & 0.751 & 0.320 & 0.744 & 0.131 & 0.360 \\
SimpleQA & 0.748 & 0.945 & 0.645 & 0.955 & 0.158 & 0.762 \\
GPQA~Diamond & 0.778 & 0.923 & 0.785 & 0.798 & 0.608 & 0.613 \\
Humanity's~Last~Exam & 0.203 & 0.519 & 0.131 & 0.312 & 0.058 & 0.111 \\
MMMLU & 0.356 & 0.838 & 0.649 & 0.727 & 0.582 & 0.602 \\
\midrule
GAIA & 0.414 & 0.760 & 0.253 & 0.701 & 0.164 & 0.428 \\
SkillsBench & 0.317 & 0.602 & 0.132 & 0.258 & 0.054 & 0.017 \\
BFCL & 0.734 & 0.683 & 0.791 & 0.764 & 0.621 & 0.610 \\
GAIA2 & 0.170 & 0.210 & 0.000 & 0.166 & 0.000 & 0.035 \\
Terminal-Bench~2.0 & 0.427 & 0.693 & 0.251 & 0.371 & 0.105 & 0.052 \\
DeepSynth & 0.289 & 0.548 & 0.069 & 0.300 & 0.030 & 0.088 \\
Seal-0 & 0.276 & 0.511 & 0.192 & 0.459 & 0.048 & 0.258 \\
WideSearch & 0.560 & 0.791 & 0.303 & 0.533 & 0.131 & 0.120 \\
\midrule
Spider~2 & 0.219 & 0.276 & 0.109 & 0.245 & 0.130 & 0.141 \\
DA-Code & 0.558 & 0.565 & 0.377 & 0.499 & 0.250 & 0.171 \\
\midrule
SpreadsheetBench & 0.643 & 0.762 & 0.452 & 0.588 & 0.263 & 0.095 \\
FinanceAgent & 0.500 & 0.767 & 0.113 & 0.533 & 0.000 & 0.213 \\
LawBench & 0.586 & 0.638 & 0.486 & 0.425 & 0.321 & 0.110 \\
PIXIU & 0.574 & 0.710 & 0.506 & 0.480 & 0.283 & 0.337 \\
MedAgentBench & 0.340 & 0.643 & 0.350 & 0.430 & 0.080 & 0.290 \\
\midrule
LAB-Bench & 0.230 & 0.731 & 0.160 & 0.475 & 0.083 & 0.120 \\
SLDBench & 0.673 & 0.795 & 0.542 & 0.707 & 0.225 & 0.075 \\
BIX-Bench & 0.300 & 0.407 & 0.193 & 0.313 & 0.073 & 0.073 \\
ReplicationBench & 0.256 & 0.607 & 0.081 & 0.233 & 0.041 & 0.041 \\
QCircuitBench & 0.496 & 0.620 & 0.358 & 0.177 & 0.142 & 0.000 \\
ResearchCodeBench & 0.006 & 0.866 & 0.000 & 0.599 & 0.002 & 0.421 \\
SciCode & 0.469 & 0.582 & 0.430 & 0.226 & 0.217 & 0.004 \\
CodePDE & 0.600 & 0.467 & 0.533 & 0.333 & 0.267 & 0.000 \\
\midrule
StrongReject & 0.959 & 0.949 & 0.935 & 0.958 & 0.961 & 0.951 \\
CyberGym & 0.533 & 0.767 & 0.633 & 0.633 & 0.167 & 0.233 \\
\midrule
MMAU & 0.660 & 0.703 & 0.597 & 0.583 & 0.537 & 0.070 \\
\bottomrule
\end{tabular}
\end{table}

\begin{table}[!ht]
\caption{Claude-series models under Terminus-2 and Claude Code across \MainEvalBenchmarkCount{} benchmarks.}
\label{tab:claude-performance-full}
\small
\centering
\begin{tabular}{@{}lcccccc@{}}
\toprule
\multirow{2}{*}{Benchmark}
& \multicolumn{2}{c}{Claude Haiku 4.5}
& \multicolumn{2}{c}{Claude Sonnet 4.6}
& \multicolumn{2}{c}{Claude Opus 4.6} \\
\cmidrule(lr){2-3} \cmidrule(lr){4-5} \cmidrule(lr){6-7}
& {\scriptsize Terminus-2}
& {\scriptsize Claude Code}
& {\scriptsize Terminus-2}
& {\scriptsize Claude Code}
& {\scriptsize Terminus-2}
& {\scriptsize Claude Code}
\\ \midrule
SWE-bench~Verified & 0.617 & 0.573 & 0.733 & 0.703 & 0.750 & 0.733 \\
SWE-Bench-Multilingual & 0.687 & 0.473 & 0.733 & 0.693 & 0.800 & 0.667 \\
SWE-smith & 0.187 & 0.231 & 0.415 & 0.289 & 0.582 & 0.412 \\
CRUST-Bench & 0.637 & 0.770 & 0.847 & 0.713 & 0.870 & 0.823 \\
GSO & 0.010 & 0.131 & 0.428 & 0.451 & 0.366 & 0.392 \\
SWE-Bench~Pro & 0.080 & 0.450 & 0.473 & 0.490 & 0.487 & 0.537 \\
SWT~Bench & 0.160 & 0.273 & 0.447 & 0.573 & 0.673 & 0.647 \\
Aider~Polyglot & 0.299 & 0.301 & 0.572 & 0.560 & 0.719 & 0.686 \\
AlgoTune & 0.096 & 0.123 & 0.260 & 0.270 & 0.259 & 0.231 \\
CompileBench & 0.822 & 0.889 & 0.867 & 0.911 & 0.933 & 0.933 \\
FeatureBench & 0.034 & 0.126 & 0.189 & 0.277 & 0.405 & 0.323 \\
BigCodeBench-Hard & 0.393 & 0.345 & 0.446 & 0.437 & 0.444 & 0.416 \\
USACO & 0.273 & 0.380 & 0.727 & 0.750 & 0.800 & 0.780 \\
SWE-Lancer & 0.337 & 0.357 & 0.453 & 0.487 & 0.583 & 0.523 \\
LiveCodeBench & 0.460 & 0.607 & 0.823 & 0.837 & 0.887 & 0.870 \\
HumanEvalFix & 0.929 & 0.994 & 1.000 & 0.998 & 1.000 & 1.000 \\
QuixBugs & 0.717 & 0.887 & 0.925 & 0.975 & 0.954 & 0.933 \\
\midrule
IneqMath & 0.733 & 0.953 & 0.970 & 0.997 & 0.860 & 0.860 \\
Omni-Math & 0.600 & 0.760 & 0.788 & 0.800 & 0.813 & 0.840 \\
Reasoning~Gym & 0.803 & 0.807 & 0.869 & 0.818 & 0.875 & 0.851 \\
KUMO & 0.909 & 0.956 & 0.948 & 0.948 & 0.964 & 0.964 \\
AIME 24 \& 25 & 0.622 & 0.861 & 0.978 & 0.967 & 0.989 & 0.994 \\
ARC-AGI-2 & 0.007 & 0.060 & 0.180 & 0.187 & 0.370 & 0.437 \\
\midrule
AA-LCR & 0.606 & 0.549 & 0.663 & 0.741 & 0.616 & 0.747 \\
SimpleQA & 0.573 & 0.920 & 0.807 & 0.803 & 0.907 & 0.867 \\
GPQA~Diamond & 0.535 & 0.709 & 0.823 & 0.854 & 0.843 & 0.884 \\
Humanity's~Last~Exam & 0.064 & 0.107 & 0.217 & 0.325 & 0.343 & 0.390 \\
MMMLU & 0.691 & 0.671 & 0.751 & 0.756 & 0.769 & 0.778 \\
\midrule
GAIA & 0.313 & 0.570 & 0.529 & 0.679 & 0.612 & 0.671 \\
SkillsBench & 0.121 & 0.254 & 0.311 & 0.489 & 0.346 & 0.454 \\
BFCL & 0.770 & 0.729 & 0.770 & 0.805 & 0.764 & 0.799 \\
GAIA2 & 0.056 & 0.213 & 0.319 & 0.384 & 0.357 & 0.358 \\
Terminal-Bench~2.0 & 0.303 & 0.322 & 0.581 & 0.569 & 0.618 & 0.607 \\
DeepSynth & 0.062 & 0.112 & 0.398 & 0.454 & 0.492 & 0.461 \\
Seal-0 & 0.060 & 0.258 & 0.342 & 0.294 & 0.390 & 0.348 \\
WideSearch & 0.326 & 0.574 & 0.511 & 0.673 & 0.617 & 0.758 \\
\midrule
Spider~2 & 0.208 & 0.255 & 0.339 & 0.370 & 0.375 & 0.396 \\
DA-Code & 0.504 & 0.510 & 0.578 & 0.580 & 0.616 & 0.580 \\
\midrule
SpreadsheetBench & 0.560 & 0.702 & 0.803 & 0.832 & 0.832 & 0.818 \\
FinanceAgent & 0.080 & 0.207 & 0.800 & 0.800 & 0.720 & 0.833 \\
LawBench & 0.538 & 0.565 & 0.608 & 0.620 & 0.654 & 0.666 \\
PIXIU & 0.485 & 0.489 & 0.569 & 0.574 & 0.607 & 0.612 \\
MedAgentBench & 0.467 & 0.537 & 0.497 & 0.503 & 0.597 & 0.557 \\
\midrule
LAB-Bench & 0.123 & 0.357 & 0.018 & 0.077 & 0.096 & 0.451 \\
SLDBench & 0.612 & 0.772 & 0.747 & 0.759 & 0.760 & 0.522 \\
BIX-Bench & 0.153 & 0.287 & 0.293 & 0.393 & 0.300 & 0.393 \\
ReplicationBench & 0.104 & 0.152 & 0.207 & 0.219 & 0.241 & 0.285 \\
QCircuitBench & 0.232 & 0.384 & 0.507 & 0.493 & 0.562 & 0.489 \\
ResearchCodeBench & 0.003 & 0.434 & 0.006 & 0.533 & 0.003 & 0.607 \\
SciCode & 0.345 & 0.366 & 0.439 & 0.452 & 0.485 & 0.488 \\
CodePDE & 0.400 & 0.333 & 0.600 & 0.333 & 0.600 & 0.400 \\
\midrule
StrongReject & 0.970 & 0.962 & 0.909 & 0.906 & 0.910 & 0.919 \\
CyberGym & 0.633 & 0.733 & 0.700 & 0.500 & 0.567 & 0.633 \\
\midrule
MMAU & 0.550 & 0.503 & 0.633 & 0.617 & 0.697 & 0.690 \\
\bottomrule
\end{tabular}
\end{table}

\begin{table}[!ht]
\caption{Gemini-series models under Terminus-2 and Gemini CLI across \MainEvalBenchmarkCount{} benchmarks.}
\label{tab:gemini-performance-full}
\small
\centering
\begin{tabular}{@{}lcccc@{}}
\toprule
\multirow{2}{*}{Benchmark}
& \multicolumn{2}{c}{Gemini 3.1 Pro Preview}
& \multicolumn{2}{c}{Gemini 3 Flash Preview} \\
\cmidrule(lr){2-3} \cmidrule(lr){4-5}
& Terminus-2
& Gemini CLI
& Terminus-2
& Gemini CLI
\\ \midrule
SWE-bench~Verified & 0.777 & 0.757 & 0.687 & 0.733 \\
SWE-Bench-Multilingual & 0.767 & 0.747 & 0.747 & 0.767 \\
SWE-smith & 0.323 & 0.442 & 0.296 & 0.211 \\
CRUST-Bench & 0.940 & 0.923 & 0.687 & 0.853 \\
GSO & 0.373 & 0.431 & 0.245 & 0.408 \\
SWE-Bench~Pro & 0.467 & 0.457 & 0.390 & 0.460 \\
SWT~Bench & 0.573 & 0.487 & 0.600 & 0.560 \\
Aider~Polyglot & 0.788 & 0.884 & 0.667 & 0.713 \\
AlgoTune & 0.239 & 0.278 & 0.182 & 0.163 \\
CompileBench & 0.933 & 0.956 & 0.956 & 1.000 \\
FeatureBench & 0.258 & 0.405 & 0.038 & 0.187 \\
BigCodeBench-Hard & 0.405 & 0.425 & 0.386 & 0.400 \\
USACO & 0.933 & 0.540 & 0.920 & 0.880 \\
SWE-Lancer & 0.630 & 0.710 & 0.477 & 0.500 \\
LiveCodeBench & 0.920 & 0.923 & 0.877 & 0.907 \\
HumanEvalFix & 1.000 & 1.000 & 0.998 & 0.998 \\
QuixBugs & 0.925 & 0.933 & 0.958 & 0.971 \\
\midrule
IneqMath & 0.947 & 0.870 & 0.867 & 0.193 \\
Omni-Math & 0.872 & 0.880 & 0.835 & 0.650 \\
Reasoning~Gym & 0.906 & 0.804 & 0.868 & 0.594 \\
KUMO & 0.969 & 0.967 & 0.953 & 0.964 \\
AIME 24 \& 25 & 0.989 & 0.994 & 0.889 & 0.811 \\
ARC-AGI-2 & 0.603 & 0.767 & 0.350 & 0.377 \\
\midrule
AA-LCR & 0.744 & 0.660 & 0.643 & 0.650 \\
SimpleQA & 0.950 & 0.960 & 0.840 & 0.910 \\
GPQA~Diamond & 0.953 & 0.971 & 0.896 & 0.704 \\
Humanity's~Last~Exam & 0.533 & 0.541 & 0.395 & 0.272 \\
MMMLU & 0.804 & 0.769 & 0.824 & 0.740 \\
\midrule
GAIA & 0.677 & 0.671 & 0.568 & 0.364 \\
SkillsBench & 0.374 & 0.582 & 0.260 & 0.564 \\
BFCL & 0.837 & 0.813 & 0.718 & 0.743 \\
GAIA2 & 0.215 & 0.331 & 0.183 & 0.214 \\
Terminal-Bench~2.0 & 0.689 & 0.633 & 0.513 & 0.524 \\
DeepSynth & 0.550 & 0.588 & 0.359 & 0.328 \\
Seal-0 & 0.420 & 0.505 & 0.393 & 0.393 \\
WideSearch & 0.641 & 0.660 & 0.641 & 0.536 \\
\midrule
Spider~2 & 0.344 & 0.349 & 0.250 & 0.323 \\
DA-Code & 0.606 & 0.601 & 0.558 & 0.561 \\
\midrule
SpreadsheetBench & 0.795 & 0.825 & 0.755 & 0.768 \\
FinanceAgent & 0.753 & 0.760 & 0.540 & 0.380 \\
LawBench & 0.697 & 0.644 & 0.648 & 0.542 \\
PIXIU & 0.748 & 0.748 & 0.711 & 0.704 \\
MedAgentBench & 0.637 & 0.550 & 0.507 & 0.307 \\
\midrule
LAB-Bench & 0.313 & 0.709 & 0.311 & 0.678 \\
SLDBench & 0.784 & 0.737 & 0.789 & 0.747 \\
BIX-Bench & 0.433 & 0.480 & 0.393 & 0.393 \\
ReplicationBench & 0.541 & 0.630 & 0.219 & 0.356 \\
QCircuitBench & 0.687 & 0.560 & 0.643 & 0.491 \\
ResearchCodeBench & 0.000 & 0.833 & 0.002 & 0.711 \\
SciCode & 0.550 & 0.537 & 0.511 & 0.144 \\
CodePDE & 0.267 & 0.133 & 0.333 & 0.200 \\
\midrule
StrongReject & 0.868 & 0.916 & 0.733 & 0.875 \\
CyberGym & 0.833 & 0.800 & 0.733 & 0.733 \\
\midrule
MMAU & 0.627 & 0.610 & 0.627 & 0.507 \\
\bottomrule
\end{tabular}
\end{table}

\section{Detailed Large-scale Quantitative Analysis}


\providecommand{\AppNumSystems}{16}
\providecommand{\AppNumBenchmarks}{54}
\providecommand{\AppNumModelsOnly}{8}

\providecommand{\AppSVDModelOnlyPCOne}{73.7}
\providecommand{\AppSVDModelOnlyPCOneTwo}{81.9}
\providecommand{\AppSVDFullPCOne}{67.1}
\providecommand{\AppSVDFullPCTwo}{10.1}
\providecommand{\AppSVDFullCumTwo}{77.2}

\providecommand{\AppNumFamilyHarnessSystems}{8}
\providecommand{\AppNumTerminusSystems}{8}
\providecommand{\AppPCTwoAgentMean}{+1.65}
\providecommand{\AppPCTwoBaseMean}{-1.65}
\providecommand{\AppPCTwoGap}{3.31}

\providecommand{\AppBaselineMedAE}{0.117}
\providecommand{\AppLogitBenchRegMedAE}{0.081}
\providecommand{\AppSVDLogitMedAE}{0.069}
\providecommand{\AppBlendMedAE}{0.063}
\providecommand{\AppBlendImprovement}{46}

\providecommand{\AppPerBenchmarkPredictTable}{\begin{table}[H]\small
\centering
\caption{Per-benchmark predictability (50\% holdout, 3~folds).
  Low MedAE = easily predicted from other benchmarks (redundant);
  high MedAE = hard to predict (unique signal).}
\label{tab:bench-predict}
\begin{tabular*}{\linewidth}{@{\extracolsep{\fill}}lr@{\hskip 2em}lr@{}}
\toprule
\multicolumn{2}{c}{\emph{Most Redundant}} & \multicolumn{2}{c}{\emph{Most Unique}} \\
\cmidrule(r){1-2} \cmidrule(l){3-4}
Benchmark & MedAE & Benchmark & MedAE \\
\midrule
HumanEvalFix & 0.004 & ResearchCodeBench & 0.221 \\
KUMO & 0.017 & FinanceAgent & 0.202 \\
BFCL & 0.029 & LAB-Bench & 0.168 \\
StrongReject & 0.030 & CodePDE & 0.164 \\
BigCodeBench & 0.031 & USACO & 0.163 \\
LawBench & 0.035 & MedAgentBench & 0.121 \\
DA-Code & 0.038 & CyberGym & 0.102 \\
Reasoning Gym & 0.042 & IneqMath & 0.099 \\
\bottomrule
\end{tabular*}
\end{table}}

\providecommand{\BenchPressRedundantOne}{HumanEvalFix}
\providecommand{\BenchPressRedundantOneScore}{0.004}
\providecommand{\BenchPressUniqueOne}{ResearchCodeBench}
\providecommand{\BenchPressUniqueOneScore}{0.221}
\providecommand{\BenchPressRedundantTwo}{KUMO}
\providecommand{\BenchPressRedundantTwoScore}{0.017}
\providecommand{\BenchPressUniqueTwo}{FinanceAgent}
\providecommand{\BenchPressUniqueTwoScore}{0.202}
\providecommand{\BenchPressRedundantThree}{BFCL}
\providecommand{\BenchPressRedundantThreeScore}{0.029}
\providecommand{\BenchPressUniqueThree}{LAB-Bench}
\providecommand{\BenchPressUniqueThreeScore}{0.168}

\providecommand{\AppBenchGreedySequence}{\begin{table}[H]
\centering
\caption{Greedy forward selection sequence: benchmarks added in order of decreasing independence. $\max|\rho|$ is the maximum Spearman correlation with any previously selected benchmark. The line separates the 12 benchmarks below the $|\rho| = 0.7$ redundancy threshold from the first benchmark that exceeds it.}
\label{tab:greedy-sequence}
\small
\begin{tabular}{lrl}
\toprule
Benchmark & $\max|\rho|$ & Domain \\
\midrule
CodePDE & 0.00 & Scientific Research \\
LiveCodeBench & 0.01 & Software Engineering \\
IneqMath & 0.11 & Mathematics \& Reasoning \\
BFCL & 0.27 & Agents, Tools \& Systems \\
ResearchCodeBench & 0.33 & Scientific Research \\
FinanceAgent & 0.47 & Professional Domains \\
SLDBench & 0.56 & Scientific Research \\
StrongReject & 0.59 & Safety \& Security \\
Reasoning Gym & 0.61 & Mathematics \& Reasoning \\
CyberGym & 0.66 & Agents, Tools \& Systems \\
LAB-Bench & 0.67 & Scientific Research \\
Omni-math & 0.69 & Mathematics \& Reasoning \\
\midrule
SWE-bench-verified & 0.70 & Software Engineering \\
\bottomrule
\end{tabular}
\end{table}}
\providecommand{\GreedyFirstAboveSevenCorr}{0.70}

\providecommand{\TaskTotalTasks}{6627}
\providecommand{\TaskZeroVarTasks}{731}
\providecommand{\TaskGoodTasks}{5896}
\providecommand{\TaskFullPCOne}{32.0}
\providecommand{\TaskFullPCTwo}{11.7}
\providecommand{\TaskModelPCOne}{41.9}
\providecommand{\TaskAgentPCTwoMean}{64.5}
\providecommand{\TaskBasePCTwoMean}{-64.5}

\providecommand{\TaskRedundNumBenchmarks}{52}
\providecommand{\TaskRedundMeanPCOne}{49.5}
\providecommand{\TaskRedundMedianPCsNinety}{8}
\providecommand{\TaskRedundMeanCompression}{15.4}
\providecommand{\TaskRedundMeanZeroVarPct}{11.0}
\providecommand{\TaskRedundRhoKOne}{0.885}
\providecommand{\TaskRedundRhoKThree}{0.923}
\providecommand{\TaskRedundRhoKFive}{0.941}
\providecommand{\TaskRedundRhoKTen}{0.953}
\providecommand{\TaskRedundAboveNFKOne}{3/52}
\providecommand{\TaskRedundAboveNFKThree}{12/52}
\providecommand{\TaskRedundAboveNFKFive}{22/52}
\providecommand{\TaskRedundAboveNFKTen}{33/52}
\providecommand{\TaskLowRankExOneName}{HumanEvalFix}
\providecommand{\TaskLowRankExOneTasks}{164}
\providecommand{\TaskLowRankExOnePCOne}{86.2}
\providecommand{\TaskLowRankExOnePCs}{2}
\providecommand{\TaskLowRankExTwoName}{ResearchCodeBench}
\providecommand{\TaskLowRankExTwoTasks}{212}
\providecommand{\TaskLowRankExTwoPCOne}{78.4}
\providecommand{\TaskLowRankExTwoPCs}{3}
\providecommand{\TaskLowRankExThreeName}{FinanceAgent}
\providecommand{\TaskLowRankExThreeTasks}{50}
\providecommand{\TaskLowRankExThreePCOne}{69.5}
\providecommand{\TaskLowRankExThreePCs}{6}
\providecommand{\TaskZeroVarExOneName}{StrongReject}
\providecommand{\TaskZeroVarExOnePct}{52}
\providecommand{\TaskZeroVarExTwoName}{Spider 2}
\providecommand{\TaskZeroVarExTwoPct}{44}
\providecommand{\TaskZeroVarExThreeName}{GAIA2}
\providecommand{\TaskZeroVarExThreePct}{32}

\providecommand{\AppPerBenchRhoTable}{\begin{table}[H]\small
\centering
\caption{Per-benchmark ranking fidelity: Spearman $\rho$ between the system ranking from the top-$k$ oracle-selected tasks and the ranking from all tasks. Benchmarks sorted by $k{=}3$ fidelity (descending).}
\label{tab:per-bench-rho}
\begin{tabular}{lrcccc}
\toprule
Benchmark & Tasks & $k{=}1$ & $k{=}3$ & $k{=}5$ & $k{=}10$ \\
\midrule
  Widesearch & 100 & 0.959 & 0.988 & 0.988 & 0.997 \\
  SWE-Lancer & 94 & 0.921 & 0.978 & 0.966 & 0.971 \\
  SWE-bench-verified & 94 & 0.912 & 0.973 & 0.977 & 0.964 \\
  DA-Code & 164 & 0.928 & 0.968 & 0.971 & 0.976 \\
  SpreadsheetBench & 196 & 0.918 & 0.966 & 0.967 & 0.967 \\
  HLE & 195 & 0.935 & 0.966 & 0.974 & 0.987 \\
  LAB-Bench & 173 & 0.928 & 0.963 & 0.968 & 0.967 \\
  DeepSynth & 35 & 0.926 & 0.962 & 0.980 & 0.989 \\
  Omni-math & 160 & 0.879 & 0.962 & 0.957 & 0.966 \\
  SkillsBench & 68 & 0.961 & 0.962 & 0.974 & 0.982 \\
  SWT Bench & 49 & 0.881 & 0.956 & 0.931 & 0.962 \\
  GAIA & 157 & 0.909 & 0.953 & 0.959 & 0.950 \\
  ResearchCodeBench & 191 & 0.927 & 0.949 & 0.949 & 0.949 \\
  LiveCodeBench & 96 & 0.903 & 0.944 & 0.958 & 0.962 \\
  TerminalBench2.0 & 83 & 0.884 & 0.943 & 0.943 & 0.946 \\
  CRUST-Bench & 96 & 0.902 & 0.940 & 0.946 & 0.964 \\
  GPQA Diamond & 181 & 0.894 & 0.939 & 0.947 & 0.956 \\
  LawBench & 181 & 0.956 & 0.938 & 0.968 & 0.982 \\
  USACO & 99 & 0.876 & 0.938 & 0.940 & 0.957 \\
  FinanceAgent & 48 & 0.925 & 0.937 & 0.956 & 0.988 \\
  GSO & 90 & 0.912 & 0.937 & 0.970 & 0.947 \\
  GAIA2 & 68 & 0.910 & 0.936 & 0.974 & 0.962 \\
  ARC-AGI-2 & 95 & 0.922 & 0.933 & 0.948 & 0.982 \\
  Aider Polyglot & 221 & 0.907 & 0.932 & 0.954 & 0.955 \\
  AA-LCR & 94 & 0.884 & 0.932 & 0.900 & 0.908 \\
  BigCodeBench & 119 & 0.760 & 0.931 & 0.960 & 0.950 \\
  SWE-Bench Pro & 74 & 0.919 & 0.931 & 0.948 & 0.952 \\
  Spider 2 & 36 & 0.863 & 0.930 & 0.953 & 0.931 \\
  QCircuitBench & 25 & 0.868 & 0.929 & 0.962 & 0.915 \\
  StrongReject & 72 & 0.880 & 0.929 & 0.943 & 0.895 \\
  Scicode & 76 & 0.887 & 0.929 & 0.929 & 0.953 \\
  PIXIU & 87 & 0.859 & 0.926 & 0.921 & 0.950 \\
  FeatureBench & 156 & 0.878 & 0.926 & 0.908 & 0.966 \\
  AlgoTune & 147 & 0.876 & 0.924 & 0.947 & 0.926 \\
  KUMO & 195 & 0.889 & 0.921 & 0.960 & 0.969 \\
  SimpleQA & 196 & 0.859 & 0.919 & 0.919 & 0.921 \\
  BIX-Bench & 39 & 0.890 & 0.918 & 0.913 & 0.964 \\
  HumanEvalFix & 153 & 0.912 & 0.914 & 0.912 & 0.962 \\
  MMAU & 91 & 0.838 & 0.913 & 0.921 & 0.906 \\
  MMMLU & 127 & 0.908 & 0.909 & 0.911 & 0.909 \\
  SWE-Bench-Multilingual & 48 & 0.885 & 0.901 & 0.902 & 0.965 \\
  MedAgentBench & 70 & 0.890 & 0.899 & 0.927 & 0.930 \\
  AIME & 51 & 0.873 & 0.897 & 0.910 & 0.942 \\
  SWE-smith & 79 & 0.881 & 0.896 & 0.937 & 0.972 \\
  CompileBench & 15 & 0.909 & 0.894 & 0.971 & 0.989 \\
  Seal0 & 101 & 0.844 & 0.882 & 0.930 & 0.955 \\
  Reasoning Gym & 530 & 0.828 & 0.881 & 0.930 & 0.931 \\
  QuixBugs & 80 & 0.812 & 0.881 & 0.923 & 0.922 \\
  ReplicationBench & 79 & 0.867 & 0.862 & 0.916 & 0.940 \\
  IneqMath & 96 & 0.859 & 0.854 & 0.923 & 0.970 \\
  BFCL & 104 & 0.744 & 0.771 & 0.771 & 0.821 \\
  CyberGym & 10 & 0.707 & 0.745 & 0.940 & 0.999 \\
\bottomrule
\end{tabular}
\end{table}}

\providecommand{\AppGoToTaskTable}{\begin{table}[H]\small
\centering
\caption{Best single representative task per benchmark, ranked by useful representativeness score (leave-one-out correlation $\times$ task variance). Full list in \texttt{task\_representative\_tasks.csv}.}
\label{tab:go-to-tasks}
\begin{tabular}{llc}
\toprule
Benchmark & Best representative task & Score \\
\midrule
ResearchCodeBench & \texttt{eomt\_scale\_block\_forward} & 0.46 \\
GAIA & \texttt{ed58682d-bc52-4baa-9eb0-4eb81e1edacc} & 0.42 \\
FinanceAgent & \texttt{financeagent-29} & 0.41 \\
LAB-Bench & \texttt{figqa-0128} & 0.41 \\
SWE-Bench Pro & \texttt{instance\_navidrome\_\_navidrome-09ae41a...} & 0.41 \\
ARC-AGI-2 & \texttt{8f215267\_0} & 0.41 \\
SWE-Bench-Multilingual & \texttt{swe-bench/swebench\_multilingual\_\_redi...} & 0.41 \\
HLE & \texttt{hle/hle\_\_671f1f4ae38f776acdad8a77} & 0.41 \\
Omni-math & \texttt{omnimath\_3346} & 0.41 \\
TerminalBench2.0 & \texttt{sqlite-db-truncate} & 0.41 \\
GSO & \texttt{gso-pandas-dev--pandas-2421931} & 0.41 \\
DeepSynth & \texttt{69} & 0.40 \\
Aider Polyglot & \texttt{polyglot\_python\_food-chain} & 0.40 \\
FeatureBench & \texttt{mwaskom\_\_seaborn.7001ebe7.test\_bar.12...} & 0.39 \\
SWE-bench-verified & \texttt{pydata\_\_xarray-6721} & 0.38 \\
GPQA Diamond & \texttt{24} & 0.38 \\
SWT Bench & \texttt{matplotlib\_\_matplotlib-24177} & 0.38 \\
Spider 2 & \texttt{quickbooks002} & 0.38 \\
USACO & \texttt{132} & 0.38 \\
SpreadsheetBench & \texttt{438-18} & 0.37 \\
PIXIU & \texttt{pixiu-mlesg-mlesg10} & 0.37 \\
SWE-smith & \texttt{oauthlib\_\_oauthlib.1fd52536.combine\_f...} & 0.37 \\
BIX-Bench & \texttt{bix-61-q2} & 0.36 \\
Reasoning Gym & \texttt{reasoning-gym-algorithmic-jugs-easy00...} & 0.36 \\
ReplicationBench & \texttt{disk\_ridges\_\_ridge\_slope} & 0.36 \\
\bottomrule
\end{tabular}
\end{table}}

\providecommand{\GlobalGoToNumTasks}{5856}
\providecommand{\GlobalGoToIndependent}{400}
\providecommand{\AppGlobalGoToTaskTable}{\begin{table}[H]\small
\centering
\caption{Top 25 globally representative go-to tasks across all benchmarks, ranked by global useful representativeness (leave-one-out correlation with the full-pool aggregate $\times$ task variance). Full list in \texttt{task\_global\_representatives.csv}.}
\label{tab:global-go-to-tasks}
\begin{tabular}{llcc}
\toprule
Benchmark & Task & Score & Difficulty \\
\midrule
GAIA & \texttt{ed58682d-bc52-4baa-9eb0-4eb81e1e...} & 0.38 & medium \\
FinanceAgent & \texttt{financeagent-48} & 0.38 & medium \\
HLE & \texttt{hle/hle\_\_6728ba13fbd2af689fc469e5} & 0.38 & medium \\
PIXIU & \texttt{pixiu-mlesg-mlesg10} & 0.38 & medium \\
SpreadsheetBench & \texttt{36277} & 0.38 & medium \\
Aider Polyglot & \texttt{polyglot\_java\_simple-linked-list} & 0.38 & medium \\
DeepSynth & \texttt{35} & 0.38 & medium \\
HLE & \texttt{hle/hle\_\_66fec7825e6051260840e060} & 0.37 & medium \\
BIX-Bench & \texttt{bix-61-q2} & 0.37 & medium \\
FinanceAgent & \texttt{financeagent-26} & 0.37 & medium \\
DeepSynth & \texttt{69} & 0.37 & medium \\
SpreadsheetBench & \texttt{48969} & 0.37 & medium \\
Aider Polyglot & \texttt{polyglot\_javascript\_killer-sudok...} & 0.37 & medium \\
FinanceAgent & \texttt{financeagent-40} & 0.37 & medium \\
SpreadsheetBench & \texttt{50811} & 0.37 & medium \\
TerminalBench2.0 & \texttt{feal-differential-cryptanalysis} & 0.37 & medium \\
ARC-AGI-2 & \texttt{58490d8a\_0} & 0.37 & medium \\
LAB-Bench & \texttt{figqa-0024} & 0.37 & medium \\
Aider Polyglot & \texttt{polyglot\_python\_zipper} & 0.37 & easy \\
BIX-Bench & \texttt{bix-12-q5} & 0.37 & easy \\
SWE-Bench-Multilingual & \texttt{swe-bench/swebench\_multilingual\_...} & 0.37 & medium \\
SWE-Bench Pro & \texttt{instance\_navidrome\_\_navidrome-09...} & 0.37 & easy \\
SpreadsheetBench & \texttt{14240} & 0.37 & medium \\
Reasoning Gym & \texttt{reasoning-gym-cognition-figlet-f...} & 0.37 & medium \\
GPQA Diamond & \texttt{97} & 0.37 & medium \\
\bottomrule
\end{tabular}
\end{table}}

\providecommand{\AppGlobalGreedySequence}{\begin{table}[H]
\centering
\caption{Global greedy task selection: tasks added in order of decreasing independence across the full pool.}
\label{tab:global-greedy-sequence}
\small
\begin{tabular}{llr}
\toprule
Benchmark & Task & $\max|\rho|$ \\
\midrule
Reasoning Gym & \texttt{reasoning-gym-algorithmic-c...} & 0.00 \\
Reasoning Gym & \texttt{reasoning-gym-graphs-shorte...} & 0.00 \\
LAB-Bench & \texttt{figqa-0109} & 0.00 \\
Aider Polyglot & \texttt{polyglot\_python\_grep} & 0.04 \\
AA-LCR & \texttt{aa-lcr/aa-lcr-95} & 0.05 \\
AIME & \texttt{aime\_ii-12} & 0.06 \\
AA-LCR & \texttt{aa-lcr/aa-lcr-44} & 0.13 \\
HLE & \texttt{hle/hle\_\_67244f264d59b659ef...} & 0.18 \\
LawBench & \texttt{lawbench-2-7-3-zero-shot} & 0.18 \\
GPQA Diamond & \texttt{194} & 0.20 \\
\bottomrule
\end{tabular}
\end{table}}

\providecommand{\GoToSetNumTasks}{159}
\providecommand{\GoToSetNumBenchmarks}{53}
\providecommand{\AppGoToTaskSetTable}{\begin{table}[H]\small
\centering
\caption{Combined go-to task set: top-3 independently informative and representative tasks per benchmark (greedy $\max|\rho|{<}0.7$, then ranked by useful representativeness). Showing first 30 of 159 tasks across 53 benchmarks.}
\label{tab:goto-task-set}
\begin{tabular}{llccc}
\toprule
Benchmark & Task & Repr.\ score & Greedy step & Difficulty \\
\midrule
AA-LCR & \texttt{aa-lcr/aa-lcr-19} & 0.33 & 32 & easy \\
AA-LCR & \texttt{aa-lcr/aa-lcr-51} & 0.30 & 45 & easy \\
AA-LCR & \texttt{aa-lcr/aa-lcr-32} & 0.29 & 46 & easy \\
\addlinespace
Aider Polyglot & \texttt{polyglot\_javascript\_rest-api} & 0.35 & 2 & easy \\
Aider Polyglot & \texttt{polyglot\_javascript\_meetup} & 0.34 & 36 & medium \\
Aider Polyglot & \texttt{polyglot\_python\_two-bucket} & 0.31 & 42 & medium \\
\addlinespace
AIME & \texttt{aime\_i-14} & 0.21 & 8 & medium \\
AIME & \texttt{aime\_ii-11} & 0.19 & 10 & easy \\
AIME & \texttt{aime\_ii-9} & 0.19 & 11 & easy \\
\addlinespace
AlgoTune & \texttt{algotune-count-riemann-zeta-zeros} & 0.32 & 64 & medium \\
AlgoTune & \texttt{algotune-dynamic-assortment-plan...} & 0.24 & 4 & medium \\
AlgoTune & \texttt{algotune-fft-cmplx-scipy-fftpack} & 0.24 & 58 & medium \\
\addlinespace
ARC-AGI-2 & \texttt{7b3084d4\_0} & 0.27 & 9 & hard \\
ARC-AGI-2 & \texttt{db695cfb\_0} & 0.26 & 2 & medium \\
ARC-AGI-2 & \texttt{e3721c99\_1} & 0.23 & 10 & hard \\
\addlinespace
BFCL & \texttt{bfcl-live-simple-123-79-0} & 0.22 & 29 & medium \\
BFCL & \texttt{bfcl-live-parallel-multiple-8-7-0} & 0.19 & 36 & easy \\
BFCL & \texttt{bfcl-live-multiple-404-140-0} & 0.18 & 5 & easy \\
\addlinespace
BigCodeBench & \texttt{bigcodebench\_618} & 0.27 & 20 & medium \\
BigCodeBench & \texttt{bigcodebench\_752} & 0.25 & 2 & easy \\
BigCodeBench & \texttt{bigcodebench\_139} & 0.24 & 58 & easy \\
\addlinespace
BIX-Bench & \texttt{bix-12-q4} & 0.32 & 24 & medium \\
BIX-Bench & \texttt{bix-24-q2} & 0.29 & 27 & medium \\
BIX-Bench & \texttt{bix-41-q4} & 0.29 & 22 & easy \\
\addlinespace
CompileBench & \texttt{jq-static-musl} & 0.29 & 4 & easy \\
CompileBench & \texttt{jq-static} & 0.28 & 5 & easy \\
CompileBench & \texttt{coreutils} & 0.23 & 6 & easy \\
\addlinespace
CRUST-Bench & \texttt{crustbench-2dpartint} & 0.32 & 13 & easy \\
CRUST-Bench & \texttt{crustbench-dict} & 0.31 & 3 & easy \\
CRUST-Bench & \texttt{crustbench-quadtree} & 0.31 & 26 & easy \\
\bottomrule
\end{tabular}
\end{table}}

\providecommand{\GoToOverlapCount}{10}
\providecommand{\GoToOverlapTotal}{54}
\providecommand{\GoToOverlapPct}{19}

\providecommand{\UnifiedHoldoutNumBenchmarks}{54}
\providecommand{\UnifiedWithinBetter}{37}
\providecommand{\UnifiedCrossBetter}{17}
\providecommand{\UnifiedWithinBetterPct}{69}
\providecommand{\UnifiedMedianWithinImp}{12.1}
\providecommand{\UnifiedMedianCrossImp}{2.3}
\providecommand{\UnifiedMedianGlobalImp}{4.2}

\providecommand{\TaskHoldoutNumBenchmarks}{53}
\providecommand{\TaskHoldoutNumImproved}{41}
\providecommand{\TaskHoldoutMedianImprovement}{24.3}
\providecommand{\TaskHoldoutBestOneName}{HumanEvalFix}
\providecommand{\TaskHoldoutBestOneImp}{67.3}
\providecommand{\TaskHoldoutWorstOneName}{CyberGym}
\providecommand{\TaskHoldoutBestTwoName}{KUMO}
\providecommand{\TaskHoldoutBestTwoImp}{56.7}
\providecommand{\TaskHoldoutWorstTwoName}{BIX-Bench}
\providecommand{\TaskHoldoutBestThreeName}{LiveCodeBench}
\providecommand{\TaskHoldoutBestThreeImp}{56.3}
\providecommand{\TaskHoldoutWorstThreeName}{LawBench}

\providecommand{\TaskGreedyHighIndepOneName}{CyberGym}
\providecommand{\TaskGreedyHighIndepOnePct}{70}
\providecommand{\TaskGreedyLowIndepOneName}{HumanEvalFix}
\providecommand{\TaskGreedyHighIndepTwoName}{BIX-Bench}
\providecommand{\TaskGreedyHighIndepTwoPct}{69}
\providecommand{\TaskGreedyLowIndepTwoName}{ResearchCodeBench}

\providecommand{\AppGLMMTable}{\begin{table}[H]\small
\centering
\caption{LMM fixed effects (Eq.~\ref{eq:glmm}).  $\hat{\sigma}^2_u = \GLMMBenchmarkVar{}$, $\hat{\sigma}^2_e = \GLMMResidualVar{}$, ICC $= \GLMMIcc{}$.  Significance: {*}$p<0.05$, {**}$p<0.01$, {***}$p<0.001$.}
\label{tab:glmm-effects}
\begin{tabular}{llrrrr}
\toprule
Factor & Level & Estimate & SE & $z$ & $p$ \\
\midrule
  Intercept & baseline & +0.472 & 0.033 & 14.26 & 4.1e-46*** \\
  Model & claude-haiku-4-5-20251001 & +0.000 & \multicolumn{3}{c}{(ref.)} \\
  Model & claude-opus-4-6 & +0.179 & 0.017 & 10.35 & 4.2e-25*** \\
  Model & claude-sonnet-4-6 & +0.140 & 0.017 & 8.14 & 4.1e-16*** \\
  Model & gemini-3-flash-preview & +0.142 & 0.021 & 6.90 & 5.3e-12*** \\
  Model & gemini-3.1-pro-preview & +0.235 & 0.021 & 11.43 & 2.8e-30*** \\
  Model & gpt-5-mini & -0.002 & 0.020 & -0.12 & 0.904 \\
  Model & gpt-5-nano & -0.216 & 0.020 & -10.84 & 2.2e-27*** \\
  Model & gpt-5.4 & +0.129 & 0.020 & 6.49 & 8.5e-11*** \\
  Harness & claude-code & +0.000 & \multicolumn{3}{c}{(ref.)} \\
  Harness & codex & +0.045 & 0.020 & 2.27 & 0.023* \\
  Harness & gemini-cli & -0.037 & 0.022 & -1.64 & 0.101 \\
  Harness & terminus-2 & -0.042 & 0.014 & -2.97 & 0.003** \\
\bottomrule
\end{tabular}
\end{table}}



This appendix details the quantitative analyses in Section 3.1. We analyze the $\AppNumSystems{} \times \AppNumBenchmarks{}$
score matrix to study cross-benchmark
redundancy, within-benchmark redundancy, model vs.\ harness effects, benchmark
progress over time, cost evaluation, and execution-latency patterns.

\subsection{Benchmark and Task Predictability Analysis}
\label{app:benchpress}
\subsubsection{Low-rank structure for benchmark space}
Singular Value Decomposition (SVD) in logit space on the \AppNumModelsOnly{}-row Terminus-2
submatrix (varying only the base
model under the fixed Terminus-2 harness) yields PC1 = \AppSVDModelOnlyPCOne{}\% of
variance, which is comparable to BenchPress's 71\%~\cite{papailiopoulos2026benchpress} on their 31$\times$10 block with scaffold-free
evaluations, confirming that a single ``general capability'' axis dominates across
\AppNumModelsOnly{} models spanning capability tiers.  The first two components together capture
\AppSVDModelOnlyPCOneTwo{}\%, leaving little independent signal beyond two dimensions.



\begin{figure}[t]
  \centering
  \includegraphics[width=\linewidth]{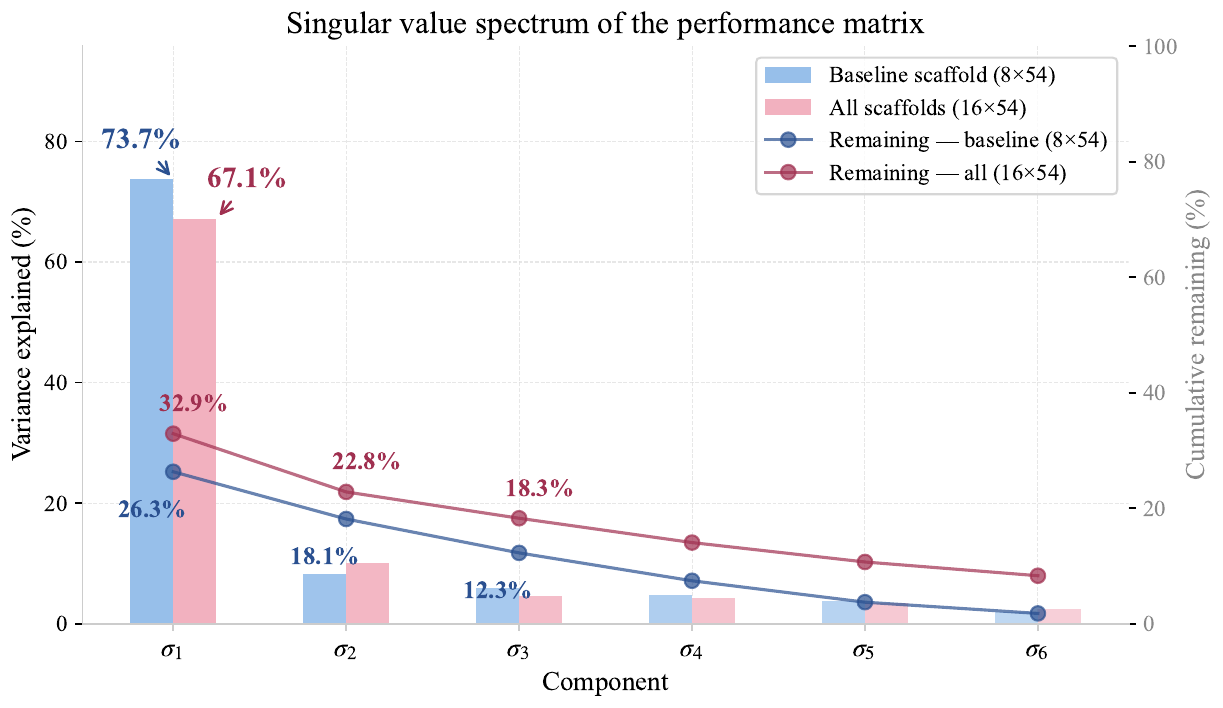}
  \caption{Singular value spectrum of the benchmark score matrix in logit space.  Blue:
\AppNumModelsOnly{}-row Terminus-2 submatrix (one fixed harness).  Purple: full
\AppNumSystems{}-row matrix (all model--harness configurations).  The drop in PC1 share and rise in PC2 when
native harnesses are included reflects the orthogonal harness-effect dimension.}
  \label{fig:svd-spectrum}
\end{figure}

On the full \AppNumSystems{}-row matrix (all model--harness configurations), the spectrum shifts:
  PC1 drops to \AppSVDFullPCOne{}\% and PC2 rises to \AppSVDFullPCTwo{}\%      
  (Figure~\ref{fig:svd-spectrum}). PC2 cleanly separates the \AppNumFamilyHarnessSystems{}
  configurations using native harnesses from the \AppNumTerminusSystems{} Terminus-2 configurations with zero
  overlap (mean gap \AppPCTwoGap{} standard units). Together the first two components
  capture \AppSVDFullCumTwo{}\% of variance, with every remaining component below 6\%:  the benchmark space is effectively rank-2, one axis for model capability and one for  
  the harness effect.

\subsubsection{Benchmark predictability.}
\label{par:benchmark-predictability}
Following BenchPress~\cite{papailiopoulos2026benchpress}, we quantify per-benchmark redundancy by predicting
held-out scores from the remaining benchmarks.  Our
setting differs in two respects: rows are model--harness configurations rather than models alone, and
all \AppNumBenchmarks{} benchmark metrics are normalized scores rescaled and bounded in
$[0,1]$. \label{par:holdout}  Given the fully-observed score
matrix $\mathbf{S} \in \mathbb{R}^{N \times B}$, for each of $T{=}5$ random seeds we hold out 50\%
of each row's benchmark scores as the test set $\mathcal{H}$; the remaining 50\% form the
training set $\mathcal{T}$.  Both prediction methods operate in logit space:
\begin{equation}
  z = \operatorname{logit}(s) = \log\!\frac{\,\mathrm{clip}(s,\,\epsilon,\,1{-}\epsilon)\,}
       {1 - \mathrm{clip}(s,\,\epsilon,\,1{-}\epsilon)},
  \quad \epsilon = 0.005,
  \label{eq:logit}
\end{equation}
where $\epsilon$ clips extreme scores away from 0 and 1 to avoid infinite logits.
The predictor combines three steps:
\begin{itemize}[leftmargin=*]
  \item \textbf{LogitBenchReg}: for each target benchmark $j$, fit univariate ordinary least 
squares (OLS) regressions in
    logit space from every other benchmark $k \ne j$:
    \begin{equation}
      z_{i,j} = \beta_k\, z_{i,k} + \gamma_k, \quad
      \text{fitted over } \{i : (i,j) \in \mathcal{T} \wedge (i,k) \in \mathcal{T}\},
      \label{eq:benchreg-ols}
    \end{equation}
    where $z_{i,j} = \mathrm{logit}(\mathrm{clip}(s_{i,j},\, \epsilon,\, 1{-}\epsilon))$ with
    $\epsilon = 0.005$ (Eq.~\ref{eq:logit}).  We select the top-$K{=}5$ predictors with
    $R^2_k \ge 0.1$ as the predictor set $\mathcal{K}_j$.  For each held-out cell
    $(i,j) \in \mathcal{H}$, the predicted score is:
    \begin{equation}
      \hat{z}_{i,j} = \frac{\sum_{k \in \mathcal{K}_j} R^2_k\, (\beta_k\, z_{i,k} + \gamma_k)}
                           {\sum_{k \in \mathcal{K}_j} R^2_k},
      \qquad
      \hat{s}_{i,j}^{\mathrm{BenchReg}} = \frac{1}{1 + e^{-\hat{z}_{i,j}}}.
      \label{eq:benchreg-predict}
    \end{equation}

  \item \textbf{SVD-Logit (Soft-Impute)}: iterative low-rank matrix completion on the
    logit-transformed, column-standardized score matrix.  Let $\mu_j$ and $\sigma_j$ be the column
    mean and standard deviation from training entries only, and $\tilde{\mathbf{Z}} =
    (\mathbf{Z} - \boldsymbol{\mu}) / \boldsymbol{\sigma}$ with held-out entries initialized to~0.
    The iteration:
    \begin{enumerate}[nosep]
      \item Compute the rank-$r$ SVD approximation ($r{=}2$):
        $\tilde{\mathbf{Z}}_{\mathrm{approx}} = \mathbf{U}_{:,1:r}\,
        \boldsymbol{\Sigma}_{1:r}\, \mathbf{V}_{1:r,:}^\top$.
      \item Replace only held-out entries:
        $\tilde{Z}_{i,j} \gets (\tilde{\mathbf{Z}}_{\mathrm{approx}})_{i,j}$ for $(i,j) \in
        \mathcal{H}$. Training entries remain fixed.
      \item Repeat until convergence:
        $\|\tilde{\mathbf{Z}}^{(t)} - \tilde{\mathbf{Z}}^{(t-1)}\|_F \big/
        \|\tilde{\mathbf{Z}}^{(t-1)}\|_F < 10^{-4}$.
    \end{enumerate}
    After convergence, de-standardize and inverse-logit:
    $\hat{s}_{i,j}^{\mathrm{SVD}} = 1/(1+e^{-(\tilde{z}_{i,j} \cdot \sigma_j + \mu_j)})$.

  \item \textbf{Blending and evaluation}: the final prediction combines both in raw score space:
    \begin{equation}
      \hat{s}_{i,j} = \alpha\, \hat{s}_{i,j}^{\mathrm{BenchReg}}
                    + (1{-}\alpha)\, \hat{s}_{i,j}^{\mathrm{SVD}},
      \quad \alpha = 0.6.
      \label{eq:blend}
    \end{equation}
    When LogitBenchReg produces no prediction (no predictor passes $R^2 \ge 0.1$), the method falls
    back to SVD-Logit alone; when both fail, it uses the column mean $\bar{s}_j$.
    All predictions are evaluated on raw scores in $[0,1]$.  The primary metric is the
    \emph{benchmark-stratified} Median Absolute Error:
    \begin{equation}
      \mathrm{MedAE}_j = \operatorname{median}_{(i,j)\in\mathcal{H}_j}
        |\hat{s}_{i,j} - s_{i,j}|,
      \label{eq:medae-bench}
    \end{equation}
    where $\mathcal{H}_j$ is the set of held-out cells for benchmark~$j$ (pooled across all seeds).
    The overall MedAE is the median across benchmarks:
    \begin{equation}
      \mathrm{MedAE} = \operatorname{median}_{j=1,\ldots,B}\; \mathrm{MedAE}_j.
      \label{eq:medae}
    \end{equation}
    Because all benchmark scores share the $[0,1]$ scale, MedAE is directly interpretable as
    the typical prediction error in percentage points.
\end{itemize}

We evaluate on our \AppNumSystems{}$\times$\AppNumBenchmarks{} matrix
(50\% holdout, 3~folds).
The simplest baseline predicts each held-out cell with the column (benchmark) mean of the
observed entries, achieving MedAE~$=$~\AppBaselineMedAE{}.
Methods that exploit cross-benchmark structure improve substantially:
LogitBenchReg reaches \AppLogitBenchRegMedAE{},
SVD-Logit ($r{=}2$) reaches \AppSVDLogitMedAE{},
and the BenchPress blend reaches \textbf{\AppBlendMedAE{}},
a \AppBlendImprovement{}\% reduction over the baseline.
The fact that one benchmark's scores can predict another's to within
\AppBlendMedAE{} points confirms substantial cross-benchmark redundancy.

\AppPerBenchmarkPredictTable{}

Benchmarks that resist prediction include \BenchPressUniqueOne{} (MedAE = \BenchPressUniqueOneScore{}),
\BenchPressUniqueTwo{} (\BenchPressUniqueTwoScore{}), and \BenchPressUniqueThree{}
(\BenchPressUniqueThreeScore{}), measuring capability axes poorly covered by the remaining suite.
By contrast, \BenchPressRedundantOne{} (MedAE = \BenchPressRedundantOneScore{}) and
\BenchPressRedundantTwo{} (\BenchPressRedundantTwoScore{}) are well-reconstructed from peers,
adding little independent signal.

\subsubsection{How many benchmarks span the evaluation space?}
\label{par:greedy-selection}
\textit{Only \GreedyBenchmarksBelowSeven{} of \AppNumBenchmarks{} benchmarks contribute independent 
signal at the $|\rho|<0.7$ threshold.}
Greedy forward selection starts from the benchmark with the lowest mean absolute correlation to all 
others and iteratively adds the benchmark whose maximum absolute correlation with the 
already-selected set is the smallest.  The first \GreedyBenchmarksBelowSeven{} selections all 
satisfy 
$\max|\rho| < 0.7$; from step~\GreedyFirstAboveSevenStep{} onward, every remaining benchmark has 
$|\rho| \ge 0.7$ with at least one already-selected benchmark.

\AppBenchGreedySequence{}
The first redundant benchmark (\GreedyFirstAboveSeven{}, step~\GreedyFirstAboveSevenStep{}, 
$\rho{=}\GreedyFirstAboveSevenCorr{}$) is largely explained by the already-selected set.  For 
practitioners with constrained budgets, $\sim$\GreedyBenchmarksBelowSeven{} well-chosen benchmarks 
across domains capture most independent signal in the full \AppNumBenchmarks{}-benchmark suite.

\textit{Difficulty and uniqueness are largely decoupled.}
The benchmarks identified as independently informative span the full difficulty spectrum---from
hard (CodePDE, mean score 0.36) to easy (StrongReject, 0.92)---rather than clustering at one
end (Table~\ref{tab:greedy-sequence}). This confirms that uniqueness is driven by the
\emph{distinctiveness} of the capability axis a benchmark measures, not by raw task difficulty.

\subsubsection{Does the same redundancy hold at the task level?}
\label{par:task-level}

We extend the SVD analysis to the \AppNumSystems{} $\times$ \TaskGoodTasks{}-task matrix
(after filtering $\TaskTotalTasks{} - \TaskGoodTasks{} = \TaskZeroVarTasks{}$ zero-variance tasks where all model--harness configurations scored identically).
At the task level, PC1 explains only \TaskFullPCOne{}\% of variance, compared with
\AppSVDFullPCOne{}\% at the benchmark level.
Benchmark aggregation therefore compresses away independent dimensions: the task-level
score matrix contains richer structure that a single overall ranking cannot capture.

\textit{Within-benchmark task redundancy varies widely.}
\label{par:intra-bench}
For each of the \TaskRedundNumBenchmarks{} benchmarks with $\ge 10$ tasks, we run PCA on the
\AppNumSystems{} $\times$ $T$ sub-matrix, where $T$ is the number of tasks in that benchmark (after removing zero-variance tasks, averaging
\TaskRedundMeanZeroVarPct{}\% per benchmark). PC1 alone explains \TaskRedundMeanPCOne{}\%
of within-benchmark variance on average, and a median of \TaskRedundMedianPCsNinety{}
components suffices for 90\% (out of a maximum of \AppNumSystems{} dimensions).
The degree of redundancy varies considerably across benchmarks:
\begin{itemize}[nosep]
  \item \textbf{\TaskLowRankExTwoName{}} (\TaskLowRankExTwoTasks{} tasks): PC1 =
\TaskLowRankExTwoPCOne{}\%, only \TaskLowRankExTwoPCs{}~components for 90\%.
  \item \textbf{\TaskLowRankExThreeName{}} (\TaskLowRankExThreeTasks{} tasks): PC1 =
\TaskLowRankExThreePCOne{}\%, \TaskLowRankExThreePCs{}~components for 90\%.
  \item \textbf{\TaskLowRankExOneName{}} (\TaskLowRankExOneTasks{} tasks): PC1 =
\TaskLowRankExOnePCOne{}\%, \TaskLowRankExOnePCs{}~components for 90\%.
\end{itemize}
Selecting the $k$ tasks most correlated with the full benchmark mean and re-ranking the
\AppNumSystems{} systems gives the ranking fidelity shown in Table~\ref{tab:task-redundancy}.
Individual benchmarks range from WideSearch ($\rho = 0.99$) to CyberGym ($\rho = 0.75$),
reflecting differences in internal task diversity; the complete per-benchmark breakdown
is given in Table~\ref{tab:per-bench-rho}.

\begin{table}[!htbp]
  \centering
  \caption{Ranking fidelity from a small task subset vs.\ the full benchmark.
  Values are mean Spearman $\rho$ across \TaskRedundNumBenchmarks{} benchmarks with $\ge 10$ tasks.}
  \label{tab:task-redundancy}
  \begin{tabular}{ccc}
    \toprule
    $k$ tasks & Mean $\rho$ & Benchmarks with $\rho > 0.95$ \\
    \midrule
    1  & \TaskRedundRhoKOne{}  & \TaskRedundAboveNFKOne{}  \\
    3  & \TaskRedundRhoKThree{} & \TaskRedundAboveNFKThree{} \\
    5  & \TaskRedundRhoKFive{}  & \TaskRedundAboveNFKFive{}  \\
    10 & \TaskRedundRhoKTen{}   & \TaskRedundAboveNFKTen{}   \\
    \bottomrule
  \end{tabular}
\end{table}

\AppPerBenchRhoTable{}

\textit{Within-benchmark holdout prediction.}
\label{par:task-holdout}
Because the full \TaskGoodTasks{}-task matrix has far more columns than rows
(\AppNumSystems{}), cross-benchmark prediction would overfit.
We therefore predict tasks \emph{within} each benchmark: for each of the
\TaskHoldoutNumBenchmarks{} benchmarks with $\ge 5$ reliable tasks, we mask 50\% of
each row (3 folds) and blend top-$k$ peer correlation with rank-2 SVD.
The blend improves over the task-mean baseline in \TaskHoldoutNumImproved{} cases
(median improvement \TaskHoldoutMedianImprovement{}\%), with the largest gains in
benchmarks whose tasks test similar capabilities, such as \TaskHoldoutBestOneName{}
(\TaskHoldoutBestOneImp{}\%) and \TaskHoldoutBestTwoName{} (\TaskHoldoutBestTwoImp{}\%).

\textit{Representative task selection.}
\label{par:per-bench-goto}
Greedy forward selection within each benchmark identifies independently informative
tasks ($\max|\rho| < 0.7$).  The fraction of independent tasks varies widely:
\TaskGreedyHighIndepOneName{} (\TaskGreedyHighIndepOnePct{}\%) and
\TaskGreedyHighIndepTwoName{} (\TaskGreedyHighIndepTwoPct{}\%) contain many
independent items, while \TaskGreedyLowIndepOneName{} and
\TaskGreedyLowIndepTwoName{} reach the $\rho > 0.7$ threshold within the first few selections,
indicating that most of their tasks measure the same underlying capability.
Within-benchmark blends outperform cross-benchmark peers in
\UnifiedWithinBetter{}/\UnifiedHoldoutNumBenchmarks{} (\UnifiedWithinBetterPct{}\%)
cases, but for the remaining \UnifiedCrossBetter{} benchmarks cross-benchmark peers
predict better, indicating shared capability axes across benchmark boundaries.


\subsection{Model vs.\ Harness Effect: Methodology}
\label{app:model-vs-agent}

To formally test whether model or harness identity explains more variance while
controlling for benchmark difficulty, we fit an LMM:
\begin{equation}
  s_{ij} = \beta_0 + \boldsymbol{\beta}_{\mathrm{model}}\, x_{\mathrm{model},i}
  + \boldsymbol{\beta}_{\mathrm{harness}}\, x_{\mathrm{harness},i}
  + u_j + \varepsilon_{ij},
  \quad u_j \sim \mathcal{N}(0, \sigma^2_u),
  \quad \varepsilon_{ij} \sim \mathcal{N}(0, \sigma^2_e),
  \label{eq:glmm}
\end{equation}
where:
\begin{itemize}[nosep,leftmargin=*]
  \item $s_{ij}$: raw score for system $i$ (model--harness configuration) on benchmark $j$
  \item $\beta_0$: intercept (expected score at reference levels)
  \item $\boldsymbol{\beta}_{\mathrm{model}}$, $\boldsymbol{\beta}_{\mathrm{harness}}$: fixed-effect
coefficients for model and harness identity
  \item $x_{\mathrm{model},i}$, $x_{\mathrm{harness},i}$: dummy-coded indicators (first alphabetical
level as reference)
  \item $u_j \sim \mathcal{N}(0, \sigma^2_u)$: benchmark-level random intercept (difficulty 
heterogeneity)
  \item $\varepsilon_{ij} \sim \mathcal{N}(0, \sigma^2_e)$: residual error
  \item $\sigma^2_u$: between-benchmark variance; $\sigma^2_e$: within-benchmark variance
\end{itemize}
Estimated via REML using \texttt{statsmodels} \texttt{MixedLM}.

The intraclass correlation coefficient (ICC) measures the proportion of total variance attributable
to between-benchmark differences:
\begin{equation}
  \mathrm{ICC} = \frac{\sigma^2_u}{\sigma^2_u + \sigma^2_e} = \GLMMIcc{},
  \label{eq:icc}
\end{equation}
confirming that \GLMMIcc{} of score variance is between benchmarks---i.e., benchmark difficulty
is the dominant source of variation, justifying the random-intercept specification.

Table~\ref{tab:glmm-effects} reports the estimated fixed effects. Each coefficient
represents the difference from the reference level (the alphabetically first model or
harness, whose effect is zero by construction).
Among the \NumModels{} models, \GLMMNumSigModels{} coefficients are significant
($p < 0.05$); among the \NumHarnesses{} harnesses, only \GLMMNumSigHarnesses{}.
We define the \emph{effect range} for a factor as the difference between its largest and
smallest coefficients (including the reference at zero). The model effect range is
$\GLMMMaxModelEffect{} - (\GLMMMinModelEffect{}) = \GLMMModelEffectRange{}$ raw score units, while the
harness effect range is $\GLMMMaxHarnessEffect{} - (\GLMMMinHarnessEffect{}) = \GLMMHarnessEffectRange{}$.
The model range is thus \GLMMModelHarnessRangeRatio{}$\times$ larger, confirming that model
choice dominates harness choice across benchmarks.

\AppGLMMTable

\subsection{Original benchmark data collections and mapping}
\label{app:benchmark-data}

To contextualize the current Harbor results, we collected historical scores of different model and/or agents over time, along with their timestamps, for each benchmark from the original paper, public leaderboards, and technical reports of new models.

In some benchmarks, multiple evaluation metrics are reported. We manually select the metric that best aligns with our Harbor evaluation results, or transform the reported metric into a Harbor-aligned score to enable direct comparison. In particular, we apply special treatment to the following papers:
\begin{itemize}
    \item \textbf{CodePDE}: The paper reports nRMSE as the primary metric \cite{li2025codepde}, where lower is better. We convert nRMSE into a Harbor-aligned pass rate by thresholding each PDE family at $\mathrm{nRMSE}<0.05$ and averaging the resulting pass rates across the five families.

    \item \textbf{AlgoTune}: Raw speedup factors are mapped to the normalized Harbor score using the benchmark's log-speedup compression, $\log(\max\{1,x\})/(\log(\max\{1,x\})+1)$, where $x$ denotes the speedup ratio. This transformation ensures that: (1) the score always lies in $[0,1)$; (2) the model receives a non-zero score only when $x>1$, i.e., when the agent-generated code is faster than the baseline; and (3) the score increases monotonically with $x$, so faster code leads to a higher score. In the original paper \cite{press2025algotune}, the overall benchmark score is defined as the harmonic mean of acceleration ratios. We therefore collect the per-task results for each model and recompute the score using our formula, yielding a normalized score in $[0,1)$.

    \item  \textbf{SLDBench}: Raw verifier rewards can lie in $[-1,1]$, so we report the display-aligned score $(x+1)/2$ on the $[0,1]$ scale.

    \item \textbf{QuixBugs}: Historical results are reported as the number of repaired programs, which we convert to pass rate by dividing by the 40 benchmark programs.

    \item \textbf{StrongReject}: Public results report jailbreak/attack success, while Harbor reports defense success. We therefore invert the metric as $1-x$.

    \item \textbf{KUMO}: Public results are split into easy and hard settings. To match the subsampled Harbor task mixture, we compute a weighted success rate, $(202 \cdot \mathrm{accuracy}_{\mathrm{easy}} + 10 \cdot \mathrm{accuracy}_{\mathrm{hard}}) / 212$.

    \item \textbf{SciCode}: Harbor uses the 80-task without-background setup and aggregates sub-step correctness at the problem level, so we align to the paper's standard subproblem pass@1 rather than other SciCode variants.

    \item \textbf{ResearchCodeBench}: the official metric is scaled pass@1, which weights each code snippet by its number of executable lines of code (LoC). We therefore align Harbor evaluation to the same LoC-weighted aggregation, instead of averaging snippet-level success uniformly.

    \item \textbf{Spider 2.0}, \textbf{SpreadsheetBench}, \textbf{SWT-Bench}, \textbf{USACO}, and \textbf{WideSearch}: public scores are matched to the Harbor-evaluated split or subset when possible, rather than to the full benchmark headline number.
\end{itemize}

FinanceAgent, GAIA2, LawBench, MLGym, PIXIU, SkillsBench, and SWE-smith were excluded from direct historical comparison when no comparable public metric or split could be identified.

We note that existing benchmark evaluations may not align with our model--harness configurations: some benchmarks evaluate raw LLM performance without any agentic scaffold; some fine-tune specialized models or design task-specific agentic scaffolds, while our study evaluates models spanning capability tiers from the Gemini, Claude, and GPT families under general-purpose harnesses (Gemini CLI, Claude Code, Codex, and Terminus-2). We present in Table~\ref{tab:external_overlap} a benchmark-level summary of this alignment for the 54 benchmarks in our taxonomy.

\begin{table}[!h]
\centering
\caption{Benchmark-level overlap between public benchmark evaluations and our Harbor model--harness configurations. Columns are not mutually exclusive: a benchmark can contain both overlapping and non-overlapping public configurations.}
    \label{tab:external_overlap}
    \begin{tabular}{@{}lccc@{}}
    \toprule
    & Has overlap with ours & Has non-Harbor configuration & Total \\ \midrule
    Evaluation includes raw LLMs & 10 & 31 & 38 \\
    Evaluation includes agents & 7 & 27 & 33 \\ \midrule
    Unique benchmarks & 16 & 52 & 54 \\ \bottomrule
\end{tabular}
\end{table}

Overall, only 16 of the 54 benchmarks have at least one overlapping model or harness configuration. Therefore, these results are often not suitable for direct head-to-head comparison against our Harbor runs. Nevertheless, they remain useful for tracking progress: by taking the best reported performance at different time points, we can estimate how the state of the art has advanced on each benchmark over time, which will be discussed in Appendix \ref{app:overtime}.

\subsection{Performance Progress Over Time}
\label{app:overtime}

\begin{figure}[t]
  \centering
  \includegraphics[width=\linewidth]{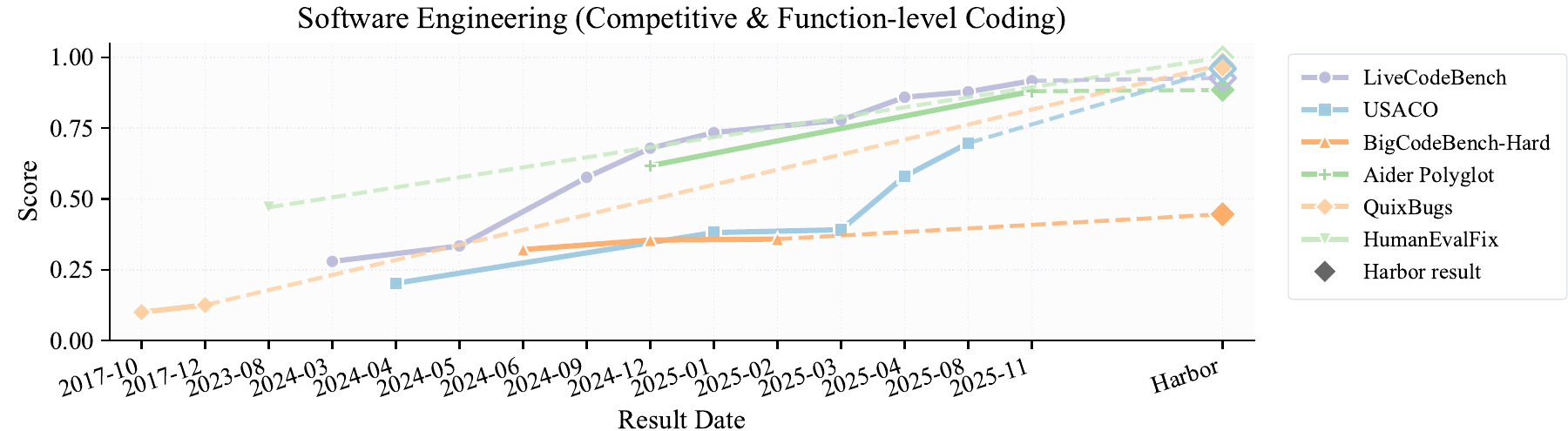}
  \includegraphics[width=\linewidth]{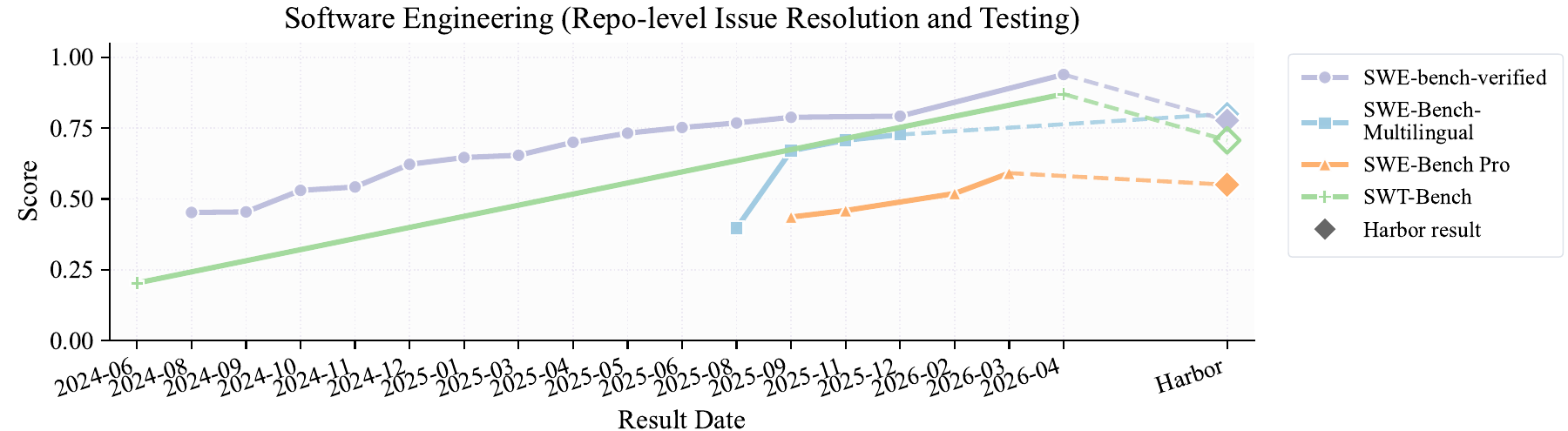}
  \includegraphics[width=\linewidth]{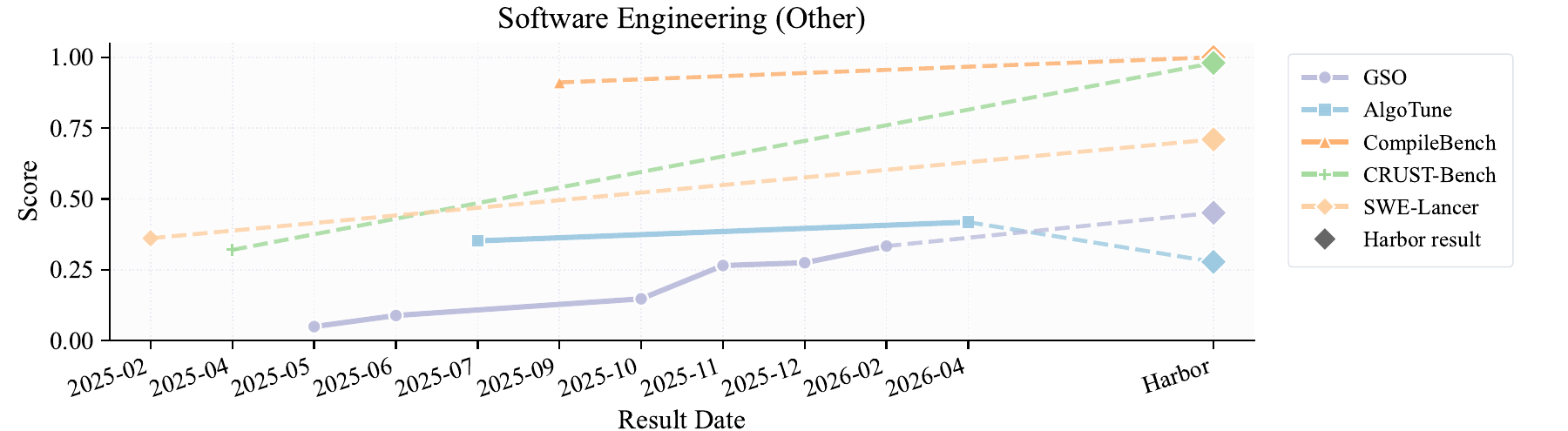}
    \caption{\textbf{Progress over time across Software Engineering benchmarks.} Each panel tracks the best reported score on representative benchmarks within a domain as a function of result date; the rightmost diamond marks the best frontier result measured in our Harbor evaluation under a unified adapter-based setup. Hollow diamonds indicate cases where the Harbor evaluation set is not identical to the original benchmark, typically because Harbor evaluates a subset, so the connected segment should be interpreted as a reference comparison rather than a strict apples-to-apples continuation. For LiveCodeBench, the historical evaluation results are based on live evaluation, while Harbor uses the fixed LiveCodeBench (V6).}
  \label{fig:progress-over-time-swe}
\end{figure}

\begin{figure}[t]
  \centering
  \includegraphics[width=\linewidth]{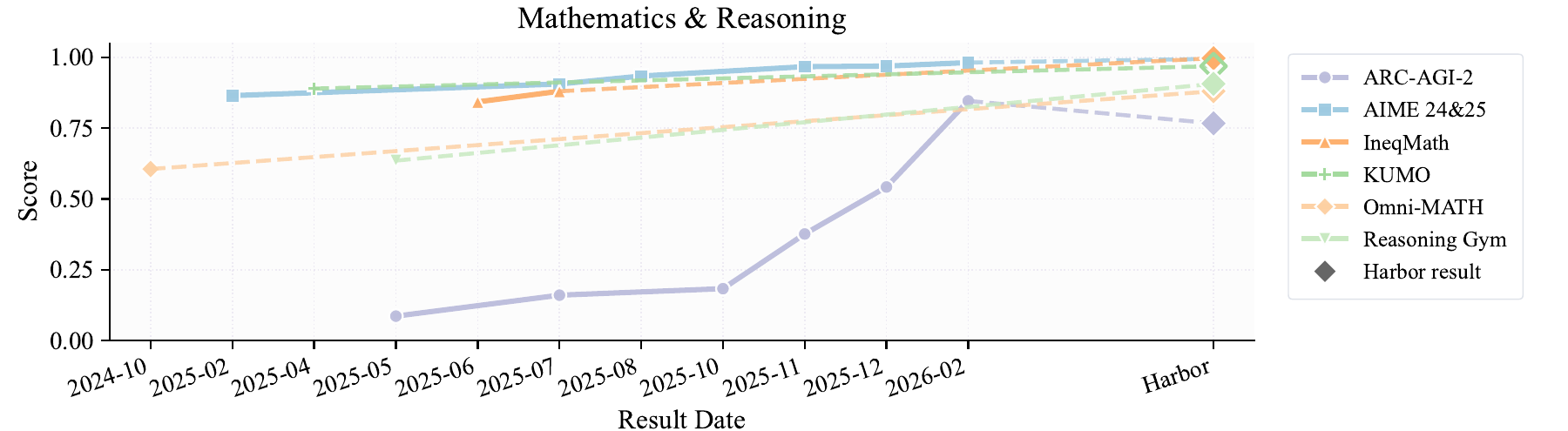}
  \includegraphics[width=\linewidth]{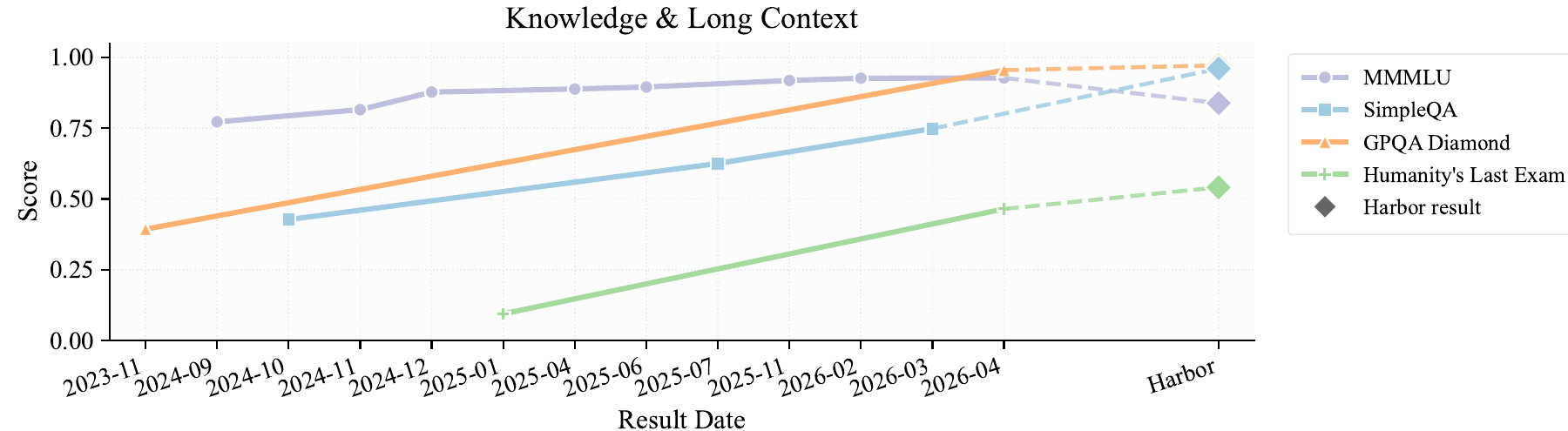}
  \includegraphics[width=\linewidth]{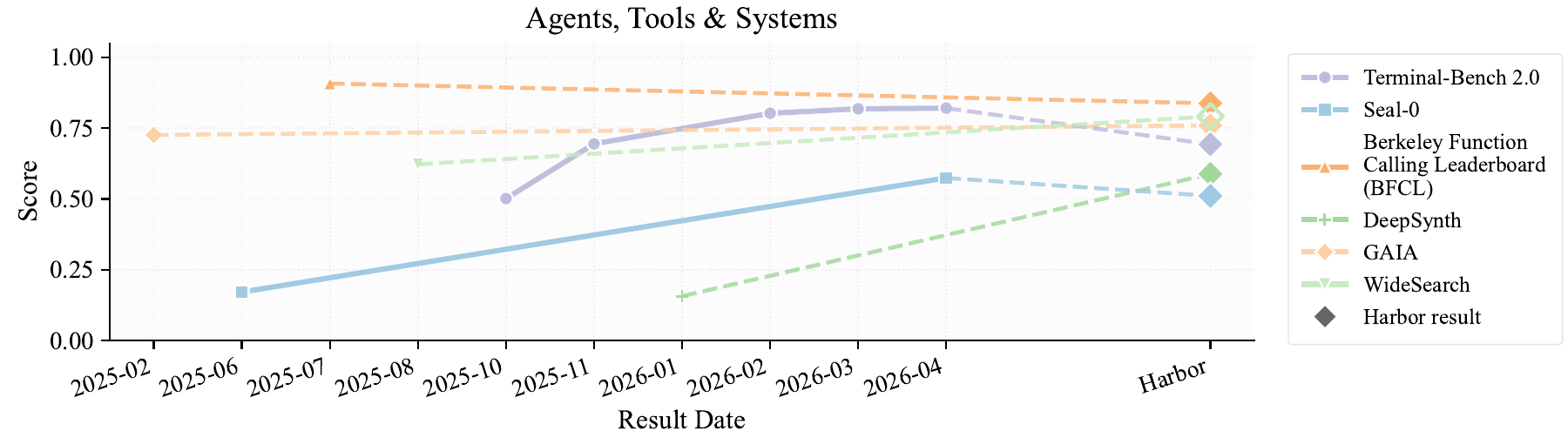}
  \includegraphics[width=\linewidth]{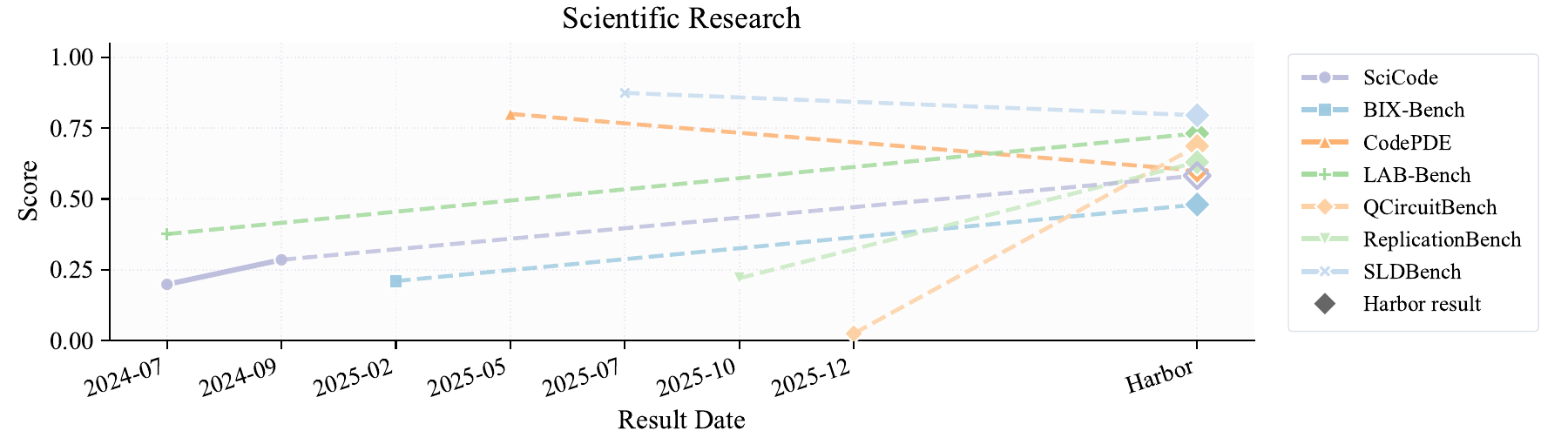}
    \caption{\textbf{Progress over time across Mathematics \& Reasoning, Knowledge \& Long Context, Agents, Tools \& Systems, and Scientific Research benchmarks.} Each panel tracks the best reported score on representative benchmarks within a domain as a function of result date; the rightmost diamond marks the best frontier result measured in our Harbor evaluation under a unified adapter-based setup. Hollow diamonds indicate cases where the Harbor evaluation set is not identical to the original benchmark, typically because Harbor evaluates a subset, so the connected segment should be interpreted as a reference comparison rather than a strict apples-to-apples continuation.}
  \label{fig:progress-over-time-other}
\end{figure}

As described in Appendix \ref{app:benchmark-data}, we have collected historical scores of different model and agent configurations over time. In Figures~\ref{fig:progress-over-time-swe} and \ref{fig:progress-over-time-other}, we present the best reported performance for each benchmark over time. While this view should only be interpreted as an observational record of community progress, the density and slope of these curves reveal where the community has concentrated its benchmarking effort, and where progress has been most visible.

A clear pattern is that Mathematics \& Reasoning and Software Engineering have attracted the most sustained leaderboard pressure. These domains contain many closely tracked benchmarks, frequent leaderboard updates, and rapid replacement of the previous state of the art. In Mathematics \& Reasoning, scores on several benchmarks rise quickly toward saturation, suggesting that once a benchmark becomes widely adopted, frontier labs and open-source systems rapidly optimize against it -- potentially because these tasks have clear, verifiable evaluation protocols. Software Engineering shows an even more heterogeneous version of the same trend: function-level and competitive-programming-style benchmarks often improve rapidly and approach saturation, while several broader engineering tasks remain substantially more challenging. This suggests that the apparent progress in ``coding'' is not uniform; it is stronger on well-specified, unit-testable tasks, and weaker on open-ended tasks requiring repository understanding, environment management, and long-horizon debugging.

By contrast, progress is less dense and often slower in domains such as Knowledge \& Long Context and Agents, Tools \& Systems. These areas have fewer historical points, fewer continuously maintained leaderboards, and less frequent apples-to-apples reporting. The resulting curves are therefore sparser, but this sparsity is itself informative: it indicates that these capabilities have not yet received the same level of repeated public measurement as math and software engineering. In Knowledge \& Long Context, many benchmarks are expensive to evaluate, sensitive to retrieval or context-window assumptions, and harder to compare across model generations. In Agents, Tools \& Systems, evaluation often depends on changing external environments, tool APIs, and multi-step execution protocols, making leaderboard progress more difficult than for static math or coding tasks.

These differences highlight an important source of bias in how progress is perceived. Domains with dense public leaderboards create a feedback loop: frequent measurement encourages optimization, which produces rapid score gains, which in turn attracts more attention. Conversely, sparse domains may appear to progress more slowly partly because they are harder to evaluate consistently and partly because fewer systems are repeatedly measured there. The fastest-moving curves identify areas where the field has built strong benchmarking infrastructure, while flatter or sparser curves point to domains where better standardized evaluation may be a prerequisite for faster scientific progress. Through our infrastructure efforts in this work, we hope that agentic evaluation can become less dependent on a small number of highly visible leaderboards and more evenly distributed across capability domains.

For several benchmarks, the best results reported externally are higher than our best results measured under Harbor, as shown in Table~\ref{tab:original-launch-vs-harbor}. These gaps arise because the external results were obtained via specialized models (e.g., on BFCL), specialized agent scaffolds (e.g., on MedAgentBench, SLDBench, and AlgoTune), and larger compute budgets for agents (e.g., on CodePDE). By contrast, Harbor evaluates a fixed set of models spanning capability tiers under a unified adapter-based setup, with the goal of improving cross-benchmark comparability rather than maximizing performance on each individual benchmark.

\begin{table}[t]
\caption{\textbf{Comparison with original or launch-time best results on selected benchmarks.} Scores are reported on a $[0,1]$ scale.}
\label{tab:original-launch-vs-harbor}

\centering
\begin{tabular}{lccccc}
\toprule
 & BFCL & SLDBench & CodePDE & MedAgentBench & AlgoTune \\
\midrule
Original / launch best & 0.9072 & 0.8741 & 0.8000 & 0.6967 & 0.3401 \\
Harbor best            & 0.8374 & 0.7952 & 0.6000 & 0.6433 & 0.2776 \\
\bottomrule
\end{tabular}

\end{table}


\subsection{Latency Study}
\label{app:token_study}
\begin{figure}[htbp]
    \centering
    \includegraphics[width=\linewidth]{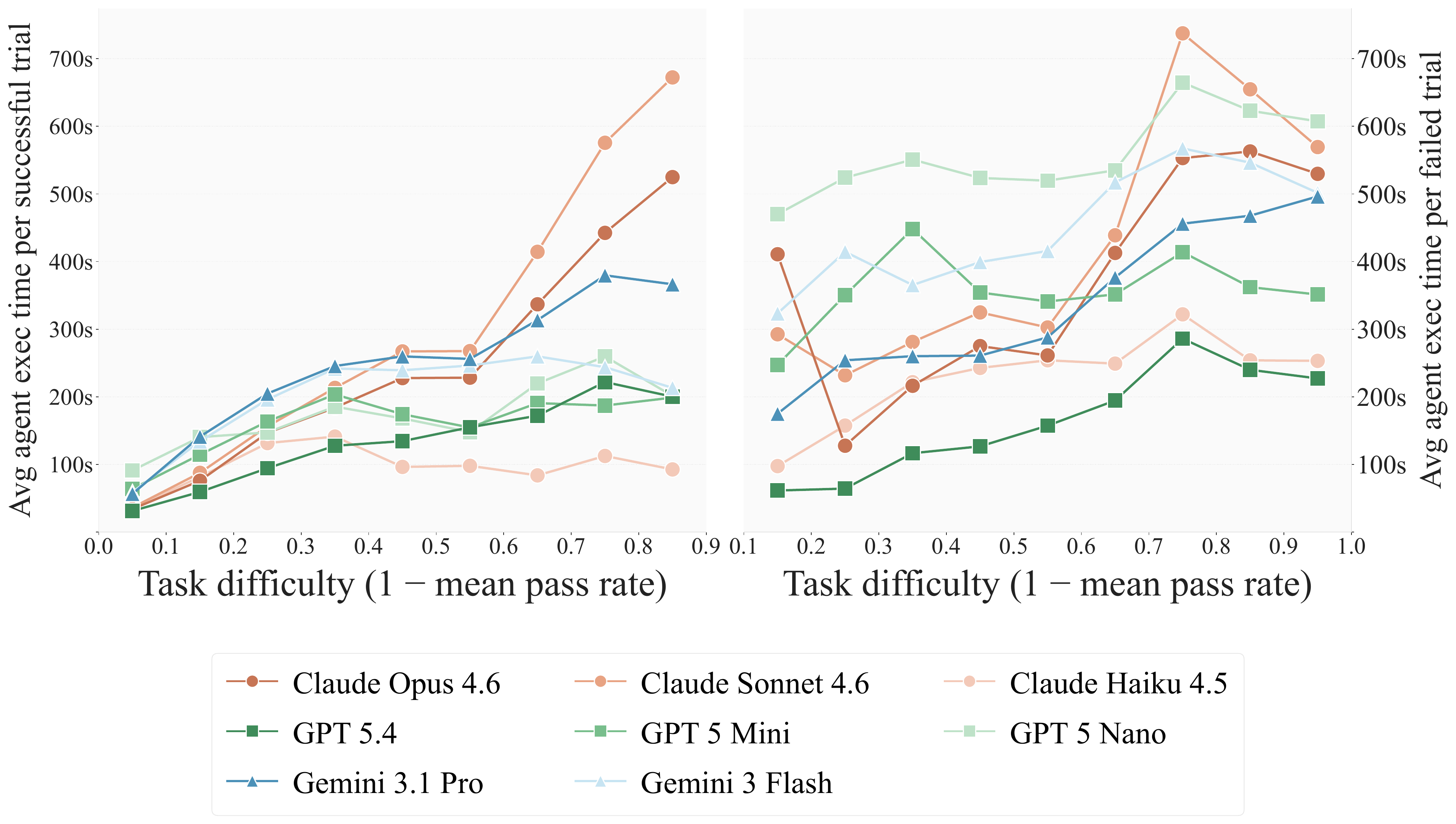}
    \caption{\textbf{Left:} Average agent execution time per successful trial. \textbf{Right:} Average agent execution time per failed trial. Per-model breakdown by task difficulty. The 0.9--1.0 bin is omitted from the success panel (most models have fewer than 25 successful trials) and the 0.0--0.1 bin from the failure panel (four models have fewer than 25 failed trials).}
    \label{fig:latency_pass_fail}
\end{figure}

Agent execution latency, the wall-clock time a model spends actively working on a task, reflects inference time, tool calls, environment interaction, and retry behavior. Using the same difficulty buckets as in \Cref{sec:main_token}, we disaggregate latency by trial outcome to ask: do all models spend time the same way when they succeed versus when they fail?

Figure~\ref{fig:latency_pass_fail} plots per-model execution time on successful trials (left) and failed trials (right). On the success side, models follow a broadly similar upward trend with difficulty, but separate into a high-latency group (Claude Sonnet 4.6, Claude Opus 4.6) and a low-latency group (GPT-5.4, Claude Haiku 4.5, GPT-5-Mini, Gemini 3 Flash), with the gap widening on harder tasks.
 
The failure side reveals two findings invisible in the success panel. First, there is a shared structural pattern across all models that failure latency peaks around difficulty 0.7--0.8 and then declines on the hardest tasks, the opposite of the success trend. This suggests that the hardest tasks trigger earlier termination, while tasks in the 0.7--0.8 range are plausible enough to sustain extended, ultimately unproductive attempts. Second, the relationship between success and failure latency is not consistent across models. GPT-5.4 terminates in similar time regardless of outcome, whereas GPT-5-Nano, despite being the weakest model, persists longest on failed trials.

\section{Case Analysis}

\subsection{Task Failure}
\label{app:task-failures}
The task-failure examples below illustrate the kinds of broken designs identified during Harbor-Index human review (Section~\ref{sec:harbor-index}). Each category appears across multiple benchmarks in the review set.

\paragraph{Broken environments.}
All four \textsc{gaia2} tasks that reached human review share the same defect: the adaptive simulation event needed to trigger the grading cascade never fires. One Claude-code trajectory ran for 270 steps, spent 36 hours but received only \texttt{\{``notifications'': []\}} throughout. The verifier waits for the simulation to advance in response to agent actions, while the adapter never advances it. Every model and harness fails for the same reason, ruling out agent capability as the cause.

On \textsc{mmau}, the Harbor adapter discards the audio payload before handing each task to the agent. Agents work from a text transcript and are graded as though they heard the audio, so tasks requiring acoustic judgement (emotion, speaker identity) cannot be solved regardless of model strength.

\paragraph{Instruction-verification mismatch.}
A task only has a clear pass condition when the instruction and the verifier agree on what success looks like. When they do not, an agent can produce a defensible solution and still score zero. We see this pattern in several forms across the review set.

In one \textsc{swebench-pro} Ansible task, 162 of 168 tests pass for most agents. The six that fail require a specific dictionary format absent from the problem description. A \textsc{teleport} task adds a new test file (\texttt{åTestAuditWriter/Backoff}) in the gold patch that is absent from the starting repository. All eleven capable agents implement the backoff mechanism correctly per the written specification, but the verifier's restore command does not include the test file the gold patch introduced.

In other cases the verifier holds a standard the instruction never states. A \textsc{humanity's last exam} task was rejected because the judge carries an internal reference answer that disagrees with the mathematically correct value; across all 18 trials, the judge rejected the correct answer with 152--280 completion tokens of active reasoning per run. An \textsc{aa-lcr} task rejects ``30.4 percentage points'' while accepting ``0.304,'' applying a unit-representation rule that does not appear in the instruction.

The mismatch can also originate on the instruction side. A \textsc{swebench-multilingual} task gives agents a raw GitHub bug report (\textit{Describe the bug, To Reproduce, Expected behavior, Your Configuration}) but no directive to fix anything. Agents have to guess from the report format that a fix is expected. An \textsc{aider-polyglot} JavaScript task shows the matrix transpose with newline-separated strings, which suggests a string contract, but the hidden verifier expects arrays. No test file is present in the working tree, so agents cannot discover the discrepancy before submitting.

\paragraph{Gameable tests.}
A \textsc{featurebench-modal} task produces byte-identical failure output (53 fails / 1 pass) across 14 runs from 5 distinct harness and model combinations. The test file imports the implementation under a specific symbol name and module path the instruction does not disclose. Every agent that uses a different but functionally correct structure fails in the same way. The task is measuring whether an agent guesses the right import path, not whether it can implement the feature.

\paragraph{Wrong gold answer.}
Several \textsc{aa-lcr} tasks were rejected because the gold answer is simply wrong. In each case, all \MainEvalConfigurationCount{} evaluated configurations spanning \MainEvalHarnessImplementationCount{} distinct harness implementations independently reached the same alternative value, and manual inspection confirmed the agent answer is correct. When every frontier model agrees and the gold disagrees, the gold needs to be checked.

\subsection{Agent Failure Modes}

\subsubsection{Methodology}
\label{app:qual-method}
The agent failure-mode taxonomy is bootstrapped \emph{bottom-up} rather than authored top-down. We draw a stratified sample of failed trajectories spanning every benchmark, model, and harness in our pool, and dispatch a Claude-4.7-Opus subagent to each trajectory in parallel. Each subagent reads the task's instruction, the verifier output, and the agent's full ATIF trajectory, and emits a free-form, single-sentence description of the failure root cause---e.g., ``deleted a previously-passing test rather than fixing the bug,'' ``terminated declaring success without running the verifier,'' ``failed to install a benchmark-specific dependency.'' The free-form descriptions are then clustered into a small set of candidate failure modes, and the authors review and merge near-duplicates, yielding $16$ candidate rubrics. Each rubric carries a short definition and several anchor examples drawn from the bootstrap descriptions, so that downstream human annotators and an LLM judge can apply the same definitions. Below we describe the human-annotation protocol, the gold-label adjudication rule, and the LLM-judge calibration that follows.

\paragraph{Human annotation.} A held-out sample of $182$ failed trials, stratified across benchmark, model, and harness, is annotated independently by two non-overlapping panels of three annotators each (Round 1 and Round 2). Each annotator works in Docent\footnote{Docent is a system for inspecting and annotating model trajectories: \url{https://transluce.org/introducing-docent}.} on a per-trial UI: they read the same task instruction, verifier output, and rendered ATIF trajectory the LLM judge will see, and apply each of the $18$ candidate rubrics independently with a binary verdict (\textsc{match} / \textsc{no match}) and a free-text explanation citing concrete trajectory evidence. We measure inter-annotator agreement by comparing the two rounds.

\paragraph{Inter-rater agreement.} Cohen's $\kappa$ between Round 1 and Round 2 is reported per rubric in Table~\ref{tab:human-irr}. The pooled $\kappa$ across all $18$ rubrics is $0.66$, and twelve of the eighteen rubrics individually clear the $\kappa \geq 0.5$ ``moderate-or-better'' threshold. The remaining six --- Calculation Mismatch (R1 over-applies it as a numerical-Q\&A catch-all), Rule Inference Error (low base rate, R2 over-applies on ARC-style tasks), Specification--Verification Mismatch (R2 over-applies on under-specified tasks), and three rubrics that R1 never used at all (Insufficient Web Research, Premature Termination, Wrong Output Location) --- are dropped from headline reporting and folded into an ``Others'' bucket.

\paragraph{Gold-label adjudication.} We then take the union of the two human annotations to generate the final gold label for calibrating the LLM judge. For each $(\textit{trial}, \textit{rubric})$ cell, the gold label is \textsc{match} iff at least one Round-1 \emph{or} Round-2 annotator selects a \textsc{match}. This union rule yields a recall-leaning gold set: a single annotator who can quote concrete trajectory evidence (a verifier-output snippet, a tool-call message number, a quoted line of agent code) is sufficient to flag a cell.

\begin{table}[h]
\centering\footnotesize
\caption{Round-1 vs.\ Round-2 inter-annotator Cohen's $\kappa$ on $182$ trials with labels in both rounds. Sorted by $\kappa$. Twelve rubrics clear the $\kappa \geq 0.5$ trustworthy threshold; six are excluded.}
\label{tab:human-irr}
\begin{tabular}{lrrrl}
\toprule
\textbf{Rubric} & \textbf{R1+} & \textbf{R2+} & \textbf{$\kappa$} & \textbf{Tier} \\
\midrule
Wrong Target / Layer & 2 & 2 & 1.000 & Trustworthy \\
Syntax / Language Error & 3 & 4 & 0.854 & Trustworthy \\
Agent Timeout & 8 & 9 & 0.815 & Trustworthy \\
Unverified Claim & 27 & 21 & 0.809 & Trustworthy \\
Plan Over Implementation & 2 & 3 & 0.797 & Trustworthy \\
Wrong Factual Answer & 37 & 31 & 0.784 & Trustworthy \\
Hidden-Test Regression & 22 & 14 & 0.755 & Trustworthy \\
False Success Claim & 23 & 16 & 0.743 & Trustworthy \\
Silent Deliverable & 6 & 3 & 0.659 & Trustworthy \\
Algorithmic Bug & 25 & 26 & 0.659 & Trustworthy \\
Environment Block & 7 & 8 & 0.652 & Trustworthy \\
Wrong Output Schema & 12 & 11 & 0.583 & Trustworthy \\
\midrule
Calculation Mismatch & 22 & 7 & 0.451 & Excluded \\
Rule Inference Error & 2 & 8 & 0.389 & Excluded \\
Specification--Verification Mismatch & 16 & 34 & 0.320 & Excluded \\
Insufficient Web Research & 0 & 3 & 0.000 & Excluded \\
Premature Termination & 0 & 3 & 0.000 & Excluded \\
Wrong Output Location & 0 & 1 & 0.000 & Excluded \\
\bottomrule
\end{tabular}
\end{table}

\paragraph{LLM-judge calibration.} The judge is a single multi-rubric shortlist prompt covering the $12$ trustworthy rubrics: each call shows the model the task context (instruction, verifier code, oracle solution, environment Dockerfile, verifier stdout), the full rendered ATIF trajectory, and all $12$ rubric definitions, and asks the model to shortlist $0$--$3$ rubrics that fire with one citation per fire. We tune the rubric texts (not the prompt template) iteratively against the gold set, alternating broaden/tighten edits per rubric. We test two judge models (\texttt{gemini-3-flash-preview} and \texttt{gemini-3.1-pro-preview}), and use the stronger Pro model after 14 rounds of prompt tuning.

A few patterns from $14$ tuning rounds:
\begin{itemize}\setlength{\itemsep}{2pt}
  \item One-rubric-at-a-time edits compose; multi-rubric edits do not. Tightening Rubric A's exclusions reliably shifts borderline cells away from A, but the same edit will typically over-correct on a related rubric (e.g.\ tightening False Success Claim drains positives into Unverified Claim) when applied alongside an unrelated change to a third rubric.
  \item Per-rubric sub-judges (one focused yes/no call per rubric per trial, $12\times$ the calls) under-perform the multi-rubric SHORTLIST prompt. The competition between rubrics for the shortlist slot is helpful: it forces the model to commit to a positive verdict rather than hedging \textsc{no match} on every rubric independently.
  \item Pro responds better to short Docent-base rubric text than to elaborate priority-firing language we wrote in some iterations. For Agent Timeout, dropping the override entirely (so Pro reads the original $200$-token rubric text) yields $\kappa = 0.48$; an elaborate override that explicitly reranks AT above content rubrics yields $\kappa = 0.0$ (the model never picks AT). Pro is conservative by default, and verbose ``do this even if you'd otherwise pick X'' instructions tend to confuse the model rather than improving rubric selection.
\end{itemize}

\paragraph{Calibration results.} The final chosen configuration achieves pooled $\kappa = 0.51$ against the $182$-trial gold set, vs.\ the human--human R1$\leftrightarrow$R2 ceiling of $0.66$ on the same trials. Six rubrics clear the $\kappa \geq 0.5$ trustworthy threshold individually; the remaining six less confident rubrics are grouped into the ``Others'' bucket. Per-rubric numbers are given in Table~\ref{tab:judge-calibration}.

\begin{table}[h]
\centering\footnotesize
\caption{Pro v12 judge agreement with the gold set on the $182$-trial calibration sample. Trials where one or more parse failures dropped a verdict are excluded from the per-rubric counts. Six rubrics clear $\kappa \geq 0.5$; the rest are folded into ``Others''.}
\label{tab:judge-calibration}
\begin{tabular}{lrrrl}
\toprule
\textbf{Rubric} & \textbf{Gold+} & \textbf{LLM+} & \textbf{$\kappa$ vs.\ gold} & \textbf{Tier} \\
\midrule
Wrong Factual Answer & 37 & 49 & 0.69 & $\geq 0.5$ (reported) \\
Syntax / Language Error & 4 & 5 & 0.66 & $\geq 0.5$ (reported) \\
Silent Deliverable & 6 & 6 & 0.66 & $\geq 0.5$ (reported) \\
Hidden-Test Regression & 20 & 11 & 0.62 & $\geq 0.5$ (reported) \\
Environment Block & 10 & 5 & 0.52 & $\geq 0.5$ (reported) \\
Plan Over Implementation & 3 & 1 & 0.50 & $\geq 0.5$ (reported) \\
\midrule
Agent Timeout & 9 & 7 & 0.48 & Others \\
False Success Claim & 19 & 10 & 0.44 & Others \\
Algorithmic Bug & 28 & 35 & 0.37 & Others \\
Wrong Output Schema & 15 & 6 & 0.25 & Others \\
Unverified Claim & 21 & 1 & 0.08 & Others \\
Wrong Target / Layer & 2 & 2 & $-0.01$ & Others \\
\midrule
\textbf{Pooled} & \textbf{183} & \textbf{138} & \textbf{0.51} & --- \\
\bottomrule
\end{tabular}
\end{table}

The mean $\kappa$ across the six reported rubrics is $0.61$, essentially matching the human--human ceiling on the same six rubrics ($0.71$): on the rubrics where the judge clears the trustworthy bar, its calls are at human-rater quality, and the gap between $0.51$ pooled and the $0.66$ human ceiling is concentrated in the Others bucket --- particularly Unverified Claim (Pro under-fires by an order of magnitude), Wrong Output Schema (Pro under-fires when key-name mismatches are involved), and Wrong Target / Layer (low base rate, $n = 2$ in the gold sample).

\paragraph{Audit at scale.} The calibrated judge is then applied to all $5{,}898$ failed trials in the full hard-task set. We retry parse-failure trials up to three times with bumped \texttt{max\_tokens}, achieving $98.4\%$ parsed coverage; the remaining $1.6\%$ are mostly trials whose rendered transcript exceeds the gateway's per-request token cap and are dropped from headline numbers.

\paragraph{Rubric definitions.} The full set of $18$ candidate rubrics, with the definitions presented to human annotators and used in the LLM judge prompts, is given below. We mark each rubric as \textsc{Trustworthy} (passes inter-annotator $\kappa \geq 0.5$) or \textsc{Excluded} (folded into ``Others'').

\smallskip\noindent\textit{Trustworthy rubrics ($\kappa \geq 0.5$, reported individually).}

\noindent{\large\bfseries Wrong Target / Layer.}\quad \textsc{Trustworthy}\par
\smallskip

\textbf{Framing.}
Agent applied a change to a target / abstraction layer / module / code path where the change cannot produce the behavior the grader checks.

\textbf{Concrete patterns.}
\begin{itemize}\setlength{\itemsep}{2pt}
  \item Fix at global-stylesheet layer when the component uses inline / CSS-in-JS styles
  \item Test that asserts file-layout when the task required a behavioral test
  \item Patch to a codepath the failing test does not exercise
  \item Shallow symptom fix when root cause is deeper in state/logic
  \item Fix in a deprecated / unused module
\end{itemize}

\textbf{Decision procedure.}
\begin{enumerate}\setlength{\itemsep}{2pt}
  \item Identify the task's intended target.
  \begin{itemize}\setlength{\itemsep}{1pt}
    \item If no target location inferrable: \textsc{no match}.
  \end{itemize}
  \item Identify where the agent actually applied changes.
  \item \textsc{match} if (a) a change was applied AND (b) the applied change is structurally incapable of affecting the graded behavior.
\end{enumerate}

\textbf{Exclusions.}
\begin{itemize}\setlength{\itemsep}{2pt}
  \item Right file but logic wrong: \textsc{no match} (Algorithmic Bug).
  \item Minimal symptom fix catching some cases: \textsc{no match} (Algorithmic Bug).
\end{itemize}

\noindent{\large\bfseries Syntax / Language Error.}\quad \textsc{Trustworthy}\par
\smallskip

\textbf{Framing.}
Agent produced code / DSL artefact that doesn't parse or uses invalid domain constructs — Python SyntaxError, OpenQASM unsupported gates, PDDL malformed, Java NPE at load time, Go won't compile, YAML rejected.

\textbf{Decision procedure.}
\begin{enumerate}\setlength{\itemsep}{2pt}
  \item Identify the domain language.
  \item Look for compile / parse / load errors in tool outputs.
  \item \textsc{match} if the artefact fails at language-level.
\end{enumerate}

\textbf{Exclusions.}
\begin{itemize}\setlength{\itemsep}{2pt}
  \item Code compiled but logic wrong: \textsc{no match} (Algorithmic Bug).
  \item Runtime exception from execution (not parse-time): \textsc{no match} unless the exception prevents any test from running.
\end{itemize}

\noindent{\large\bfseries Algorithmic Bug.}\quad \textsc{Trustworthy}\par
\smallskip

\textbf{Framing.}
On a code-writing or bug-fixing task, the agent's overall APPROACH IS DEFENSIBLE but the implementation contains a SPECIFIC, NAMEABLE DEFECT — wrong axis, off-by-one, parameter swap, missed edge case, flawed API usage, wrong control-flow branch. You must be able to point to the bug.

\textbf{Decision procedure.}
\begin{enumerate}\setlength{\itemsep}{2pt}
  \item Confirm the task is code-writing or bug-fixing (not factual Q\&A, not Calc, not numerical-only).
  \item Confirm runnable code WAS produced and executed without an unhandled crash (no SyntaxError, no ImportError, no environment failure).
  \item Confirm reward=0.
  \item POINT TO THE BUG. You must quote either:  If you cannot quote a SPECIFIC defect — output 'no match'.
  \begin{itemize}\setlength{\itemsep}{1pt}
    \item The buggy line in the agent's code AND describe the specific defect (e.g., 'used i+1 instead of i-1 on line 42'), OR
    \item A specific failing test output that pinpoints the logical error (e.g., 'expected 5 got 4 on input [1,2,3,4]').
  \end{itemize}
\end{enumerate}

\textbf{Exclusions.}
\begin{itemize}\setlength{\itemsep}{2pt}
  \item Agent took the WRONG OVERALL APPROACH (e.g., downloaded pre-trained weights when the task said to fine-tune; implemented a feature in the wrong file; chose the wrong abstraction): \textsc{no match} (Wrong Target / Layer).
  \item Code did not parse / compile: \textsc{no match} (Syntax / Language Error).
  \item Sample tests passed but hidden tests failed: \textsc{no match} (Hidden-Test Regression).
  \item Numerical result outside tolerance: \textsc{no match} (Calculation Mismatch).
  \item No code emitted (only narrative / plan): \textsc{no match} (Plan Over Implementation / Silent Deliverable).
  \item Crash from missing dependency / broken environment: \textsc{no match} (Environment Block).
  \item Verifier's spec doesn't match the prompt: \textsc{no match} (Specification–Verification Mismatch).
  \item Wrong file/path/abstraction: \textsc{no match} (Wrong Target / Layer).
  \item Format / shape only: \textsc{no match} (Wrong Output Schema).
  \item Failure on a short-answer Q\&A task: \textsc{no match} (Wrong Factual Answer). If you would write 'the agent's solution must have a bug because reward=0' — that is NOT enough. Output 'no match'. You need to identify WHICH bug. This rubric is HIGH-PRECISION: prefer 'no match' unless you can name and quote the specific defect.
\end{itemize}

\noindent{\large\bfseries Wrong Factual Answer.}\quad \textsc{Trustworthy}\par
\smallskip

\textbf{Framing.}
The task asks for a SHORT, LITERAL ANSWER (a single value, a year, a name, an MCQ letter, a number with a fixed unit, a yes/no) and the agent emitted that answer in the required slot, but the value is wrong by a plain reading of the prompt.

\textbf{Decision procedure.}
\begin{enumerate}\setlength{\itemsep}{2pt}
  \item Confirm the deliverable is a short literal answer — not extended code, not a multi-step calculation pipeline that itself can have algorithmic bugs.
  \item Confirm the agent emitted a literal answer in the required slot (file, message, function call) and that the answer is concretely wrong (you can quote the agent's value AND the verifier-expected value).
  \item \textsc{match} only when steps 1 + 2 hold AND the failure isn't better explained by a more-specific rubric below.
\end{enumerate}

\textbf{Exclusions.}
\begin{itemize}\setlength{\itemsep}{2pt}
  \item Numerical answer outside tolerance from a calculation pipeline: \textsc{no match} (Calculation Mismatch).
  \item Code-writing task where reward=0 because of a bug in the code: \textsc{no match} (Algorithmic Bug).
  \item Format/shape mismatch (right content, wrong shape): \textsc{no match} (Wrong Output Schema).
  \item Wrong because of a rule-inference mistake on a pattern task: \textsc{no match} (Rule Inference Error).
  \item Missing answer / no deliverable: \textsc{no match} (Silent Deliverable).
  \item Agent could not verify against an available checker: \textsc{no match} (Unverified Claim). This rubric COVERS short-answer Q\&A, MCQ, simple lookups, even when the underlying reasoning is complex — as long as the deliverable is just the literal answer and the answer is wrong.
\end{itemize}

\noindent{\large\bfseries Wrong Output Schema.}\quad \textsc{Trustworthy}\par
\smallskip

\textbf{Framing.}
The agent's content is at least partly correct, but the FORMAT / SHAPE / ENCODING / KEY-NAMES are rejected by the verifier. This rubric covers a wide range: JSON shape mismatch, wrong scalar type (e.g. 4.0 vs 4, '0' vs 'zero'), expected string got number, missing required key, extra prose around the answer, wrong delimiters, markdown vs plain text, list-vs-dict, and key-name mismatches (e.g. snake\_case vs camelCase) where the verifier checks the literal key.

\textbf{Firing trigger.}
\begin{itemize}\setlength{\itemsep}{2pt}
  \item The verifier output contains shape/type/encoding error patterns: 'expected <type> got <type>', 'KeyError', 'invalid JSON', 'expected list got dict', 'AttributeError on dict.<method>', 'JSON decode error', schema-validation messages.
  \item The verifier complains the agent's output cannot be parsed.
  \item The verifier expected a specific key/name that the agent used a different spelling/case for, where the underlying values are recognizable.
  \item The agent emitted markdown / prose where a bare value was expected. Output \textsc{match} if ANY of the above patterns appears AND the agent's content is at least partly recognizable.
\end{itemize}

\textbf{Decision procedure.}
\begin{enumerate}\setlength{\itemsep}{2pt}
  \item Inspect verifier\_output for shape/type/encoding/key error patterns.
  \item Confirm the agent's content is at least partly correct — the right answer or recognizable pieces are present, just in the wrong shape.
  \item Quote the specific divergence (which field, what shape).
  \item \textsc{match} if 1+2+3 hold.
\end{enumerate}

\textbf{Exclusions.}
\begin{itemize}\setlength{\itemsep}{2pt}
  \item Content is wholly wrong, formatting is fine: \textsc{no match} (Algorithmic Bug / Wrong Factual Answer).
  \item Content correct, location wrong: \textsc{no match} (Wrong Output Location).
  \item No content produced at all: \textsc{no match} (Silent Deliverable).
\end{itemize}

\noindent{\large\bfseries Hidden-Test Regression.}\quad \textsc{Trustworthy}\par
\smallskip

\textbf{Framing.}
The agent ran SCAFFOLD-PROVIDED test infrastructure or correctness checks (sample tests, the repo's own test suite, a check script that ships with the task, a make-target the task points at, a CI command, a known oracle command) on its deliverable and they PASSED — but the verifier's hidden / fuller test suite still rejects with reward=0.

Key signal: the agent ran something that came WITH the task (not something the agent wrote / fabricated) and saw a positive result.

FIRING TRIGGER (high-recall): trajectory contains BOTH:
  (a) An invocation of a scaffold-shipped test/check (pytest tests/, make test, gradlew test, npm test, repo-provided sample inputs, a check script in the workspace), AND
  (b) A positive result from that invocation (the agent quoted 'passed', 'OK', '0 failed', 'tests passed', or similar).

If both fire and reward=0, output \textsc{match} unless a strong exclusion applies.

\textbf{Decision procedure.}
\begin{enumerate}\setlength{\itemsep}{2pt}
  \item Quote the scaffold-provided test/check the agent ran (with msg number).
  \item Quote the positive output the agent saw.
  \item Confirm reward=0.
  \item Confirm the test was scaffold-shipped, NOT agent-fabricated. Ad-hoc inline assertions the agent wrote do NOT qualify (those are False Success Claim).
  \item \textsc{match} when 1+2+3+4 hold.
\end{enumerate}

\textbf{Exclusions.}
\begin{itemize}\setlength{\itemsep}{2pt}
  \item Agent ran its own ad-hoc check (an inline \texttt{assert ...} it crafted, a smoke-test script, a manual eyeballing): \textsc{no match} (False Success Claim).
  \item Agent never ran any test: \textsc{no match} (Unverified Claim).
  \item Agent's tests failed and it ignored them: \textsc{no match}.
  \item Hidden tests enforce something genuinely absent from the prompt: \textsc{no match} (Specification–Verification Mismatch).
  \item Crash / syntax error: \textsc{no match} (Syntax / Language Error). This rubric is HIGH-RECALL: when you see scaffold-provided tests + positive result + reward=0, prefer 'match' over the broader False Success Claim or Unverified Claim. The signal is most specific.
\end{itemize}

\noindent{\large\bfseries False Success Claim.}\quad \textsc{Trustworthy}\par
\smallskip

\textbf{Framing.}
The agent SAW POSITIVE EVIDENCE for its deliverable (a passing assertion, a matching expected output, an analysis it concluded was correct, a successful smoke-test) and DECLARED THE TASK DONE — but the verifier disagrees with reward=0.

\textbf{Firing trigger.}
(a) The trajectory contains an agent-generated positive correctness signal: a quoted phrase like 'passed', 'OK', 'matches expected', 'all tests pass', 'correct', 'looks good', 'sanity check passes', or an inline assertion that returned True / didn't raise.
  (b) The agent declared completion (final 'task done', 'finish', 'submit', or stopped confidently after writing the deliverable).
  (c) Reward = 0.

If all three fire, output \textsc{match} unless a strong exclusion applies.

Note on the FSC vs HTR boundary: when the agent ran scaffold-shipped tests that passed, prefer Hidden-Test Regression (more specific). When the agent's positive signal came from its own ad-hoc check (an inline assertion / one-off script / manual eyeballing), prefer FSC.

\textbf{Decision procedure.}
\begin{enumerate}\setlength{\itemsep}{2pt}
  \item Find the agent-generated positive correctness signal in the trajectory and quote it (with msg number).
  \item Confirm completion declaration (or implicit stop) and reward = 0.
  \item If the positive signal came from scaffold-shipped tests (pytest, make test, check script that ships with the task), this is HTR territory — output \textsc{no match}.
  \item Otherwise output \textsc{match}.
\end{enumerate}

\textbf{Exclusions.}
\begin{itemize}\setlength{\itemsep}{2pt}
  \item Agent's check returned negative and agent declared done anyway: \textsc{no match}.
  \item Agent ran NO check at all: \textsc{no match} (Unverified Claim).
  \item Lint/syntax/existence checks only: \textsc{no match} (Unverified Claim).
  \item Agent admitted incompleteness: \textsc{no match}.
\end{itemize}

\noindent{\large\bfseries Unverified Claim.}\quad \textsc{Trustworthy}\par
\smallskip

\textbf{Framing.}
\begin{itemize}\setlength{\itemsep}{2pt}
  \item The agent declared the task done WITHOUT running a substantive verification of its deliverable when one was available. Unverified Claim is a RESIDUAL rubric — it applies when the agent skipped verification AND no more-specific failure mode explains the verifier's rejection. PRIORITY ORDER — check more-specific rubrics FIRST, only fire UC if none of these apply to the failure:
  \item Verifier output is a syntax / compilation error → Syntax / Language Error, NOT UC.
  \item Verifier output is a shape / type / schema error → Wrong Output Schema, NOT UC.
  \item Verifier output is a missing-file / no-deliverable error → Silent Deliverable, NOT UC.
  \item Verifier output names a specific failing test / off-by-one / wrong axis → Algorithmic Bug, NOT UC.
  \item Trial timed out → Agent Timeout, NOT UC.
  \item Crash from missing dep / broken env → Environment Block, NOT UC.
  \item Agent ran a substantive check that returned positive → False Success Claim or Hidden-Test Regression, NOT UC.
  \item Agent emitted code in the wrong file / abstraction → Wrong Target / Layer, NOT UC. Only after ruling out the above do you consider UC.
\end{itemize}

\textbf{Firing trigger.}
\begin{itemize}\setlength{\itemsep}{2pt}
  \item The agent's deliverable is a defensible attempt (not silently missing, not syntactically broken, not blatantly wrong format) AND the agent did NOT run a substantive verification before declaring complete AND a verification mechanism was available (sample tests, oracle script, the repo's pytest, or simply running the deliverable end-to-end).
\end{itemize}

\textbf{Decision procedure.}
\begin{enumerate}\setlength{\itemsep}{2pt}
  \item Confirm none of the priority-rubrics above explain the failure.
  \item Confirm the agent declared complete without running substantive verification.
  \item Confirm a substantive verification was AVAILABLE (the task had tests, an oracle, or an end-to-end runnable check).
  \item \textsc{match} if 1+2+3 hold.
\end{enumerate}

\textbf{Exclusions.}
\begin{itemize}\setlength{\itemsep}{2pt}
  \item A more-specific rubric explains the failure: \textsc{no match} (use that rubric).
  \item Agent ran a substantive check successfully: \textsc{no match} (FSC or HTR).
  \item Agent admitted incompleteness: \textsc{no match}.
  \item Crashed/timed out before reaching verification: \textsc{no match}.
  \item Trial is short literal Q\&A with no programmatic checker: \textsc{no match}.
\end{itemize}

\noindent{\large\bfseries Plan Over Implementation.}\quad \textsc{Trustworthy}\par
\smallskip

\textbf{Framing.}
The agent narrates a CONCRETE NEXT ACTION ("I'll now write the file", "Let me modify X to do Y") OR an analysis/plan, and then the trajectory ends WITHOUT executing that action — no tool call, no file write, no patch applied. The required deliverable was never produced because the agent ran out of steps mid-plan.

Distinguishes from Silent Deliverable: SD = agent acted as if task was complete despite no artifact; PoI = agent stopped mid-plan, often with the next step explicitly described.

\textbf{Decision procedure.}
\begin{enumerate}\setlength{\itemsep}{2pt}
  \item Identify the required deliverable.
  \item Search the trajectory for tool calls that produced any version of the deliverable. If found: \textsc{no match}.
  \item Inspect the agent's final message(s). \textsc{match} if the agent describes a concrete next step ("I'll write...", "now let me modify...", "the next thing is...") OR finishes mid-analysis without ever executing the implementation step.
\end{enumerate}

\textbf{Exclusions.}
\begin{itemize}\setlength{\itemsep}{2pt}
  \item Deliverable was produced (even partially): \textsc{no match}.
  \item Agent declared 'task complete' or similar without an artifact: \textsc{no match} (Silent Deliverable).
  \item Agent admitted incompleteness, asked for help, or hit explicit timeout: \textsc{no match}.
  \item Crash before reaching implementation phase: \textsc{no match}.
\end{itemize}

\noindent{\large\bfseries Silent Deliverable.}\quad \textsc{Trustworthy}\par
\smallskip

\textbf{Framing.}
The task asked for a CONCRETE ARTIFACT at a specific location (a file, a function definition, a saved answer, a written patch) and NO EVIDENCE that the artifact was produced exists in the trajectory. The agent did not deliver.

PRIORITY EXCLUSION (check FIRST, before the firing triggers below): if the agent's FINAL message describes a concrete next action that was not executed ("I'll now write the file", "now let me modify X", "the next step is to write...") AND the trajectory ends mid-step, OUTPUT 'no match' here — that case is Plan Over Implementation, NOT Silent Deliverable. Silent Deliverable fires when the agent treats the task as done despite no artifact; Plan Over Implementation fires when the agent stops mid-narrative without finishing.

\textbf{Firing trigger.}
\begin{itemize}\setlength{\itemsep}{2pt}
  \item Verifier output explicitly reports the deliverable is missing / empty: phrases like 'No <file> found', '<file> does not exist', 'file not found', 'no answer.txt', 'missing output', 'output is empty', a Python \texttt{FileNotFoundError} / \texttt{KeyError} on the deliverable path, or any verifier message indicating the artifact wasn't there.
  \item Reward is null / missing AND no deliverable-producing tool call appears in the trajectory.
  \item The agent's trajectory lacks any tool call that would have written the deliverable (no Edit, no Write, no apply\_patch, no bash redirect to the required path) AND the agent treated the task as complete.
\end{itemize}

\textbf{Decision procedure.}
\begin{enumerate}\setlength{\itemsep}{2pt}
  \item Identify the required deliverable artifact and its location from the task instruction or verifier code.
  \item Check the verifier output for an explicit missing-file complaint (any of the trigger phrases above). If present → output \textsc{match}.
  \item Otherwise, search every tool call (file write, edit, apply\_patch, bash output, function call) for evidence the artifact was produced. If no such evidence exists AND the agent treated the task as complete → output \textsc{match}.
\end{enumerate}

\textbf{Exclusions.}
\begin{itemize}\setlength{\itemsep}{2pt}
  \item Deliverable IS produced (even partially or wrongly) somewhere in the trajectory: \textsc{no match} (use a content rubric).
  \item Agent narrated a concrete next write step but trajectory was cut off mid-action: \textsc{no match} (Plan Over Implementation).
  \item Agent admitted the work was incomplete or asked for help: \textsc{no match}.
  \item Hard timeout/crash before any deliverable could be produced: \textsc{no match}. When in doubt — and any of the strong firing triggers fired — lean \textsc{match}. This rubric is HIGH-RECALL on those signals; do not let broader rubrics (Algorithmic Bug, Plan Over Implementation) win when the artifact is genuinely missing.
\end{itemize}

\noindent{\large\bfseries Agent Timeout.}\quad \textsc{Trustworthy}\par
\smallskip

\textbf{Framing.}
The trial ended because the agent hit its wall-clock or step-count budget. Agent Timeout MAY co-occur with content rubrics (Algorithmic Bug, Wrong Factual Answer, Wrong Output Schema) when the agent's partial output also has issues — in those cases, BOTH rubrics should be shortlisted, not just one.

\textbf{Firing trigger.}
\begin{itemize}\setlength{\itemsep}{2pt}
  \item Verifier output / error fields contain \texttt{AgentTimeoutError}, \texttt{TimeoutError}, 'agent execution timed out', 'agent timeout exceeded', 'hit Ns timeout', 'timed out after N seconds'.
  \item Trial runtime is within \textasciitilde{}5\% of the agent's timeout budget.
  \item The agent's final message is mid-action ("I'll now write...", "let me run...", "next I'll...") and the trajectory was cut off without execution.
  \item \texttt{error\_type} / \texttt{terminate\_reason} indicates timeout / wall-clock / cut-off.
  \item \texttt{exception\_info} / traceback mentions timeout-related exceptions. When ANY trigger fires, output \textsc{match} for Agent Timeout. Note: this can co-occur with a content rubric — if the agent's partial output ALSO has a clear bug or wrong answer, shortlist BOTH (Agent Timeout AND the content rubric). The SHORTLIST allows up to 3 rubrics; co-occurring AT + content is common.
\end{itemize}

\textbf{Decision procedure.}
\begin{enumerate}\setlength{\itemsep}{2pt}
  \item Inspect verifier\_output / error fields / exception\_info / final agent message / runtime data for ANY timeout indicator.
  \item If found, output \textsc{match} for AT. Quote the timeout signal.
  \item If you'd ALSO have fired a content rubric (AB, WFA, WOS) on the partial output, you may shortlist that one as a co-rubric — but AT must be in the shortlist.
\end{enumerate}

\textbf{Exclusions.}
\begin{itemize}\setlength{\itemsep}{2pt}
  \item Trial finished naturally (agent declared done, no timeout signal): \textsc{no match}.
  \item Crash from missing dep / broken env: \textsc{no match} (Environment Block — different failure mode).
  \item Verifier timed out, not the agent: \textsc{no match}.
\end{itemize}

\noindent{\large\bfseries Environment Block.}\quad \textsc{Trustworthy}\par
\smallskip

\textbf{Framing.}
Agent attempted its work in good faith but hit a real environment constraint — missing data, blocked network, broken dependency, OR a required capability that the task scaffold / adapter did not install. Includes adapter-induced under-tooling: e.g., a figure-QA task where the adapter ships raw PNGs but no OCR / vision library.

\textbf{Decision procedure.}
\begin{enumerate}\setlength{\itemsep}{2pt}
  \item Scan for environment-failure signals:
  \begin{itemize}\setlength{\itemsep}{1pt}
    \item "network is unreachable", "Connection refused", "403 Forbidden", "404 Not Found"
    \item "package not found", "pip install failed", "ImportError: No module"
    \item Missing file / table / input / credential
    \item Agent tries vision/ASR/OCR/search tools that aren't installed (tesseract, pillow, cv2, whisper, curl)
    \item dbt source tables / DB schemas not present
  \end{itemize}
  \item Check that the agent attempted to use a legitimate means of accomplishing the task and got blocked — not merely skipped a step it could have done.
  \item \textsc{match} if (a) a clear environment constraint documented AND (b) agent didn't work around it AND (c) no reasonable workaround was available in the scaffold.
\end{enumerate}

\textbf{Exclusions.}
\begin{itemize}\setlength{\itemsep}{2pt}
  \item The "environment issue" is actually agent misuse (wrong command, wrong path): \textsc{no match}.
  \item Agent worked around the constraint, failed for other reasons: \textsc{no match}.
  \item Agent timed out trying to resolve: \textsc{no match} (Agent Timeout).
\end{itemize}

\medskip\noindent\textit{Excluded rubrics ($\kappa < 0.5$, folded into ``Others'').}

\noindent{\large\bfseries Calculation Mismatch.}\quad \textsc{Excluded}\par
\smallskip

\textbf{Framing.}
Tasks asking for a numerical result where the agent ran a pipeline to completion and produced a concrete number, but the value is outside the grader's tolerance.

\textbf{Decision procedure.}
\begin{enumerate}\setlength{\itemsep}{2pt}
  \item Task asks for numerical / statistical result.
  \item Agent ran a real pipeline: loaded data, applied a method, computed a number.
  \item Agent wrote the resulting number to the required location.
  \item reward=0, and failure is not a code crash / timeout / missing file.
  \item \textsc{match} if 1-4 hold.
\end{enumerate}

\textbf{Exclusions.}
\begin{itemize}\setlength{\itemsep}{2pt}
  \item Agent guessed without a pipeline: \textsc{no match} (Wrong Factual Answer).
  \item Pipeline crashed: \textsc{no match} (Syntax / Algorithmic).
  \item Task required symbolic, agent gave numerical: \textsc{no match}.
\end{itemize}

\noindent{\large\bfseries Rule Inference Error.}\quad \textsc{Excluded}\par
\smallskip

\textbf{Framing.}
The task tests the agent's ability to infer / apply a rule, scheme, mapping, or pattern from given examples. The agent demonstrably misunderstood or mis-extrapolated the rule itself. This is a content failure of REASONING, not coding, not retrieval, not formatting.

\textbf{Decision procedure.}
\begin{enumerate}\setlength{\itemsep}{2pt}
  \item Confirm the task is rule-inference style: visual puzzle, IQ-puzzle, pattern-completion, instruction-following with explicit examples-then-apply structure.
  \begin{itemize}\setlength{\itemsep}{1pt}
    \item If the task is straightforward Q\&A without explicit examples to generalize from: \textsc{no match} (Wrong Factual Answer).
    \item If the task is code-writing: \textsc{no match}.
  \end{itemize}
  \item Quote the specific rule the agent stated or applied, and the specific way it diverges from the correct rule (visible from the verifier output or oracle solution).
  \item \textsc{match} only when steps 1 + 2 hold and the failure is in rule inference / application — not other modes.  Default: when in doubt, output 'no match'.
\end{enumerate}

\noindent{\large\bfseries Specification–Verification Mismatch.}\quad \textsc{Excluded}\par
\smallskip

\textbf{Framing.}
The agent's deliverable is a DEFENSIBLE READING of the user-facing instruction, but the verifier rejects it because the verifier enforces a constraint NOT stated (or genuinely under-specified) in the instruction. Examples: verifier requires a specific function name not in the prompt; verifier requires a hardcoded path not in the prompt; verifier insists on the exact prose phrasing the prompt only suggested.

\textbf{Decision procedure.}
\begin{enumerate}\setlength{\itemsep}{2pt}
  \item Locate the user-facing instruction (instruction.md / first user message in <task\_context>).
  \item Locate the verifier's actual check (verifier\_code in <task\_context> / verifier\_output / test\_sh).
  \item Identify the SPECIFIC requirement the verifier enforced that caused the failure. Quote the exact failing assertion / expected value.
  \item Compare against the instruction. \textsc{match} only if BOTH:   (a) The agent's deliverable reasonably satisfies the instruction as a competent reader would understand it, AND   (b) The verifier's failing requirement is genuinely absent or contradictory in the instruction.
\end{enumerate}

\textbf{Exclusions.}
\begin{itemize}\setlength{\itemsep}{2pt}
  \item The verifier's requirement IS present in the instruction (even if obscure): \textsc{no match}.
  \item The agent's output is wrong by any plausible reading of the instruction: \textsc{no match} (use the appropriate content rubric).
  \item Agent didn't produce the deliverable at all: \textsc{no match}.
  \item Mismatch is purely formatting: \textsc{no match} (Wrong Output Schema / Wrong Output Location).
  \item Mismatch is about hidden tests probing behavior the prompt described: \textsc{no match} (Hidden-Test Regression).
  \item You cannot quote a specific verifier requirement absent from the instruction: \textsc{no match}. This rubric is for genuine UNDER-SPECIFICATION cases. Do not output 'match' merely because the agent failed and the verifier was strict. Default: when in doubt, output 'no match'.
\end{itemize}

\noindent{\large\bfseries Insufficient Web Research.}\quad \textsc{Excluded}\par
\smallskip

\textbf{Framing.}
Insufficient research applies to open-web / factual QA tasks where the agent SHOULD HAVE retrieved external information. The failure is that the retrieval was either absent when retrieval was feasible, single-sourced when corroboration was needed, or through a wrong channel.

\textbf{Decision procedure.}
\begin{enumerate}\setlength{\itemsep}{2pt}
  \item Confirm the task required web research.
  \begin{itemize}\setlength{\itemsep}{1pt}
    \item Internal tasks (code / math / local data): \textsc{no match}.
  \end{itemize}
  \item Check whether retrieval was FEASIBLE — did the scaffold provide web\_search / browser / curl to external URLs?
  \begin{itemize}\setlength{\itemsep}{1pt}
    \item Evidence of a shipped search tool: proceed.
    \item No search tool available, agent had nothing to call: \textsc{no match} (route to Environment Block).
  \end{itemize}
  \item Inventory retrieval actions. \textsc{match} if any of:   (a) Search tool was available but agent made zero retrieval attempts; answered from training knowledge   (b) Exactly one source consulted on an accuracy-sensitive question without corroboration   (c) Retrieval attempts uniformly blocked (403/404) and agent proceeded anyway despite alternative channels   (d) Used bash-scraping when a proper web tool was available
\end{enumerate}

\textbf{Exclusions.}
\begin{itemize}\setlength{\itemsep}{2pt}
  \item $\geq 2$ lookups with reconciliation, even if final answer is wrong: \textsc{no match} (Wrong Factual Answer).
  \item Tasks where training knowledge is plausibly sufficient: \textsc{no match}.
  \item Scaffold provided no retrieval tool: \textsc{no match} (Environment Block).
\end{itemize}

\noindent{\large\bfseries Premature Termination.}\quad \textsc{Excluded}\par
\smallskip

\textbf{Framing.}
Agent narrates a concrete next action ("I'll now write the file") in a final message, and the trajectory ends before the narrated action is executed. Often a scaffold artefact (e.g., codex+gpt-5-mini end-of-turn interpreted as \texttt{task\_complete}).

\textbf{Decision procedure.}
\begin{enumerate}\setlength{\itemsep}{2pt}
  \item Inspect the last 1-3 agent messages. Look for forward-looking commitment.
  \begin{itemize}\setlength{\itemsep}{1pt}
    \item Final message is a conclusion ("done", "here is X") without forward-looking verb: \textsc{no match}.
  \end{itemize}
  \item Check whether the narrated action was actually executed afterward (tool call, file write).
  \item \textsc{match} if (a) concrete next-action narrated AND (b) no corresponding tool action before trajectory ended.
\end{enumerate}

\textbf{Exclusions.}
\begin{itemize}\setlength{\itemsep}{2pt}
  \item Narrated and executed successfully: \textsc{no match}.
  \item Timed out mid-execution: \textsc{no match} (Agent Timeout).
  \item Never narrated anything concrete: \textsc{no match} (Silent Deliverable).
\end{itemize}

\noindent{\large\bfseries Wrong Output Location.}\quad \textsc{Excluded}\par
\smallskip

\textbf{Framing.}
Agent produced correct content but wrote it to the wrong path, wrong filename, wrong cell range.

\textbf{Decision procedure.}
\begin{enumerate}\setlength{\itemsep}{2pt}
  \item Identify the required output location.
  \item Observe where the agent actually wrote.
  \item \textsc{match} if content was written AND location deviates from required.
\end{enumerate}

\textbf{Exclusions.}
\begin{itemize}\setlength{\itemsep}{2pt}
  \item Never written: \textsc{no match} (Silent Deliverable).
  \item Correct location but wrong format: \textsc{no match} (Wrong Output Schema).
  \item Correct location but wrong content: \textsc{no match}.
\end{itemize}

\subsubsection{Agent Failure Mode Full Results}
\label{app:qual-results}
Table~\ref{tab:qual-full} reports per-rubric failure-mode prevalence for all six (model, harness) combinations across $N=6{,}028$ hard-task trajectories. The six reported rubrics are those with per-rubric judge--gold $\kappa \geq 0.5$; the remaining six rubrics (Wrong Target\,/\,Layer, Agent Timeout, Unverified Claim, Plan Over Implementation, False Success Claim, Wrong Output Schema) are aggregated into Others.

\begin{table}[H]
\centering
\small
\caption{Per-rubric failure-mode prevalence (\%) for all six (model, harness) combinations.
  WFA\,=\,Wrong Factual Answer, AB\,=\,Algorithmic Bug, HTR\,=\,Hidden-Test Regression,
  SD\,=\,Silent Deliverable, EB\,=\,Environment Block, Syn\,=\,Syntax\,/\,Language Error.
  Others aggregates six rubrics with judge--gold $\kappa < 0.5$.
  $N$ is the number of hard-task trials.}
\label{tab:qual-full}
\setlength{\tabcolsep}{5pt}
\begin{tabular}{llrrrrrrrr}
\toprule
\textbf{Model} & \textbf{Harness} & $N$ & \textbf{WFA} & \textbf{AB} & \textbf{HTR} & \textbf{SD} & \textbf{EB} & \textbf{Syn} & \textbf{Others} \\
\midrule
GPT-5.4        & Codex        & 991 & 27.3 & 20.2 & 9.6 & 1.4 & 3.5 & 0.6 & 17.9 \\
GPT-5.4        & Terminus-2   & 983 & 26.6 & 18.8 & 4.9 & 4.9 & 9.3 & 1.4 & 18.2 \\
Opus 4.6       & Claude Code  & 973 & 26.4 & 22.5 & 6.9 & 4.1 & 2.9 & 0.8 & 17.7 \\
Opus 4.6       & Terminus-2   & 974 & 23.2 & 17.7 & 10.0 & 5.0 & 4.2 & 1.4 & 24.3 \\
Gemini 3.1 Pro & Gemini CLI   & 995 & 25.0 & 17.7 & 8.7 & 6.9 & 3.9 & 1.2 & 16.7 \\
Gemini 3.1 Pro & Terminus-2   & 973 & 24.5 & 15.3 & 8.8 & 7.3 & 4.1 & 1.4 & 17.9 \\
\bottomrule
\end{tabular}
\end{table}

\paragraph{Reading the table.}
A cell reports the fraction of trials in that (model, harness) row where the rubric fires; because the judge may assign multiple rubrics to a single trial, row percentages do not sum to 100\%. The capability rubrics (WFA, AB, HTR) are broadly conserved across harnesses within each model family, while the operational rubrics (SD, EB) show harness-dependent variation. Environment Block is notably elevated for GPT-5.4 under Terminus-2 (9.3\% vs.\ 3.5\% under Codex), consistent with the protocol-incompatibility pattern described in \S\ref{sec:analysis-quali}. Silent Deliverable is highest on Gemini CLI (6.9\%) and Opus~4.6 under Terminus-2 (5.0\%), reflecting differences in submission-protocol enforcement across harnesses.

\section{Harbor-Index Selection Pipeline}
\label{app:harbor_index}
\label{sec:harbor_index_appendix}

This appendix documents the per-cell sampling protocol used to derive the difficulty filter (\Cref{app:harbor_index_sampling}), the full task-quality rubric used by both the AI auditor and human reviewers (\Cref{app:harbor_index_rubric}), the judge orchestration and prompt (\Cref{app:harbor_index_judge}), the human-review protocol (\Cref{app:harbor_index_human_review}), and the audit-and-fix stage that yields the final \HarborIndexTaskCount{}-task release (\Cref{app:harbor_index_fix}).

\subsection{Sampling Protocol}
\label{app:harbor_index_sampling}

Each candidate task is summarized by an 18-trial sample drawn from the live Harbor trial database. We define the \emph{filtering mix} as the Cartesian product of three leading model families at the time of filtering and two harness configurations:
\begin{itemize}
\item \textbf{Models:} Claude Opus 4.6, GPT-5.4, Gemini 3.1 Pro.
\item \textbf{Harnesses:} each model's native harness (Claude-Code for Anthropic, Codex for OpenAI, Gemini-CLI for Google), plus the cross-vendor Terminus-2 agent.
\end{itemize}
This yields 6 $(\text{model}, \text{harness})$ pairs. For each pair we keep 3 valid trials, giving 18 trials per task.

\paragraph{Difficulty filter.}
A task passes the filter iff $n_\mathrm{succ} \leq 6$ over its 18-trial sample, where $n_\mathrm{succ}$ denotes the number of successful trials (i.e., the success rate over the filtering mix must be at most $33\%$). Success is benchmark-aware. For most benchmarks, a trial counts as a success iff $\text{reward} > 0$. For benchmarks whose reward semantics differ from a binary pass/fail, we follow the per-benchmark scoring cutoff established by the main experiment.

\subsection{Task Quality Bar}
\label{app:harbor_index_rubric}

The auditor and human reviewers evaluate each candidate against two criteria. A task is rejected if it fails either; a reviewer may also mark a task \emph{unsure} when the evidence is mixed. The same rubric is shared between the AI judge and the human review rounds so that the two stages enforce a consistent definition of task quality.

\paragraph{Test--instruction alignment.}
Every test assertion must trace back to a requirement stated in the instruction or implied by the environment, and every requirement in the instruction must have corresponding test coverage. The criterion fails when:
\begin{itemize}
\item the verifier checks something the instruction never asks for or implies, such as a specific numerical value, exact error wording, file format, or schema;
\item the instruction is ambiguous or under-specified about the required output while the verifier asserts a single specific answer;
\item multiple correct interpretations of the instruction would each produce a different verifier outcome;
\item the agent gets the right idea but the verifier rejects it on a clerical detail not foreshadowed by the instruction.
\end{itemize}

\paragraph{Essential difficulty.}
Difficulty must come from genuine reasoning, algorithmic thinking, domain expertise, long-horizon interaction, or multi-step execution, not from minor formatting details, arbitrary precision, ambiguous output schemas, or clerical detail. The criterion fails when trials repeatedly show models arriving at the right idea but being tripped up by:
\begin{itemize}
\item exact whitespace, decimal precision, or output column ordering not specified in the instruction;
\item magic strings (function names, error messages, JSON keys) the instruction does not pin down;
\item tolerance bounds the verifier applies that the instruction does not disclose.
\end{itemize}

\paragraph{Calibration set.}
The auditor receives a small set of pre-labeled rejected tasks as in-context calibration examples (e.g., a SWE-bench Multilingual task whose verifier outcome depends on wall-clock year; a SWE-bench Pro task whose verifier requires the literal wording of an \texttt{AnsibleError} not pinned down by the PR; a Spider2 task whose gold output was generated by a buggy Haversine formula and whose verifier penalizes correct fixes). Tasks exhibiting similar smells, such as verifier--instruction mismatch, fragile string equality on novel content, under-specified output format, or gold that does not pass the verifier, are rejected by analogy.

\subsection{Task Audit Process}
\label{app:harbor_index_judge}

\paragraph{Setup.}
The AI auditor used at the second pipeline stage is based on Gemini-3-Flash, run inside a per-task Daytona sandbox via Claude Code in non-interactive print mode. The orchestrator packages, per task, a tarball containing the rubric, a corpus JSON file (18 trials with verifier stdouts and ATIF trajectory paths), the trajectories themselves, and the per-trial verifier stdouts; uploads it to a fresh sandbox; runs the audit; and downloads the resulting verdict. The judge is granted its native filesystem and shell tools (\texttt{Read}, \texttt{Grep}, \texttt{Bash}, \texttt{Write}) and is required to consult between two and four trajectories per task---mixing successes, failures, and exception cases---and to cite specific step IDs, tool calls, and observations in its rationale. The verdict is emitted as a single JSON object written to a file via the \texttt{Write} tool; a Python fallback parses an inline JSON object from the streamed assistant turn if the tool call is missing.

\paragraph{Cross-judge calibration.}
We ran the same harness with Opus-4.7 as the auditor model over a 74-task subset of the candidate pool to gauge inter-judge reliability against Gemini-3-Flash. The two judges agree on $66/74 = 89.2\%$ of decisions across the three-way \{accept, reject, unsure\} label space, with Cohen's $\kappa = 0.747$ (substantial agreement). Collapsed to the operationally relevant binary \{accept, non-accept\}, agreement rises to $69/74 = 93.2\%$: the two judges agree on accepts (16) and non-accepts (53) together for 69 of the 74 shared tasks, with the five disagreements concentrated in tasks Opus-4.7 marked \emph{accept} but Gemini-3-Flash marked \emph{reject}. Thus, the cheaper \texttt{gemini-3-flash-preview} judge is more conservative on this subset: there are no cases where it accepts a task that \texttt{opus-4.7} rejects. We use \texttt{gemini-3-flash-preview} as the production auditor on this basis.

\paragraph{Verdict schema.}
Every verdict is a JSON object with the following fields: \texttt{decision} (one of \emph{accept}, \emph{reject}, \emph{unsure}); \texttt{primary\_reasons} (short tags); \texttt{rubric} (per-criterion verdict and evidence); \texttt{failure\_attribution} (\emph{task\_design}, \emph{agent\_weakness}, \emph{infra\_or\_experiment}, or \emph{mixed}); \texttt{trial\_comparison\_notes} (cross-trial observations grounded in trajectory step IDs); \texttt{trajectories\_consulted}; \texttt{argument\_reasoning\_evidence} (detailed argumentative reasoning grounded in cited code lines, commands, or agent steps); and \texttt{summary} (a one-paragraph rationale).

\paragraph{Prompt.}
We provide the full judge prompt in \Cref{lst:harbor_index_judge_prompt}.

\begin{lstlisting}[
  basicstyle=\ttfamily\footnotesize,
  breaklines=true,
  breakatwhitespace=true,
  columns=fullflexible,
  frame=single,
  framesep=4pt,
  framerule=0.4pt,
  xleftmargin=6pt,
  xrightmargin=6pt,
  showstringspaces=false,
  captionpos=b,
  caption={Harbor-Index task-quality judge prompt (\texttt{judge\_prompt.md}).},
  label={lst:harbor_index_judge_prompt}
]
# Task quality audit

You are an expert reviewer evaluating the quality of tasks.

## What you receive

A single corpus JSON file describing one candidate task with:
- benchmark, task_path, n_succ (= number of successful trials out of 18
  frontier-model attempts; range 0-6 for the bulk pool, may be higher for
  under-rep fill-ins)
- task_files: full text of instruction.md, task.toml,
  environment/Dockerfile, solution/solve.sh, tests/test.sh
- trials: each with model, agent, reward, exception_type, test_stdout
  (verifier stdout, head/tail-clipped to ~32 KB), and trajectory_path
  (path to ATIF agent trajectory JSON)

You may also be given a task_dir path and file-system access to read
additional files in that directory if needed.

## Your job

Decide accept, reject, or unsure on the task based on quality audit. We
accept a task if it passes the rubrics that demonstrate high quality; we
reject a task if it has noticeable problems. Mark "unsure" if you are not
confident in your judgment. List detailed reasoning, evidence, and
arguments along with your final verdict.

## Quality rubric

### 1. test_instruction_alignment
Every test assertion must trace back to a requirement stated in
instruction.md or implied by the environment, and every requirement in
the instruction must have corresponding test coverage. Tests should NOT
introduce requirements beyond what the instruction describes. FAIL if:
- The verifier checks something the instruction never asks for or never
  implied (e.g., a specific numerical value, exact error wording, file
  format, or schema not stated in the instruction).
- The instruction is ambiguous or under-specified about what the agent
  should produce, while the verifier asserts a specific answer.
- Multiple correct interpretations of the instruction would each produce
  a different verifier outcome.
- The agent gets the right *idea* but the verifier rejects them on a
  clerical detail not foreshadowed by the instruction.

### 2. essential_difficulty
Difficulty must come from genuine reasoning, algorithmic thinking, domain
expertise, long-horizon interactions, multi-step execution, etc -- NOT
from formatting minutiae, arbitrary precision, ambiguous output schemas,
or clerical detail. FAIL if the trials show models repeatedly arriving
at the right idea but getting tripped up on:
- Exact whitespace / decimal precision / output column ordering not
  specified in instruction.
- Specific magic strings (function names, error messages, JSON keys) the
  instruction doesn't pin down.
- Tolerance bounds the verifier applies that the instruction doesn't
  disclose.

## How to use trial data

You have several test_stdouts plus task files. Use them to ground your
judgment:

1. Compare successful vs failed trial test_stdouts. What does the success
   pattern look like -- was it luck on a fragile verifier, or genuine
   task completion following instruction? What did the failures actually
   fail on -- insufficient domain knowledge and/or task understanding,
   wrong solution idea/execution/calculation, formatting, timeout, infra?
   If unclear, read the trajectories.
2. Compare successful trial output to solve_sh. NOTE that solve_sh is a
   rough reference solution -- it does not mean that the agent or "only
   correct" solution should do the same thing to complete the task. It
   is only material for you to compare a potentially correct approach
   against agent approaches. All your judgment should still be grounded
   in the task description and test verifiers.
3. Look for infra/experiment failures masquerading as task failures.
   Patterns to flag (mark as unsure):
   - 429 Too Many Requests, rate limiting from model API.
   - No space left on device, OOM errors mid-execution.
   - Network errors / DNS / proxy connect failures.
   - Container build failures unrelated to the task design.
   - Truncated transcripts due to step-count exhaustion.
   These are experiment issues, not task issues. If 5 of 6 trials failed
   for infra reasons and 1 succeeded cleanly, the task may still be
   high-quality -- accept if other criteria pass.

## Output format

Output a single JSON object (no preamble, no trailing markdown):

{
  "decision": "accept" | "reject" | "unsure",
  "primary_reasons": ["short tag", ...],
  "rubric": {
    "test_instruction_alignment": {"verdict": "pass"|"fail"|"na",
                                   "evidence": "..."},
    "essential_difficulty":       {"verdict": "pass"|"fail"|"na",
                                   "evidence": "..."}
  },
  "failure_attribution": "task_design" | "agent_weakness" |
                         "infra_or_experiment" | "mixed",
  "trial_comparison_notes": "What you noticed comparing trajectories and
    test_stdouts (successes vs failures, cross-model). Cite specific
    step IDs / tool calls / observations from at least one trajectory
    you rendered.",
  "trajectories_consulted": ["<trial_id_1>", "<trial_id_2>", ...],
  "argument_reasoning_evidence": "Show your judgment of the task quality.
    Claim your arguments. Show detailed reasoning with concrete evidence
    (specific to words/commands/code-lines/agent-steps if helpful) to
    demonstrate. If you claim the task to be 'unsure', show what you
    are confident and unconfident about respectively, and what you want
    human reviewers to double-check.",
  "summary": "One-paragraph rationale for the decision, citing specific
    files/lines/strings."
}
\end{lstlisting}

\subsection{Human Review Protocol}
\label{app:harbor_index_human_review}

Human review is conducted by a pool of 14 domain-experienced reviewers using the same task-quality rubric as the AI auditor. The process has three stages:
\begin{enumerate}
\item \textbf{Initial quality screen.} A task handled by a senior reviewer receives at least one review; a task handled by junior reviewers receives at least two reviews. Reviewers inspect the task specification, environment, verifier, and available execution evidence, and may accept, flag for repair, reject, or escalate an uncertain case. This screen leaves more than 110 candidates.
\item \textbf{Panel selection.} A panel of three senior reviewers discusses the surviving candidates and selects an intermediate pool of 100 based on task quality, difficulty, diversity, and the insight offered by model behavior.
\item \textbf{Trajectory-grounded audit.} Each of the 100 selected tasks is re-examined by at least two senior reviewers. Reviewers inspect trajectories and verifier outcomes, perform failure analysis, and distinguish true and false positives and negatives. They repair fixable task defects and remove tasks that remain broken or become too easy after repair, producing the final \HarborIndexTaskCount{}-task release.
\end{enumerate}

This process is an iterative engineering curation workflow rather than a fixed-panel annotation study: the review method was refined during construction, and reviewer counts therefore vary by task. We report the minimum review coverage and adjudication process above rather than a single human inter-rater-agreement statistic. Additional audit examples and task-level artifacts are available at \url{https://harbor-index.org}.

\subsection{Audit-and-Fix Loop}
\label{app:harbor_index_fix}

The senior panel's initial cut yields an intermediate pool of 100 candidate tasks. We then run repeated rounds of automated audit and reviewer repair: an LLM judge grades the verifier in a fresh copy of each task's sandbox by reading agent trajectories, verifier outputs, tests, and the reference solution, and labels each rollout as a true/false positive or negative with citations to specific steps or files. We prioritize false positives that let agents pass without solving the task and false negatives that reject correct solutions. Tasks are repaired (or dropped when irreparable), the filtering-mix models are re-run, and the audit repeats until the set stabilizes. Tasks that remain broken or become too easy after repair are removed, leaving the final Harbor-Index~1.0 release of \textbf{\HarborIndexTaskCount{} tasks across \HarborIndexBenchmarkCount{} benchmarks}. We also tighten per-task timeouts in this stage and adopt Harbor features such as separate verifier sandboxes to close shared-container reward hacks across the suite.

\subsection*{Harbor-Index~1.0 Task Catalog}

Table~\ref{tab:harbor-index-task-catalog} summarizes the \HarborIndexTaskCount{} release tasks by domain and benchmark.
Task IDs and interactive per-task scores for all \HarborIndexRolloutCount{} rollouts are published at \url{https://harbor-index.org} and on the \href{https://hub.harborframework.com/datasets/harbor-index/harbor-index-1.0}{Harbor Hub}.

\begin{table}[t]
\centering
\caption{Harbor-Index~1.0 task counts by domain and benchmark (\HarborIndexTaskCount{} tasks / \HarborIndexBenchmarkCount{} benchmarks).}
\label{tab:harbor-index-task-catalog}
\small
\begin{tabular}{llr}
\toprule
Domain & Benchmark & \#Tasks \\
\midrule
\multirow{10}{*}{Software Engineering} & GSO & 7 \\
 & SWE-Bench Verified & 5 \\
 & AlgoTune & 5 \\
 & FeatureBench & 4 \\
 & SWE-Bench Pro & 4 \\
 & SWE-Lancer & 2 \\
 & BigCodeBench & 1 \\
 & USACO & 1 \\
 & SWE-smith & 1 \\
 & SWT Bench & 1 \\
\midrule
\multirow{7}{*}{Scientific Research} & BIX-Bench & 5 \\
 & LAB-Bench & 4 \\
 & SciCode & 3 \\
 & SLDBench & 1 \\
 & Replication Bench & 1 \\
 & CodePDE & 1 \\
 & QCircuit Bench & 1 \\
\midrule
\multirow{5}{*}{Agents, Tools \& Systems} & GAIA2 & 5 \\
 & GAIA & 3 \\
 & Terminal Bench 2 & 3 \\
 & SkillsBench & 2 \\
 & WideSearch & 1 \\
\midrule
\multirow{2}{*}{Knowledge} & HLE & 8 \\
 & GPQA Diamond & 1 \\
\midrule
\multirow{2}{*}{Mathematics \& Reasoning} & ARC-AGI-2 & 5 \\
 & OmniMath & 2 \\
\midrule
\multirow{2}{*}{Data \& Analytics} & Spider 2 & 2 \\
 & DA-Code & 1 \\
\midrule
\multirow{1}{*}{Safety \& Security} & Cyber Gym & 2 \\
\bottomrule
\end{tabular}
\end{table}

\section{Limitations}
\label{sec:app_limitations}

Apart from the limitations listed in~\Cref{sec:conclusion}, we discuss extended limitations here.



\textbf{Evaluation cost and scalability.}
Large-scale agent evaluation remains expensive. Our study uses substantial compute resources and token budgets, which may limit reproducibility for smaller research groups. While Harbor-Index reduces cost, it is still a proxy for the full benchmark distribution.

\textbf{LLM-as-a-judge limitations.}
Some tasks rely on automated evaluation using language models as judges. Even though we aggregate the evaluation from 4 different models from different model providers and repeated runs for each model, these evaluations may still introduce bias, variance, or systematic errors, especially for open-ended or subjective tasks. 



\textbf{Ethical considerations.}
Our work focuses on evaluation infrastructure and does not directly address downstream societal impacts such as misuse of agents or deployment risks. Future work should consider safety, alignment, and governance aspects of agentic systems.

Addressing these limitations is an important direction for future work, including expanding benchmark diversity, improving adapter verification, reducing evaluation cost, and designing more realistic and dynamic evaluation protocols.

\section{Broader Impacts}
\label{sec:app_broader_impacts}

This work introduces Harbor Adapters and Harbor-Index, aiming to improve the scalability, reproducibility, and coverage of agentic evaluation. We discuss potential positive impacts as well as risks and mitigation strategies.

\textbf{Positive impacts.}
Our infrastructure lowers the barrier to evaluating language-model agents across diverse benchmarks, promoting more comprehensive and reproducible comparisons. By enabling large-scale and standardized evaluation, this work may help the community better understand model capabilities, identify failure modes, and avoid overfitting to a small set of popular benchmarks. In the long term, improved evaluation can contribute to the development of more reliable and robust AI systems in domains such as software engineering, scientific research, and data analysis.

\textbf{Risks and potential misuse.}
Standardized evaluation frameworks may also accelerate competitive benchmarking and optimization, potentially encouraging overfitting to benchmark suites rather than real-world performance. In addition, improved evaluation infrastructure could indirectly support the development of more capable autonomous agents, which may raise concerns around misuse, including automation of harmful tasks, large-scale exploitation, or deployment without sufficient safeguards.


\textbf{Mitigation and responsible use.}
We partially mitigate these risks by emphasizing compactness, diversity, difficulty, and quality in Harbor-Index construction, and by releasing detailed analyses of failure modes. We encourage users to interpret results cautiously, avoid over-reliance on single benchmarks, and complement our evaluation with domain-specific and real-world testing. Future work should further incorporate safety, alignment, and governance considerations into agent evaluation frameworks.

Overall, we believe that improving evaluation infrastructure is a necessary step toward building more transparent, reliable, and accountable AI systems, but it must be accompanied by responsible use and continued scrutiny.
\end{document}